\pdfoutput=1

\documentclass{article}

    \usepackage[preprint]{neurips_2022}

\usepackage[utf8]{inputenc} % allow utf-8 input
\usepackage[T1]{fontenc}    % use 8-bit T1 fonts
\usepackage{url}            % simple URL typesetting
\usepackage{booktabs}       % professional-quality tables
\usepackage{amsfonts}       % blackboard math symbols
\usepackage{nicefrac}       % compact symbols for 1/2, etc.
\usepackage{microtype}      % microtypography
\usepackage{xcolor}         % colors
\usepackage{amsmath}
\usepackage{tcolorbox}
\usepackage{hyperref}
\hypersetup{colorlinks = true,
            linkcolor = blue,
            urlcolor  = blue,
            citecolor = blue,
            anchorcolor = blue}
\usepackage{multirow}
\usepackage{xcolor}
\usepackage{adjustbox}

\newcommand{\cmmnt}[1]{\ignorespaces}

\usepackage{amssymb}
\definecolor{tableau10_C2}{RGB}{44, 160, 44}
\definecolor{tableau10_C3}{RGB}{214, 39, 40}
\definecolor{purple}{RGB}{60, 19, 97}
\definecolor{darkgreen}{RGB}{0, 100, 0}
\usepackage{subfigure}
\usepackage{tikz}

\title{The Effect of Diversity in Meta-Learning}

\newcommand\blfootnote[1]{%
  \begingroup
  \renewcommand\thefootnote{}\footnote{#1}%
  \addtocounter{footnote}{-1}%
  \endgroup
}

\author{%
  Ramnath Kumar $^{1,*}$, 
  Tristan Deleu$^{2}$,
  Yoshua Bengio$^{2,3}$
}

\begin{document}

\maketitle

\begin{abstract}
Few-shot learning aims to learn representations that can tackle novel tasks given a small number of examples. Recent studies show that task distribution plays a vital role in the model's performance. Conventional wisdom is that task diversity should improve the performance of meta-learning. In this work, we find evidence to the contrary; we study different task distributions on a myriad of models and datasets to evaluate the effect of task diversity on meta-learning algorithms. For this experiment, we train on multiple datasets, and with three broad classes of meta-learning models - Metric-based (i.e., Protonet, Matching Networks), Optimization-based (i.e., MAML, Reptile, and MetaOptNet), and Bayesian meta-learning models (i.e., CNAPs). Our experiments demonstrate that the effect of task diversity on all these algorithms follows a similar trend, and task diversity does not seem to offer any benefits to the learning of the model. Furthermore, we also demonstrate that even a handful of tasks, repeated over multiple batches, would be sufficient to achieve a performance similar to uniform sampling and draws into question the need for additional tasks to create better models. \blfootnote{$^{*}$ Work done during an internship at Mila; Correspondence author \href{mailto:ramnathk@google.com}{ramnathk@google.com}. $^1$ Google Research, India. $^2$Mila, Qu\'ebec Artificial Intelligence Institute, Université de Montréal. $^{3}$CIFAR, IVADO.}
\end{abstract}

\section{Introduction}

It is widely recognized that humans can learn new concepts based on very little supervision, i.e., with few examples (or "shots"), and generalize these concepts to unseen data as mentioned by \cite{lake2011one}. On the other hand, recent advances in deep learning have primarily relied on datasets with large amounts of labeled examples, primarily due to overfitting concerns in low data regimes. Although better data augmentation and regularization techniques can alleviate these concerns, many researchers now assume that future breakthroughs in low data regimes will emerge from meta-learning, or "learning to learn."\newline

Here, we study the effect of task diversity in the low data regime and its impact on various models. In this meta-learning setting, a model is trained on a handful of labeled examples at a time under the assumption that it will learn how to correctly project examples of different classes and generalize this knowledge to unseen labels at test time.
Although this setting is often used to illustrate the remaining gap between human capabilities and machine learning, we could argue that the domain of meta-learning is still nascent. The field of task selection has mainly remained under-explored in this setting. Hence, our exploration of this setting is much warranted. To the best of our knowledge, no previous work attempts to work with task diversity and its effect in the meta-learning setting. \newline

Conventional wisdom is that the model's performance will improve as we train on more diverse tasks. This does seem intuitively sound: training on a diverse and large amount of classes should bring about a more extensive understanding of the world, thus learning multiple concepts of, let's say, the "world model". To test this hypothesis, we define task samplers that either limit task diversity by selecting a subset of overall tasks or improve task diversity using approaches such as Determinantal Point Processes (DPPs) proposed by \cite{macchi1975coincidence}. This problem is interesting since understanding the effect of diversity in meta-learning is closely linked to the model's ability to learn. In hindsight, this study is also an excellent metric to test the efficacy of our models, as will become more substantial in further sections.

\subsection{\textbf{Contributions}} 
In this section, we present the main contributions of the paper:
\begin{itemize}
    \item We show that limiting task diversity and repeating the same tasks over the training phase allows the model to obtain performances similar to models trained on Uniform Sampler without any adverse effects. (Section~\ref{exp},\ref{disc})
    \item We also show that increasing task diversity using sophisticated samplers such as DPP or Online Hard Task Mining (OHTM) Samplers does not significantly boost performance. Instead, this also harms the performance of the learner in certain instances. (Section~\ref{exp},\ref{disc})
    \item We also propose a suitable theoretical explanation for our findings from the connection to Simpson's paradox phenomenon from the discipline of causality as discussed briefly in Appendix~\ref{theory}. 
    \item We also propose a metric to compute task diversity in the meta-learning setting, something non-existent in previous literature. (Section~\ref{td})
    \item Our findings bring into question the efficiency of the model and the advantage it gains with access to more data using samplers such as the standard sampling regime -- Uniform Sampler. If we can achieve similar performances with fewer data, the existing models have not taken advantage of the excess data it is provided with.
\end{itemize}

\section{Background}

\begin{figure*}[th]
    \centering
    \begin{adjustbox}{center}
    \includegraphics[width=0.99\linewidth]{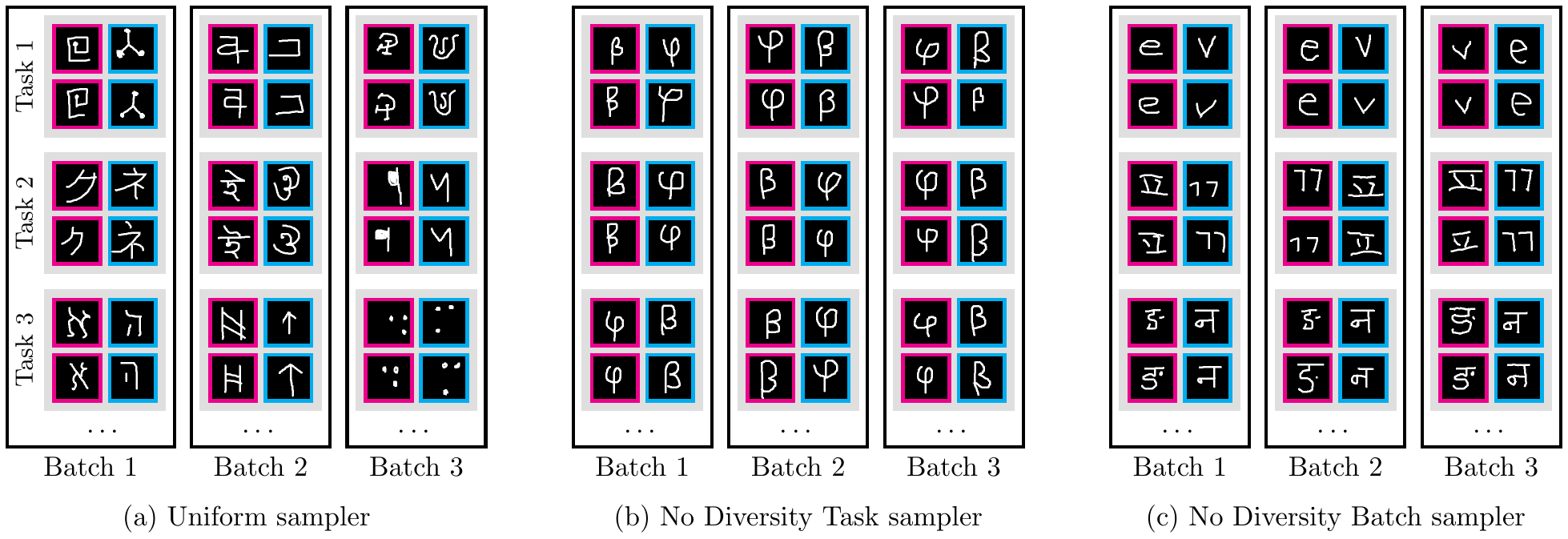}
    \end{adjustbox}
    \vspace*{-1.5em}
    \caption{Illustration of (a) the Uniform Sampler, (b) the No Diversity Task Sampler, and (c) the No Diversity Batch Sampler.}
    \label{fig:task_sampling_low_diversity}
\end{figure*}

Here, we review some of the fundamental ideas required to better understand our few-shot learning experiments.
\subsection{Episodic few-shot learning}
In episodic few-shot learning, an episode is represented as a $N$-way, $K$-shot classification problem where $K$ is the number of examples per class and $N$ is the number of unique class labels. During training, the data in each episode is provided as a support set $S = \{(x_{1,1},y_{1,1}), ..., (x_{N,K}, y_{N,K})\}$ where $x_{i,j} \in \mathbb{R}^D$ is the $i$-th instance of the $j$-th class, and $y_{j} \in \{0,1\}^N$ is its corresponding one-hot labeling vector. Each episode aims to optimize a function $f$ that classifies new instances provided through a ``query'' set $Q$, containing instances of the same class as $S$. This task is difficult because $K$ is typically very small (e.g. 1 to 10). The classes change for every episode. The actual test set used to evaluate a model does not contain classes seen in support sets during training. In the task-distribution view, meta-learning is a general-purpose learning algorithm that can generalize across tasks and ideally enable each new task to be learned better than the last. We can evaluate the performance of $\omega $ over a distribution of tasks $p(\tau)$. Here we loosely define a task to be a dataset and loss function $\tau = \{\mathcal{D}_{\tau},\mathcal{L}_{\tau}\}$. Learning how to learn thus becomes:
\begin{equation}
    \min_{\omega}\underset{\tau \sim p(\tau)}{\mathbb{E}}\left[ \mathcal{L}_{\tau}(\mathcal{D}_{\tau};\omega)\right],
\end{equation}
where $\mathcal{L}(\mathcal{D};\omega)$ measures the performance of a model trained using network parameters $\omega$ on dataset $\mathcal{D}$, and $p(\tau)$ indicates the task distribution. In our experiments, we extend this setting such that we vary the task diversity of the train split to study the effects on test split, which remains unchanged (i.e. uniformly sampling test tasks).

\subsection{Determinantal Point Processes (DPPs)}
A Determinantal Point Process \citep[DPP;][]{kulesza2012determinantal} is a probability distribution over subsets of a ground set $\mathcal{Y}$, where we assume $\mathcal{Y}=\left \{ 1,2,\ldots,N \right \}$ and $N = |\mathcal{Y}|$. An $\mathbf{L}$-ensemble defines a DPP using a real, symmetric, and positive-definite matrix $\mathbf{L}$ indexed by the elements of $\mathcal{Y}$. The probability of sampling a subset $Y=A\subseteq \mathcal{Y}$ can be written as:
\begin{equation}
    P(Y=A)\propto \det \mathbf{L}_A, 
\end{equation}
where $\mathbf{L}_A := [L_{i,j}]_{i,j\in A} $ is the restriction of $\mathbf{L}$ to the entries indexed by the elements of $A$. As $\mathbf{L}$ is a positive semi-definite, there exists a $d\times N$ matrix $\Psi$ such that $\textbf{L} = \Psi^T\Psi$ where $d \leq N$. Using this principle, we define the probability of sampling as:
\begin{equation}
\label{dpp}
    P(Y=A) \propto \det \textbf{L}_A = \mathrm{vol}^2(\{\Psi_i\}_{i\in A}),
\end{equation}
where the RHS is the squared volume($\mathrm{vol}$) of the parallelepiped spanned by $\{\Psi_i\}_{i\in A}$. In Eq. \ref{dpp}, $\Psi_i$ is defined as the feature vector of element $i$, and each element $L_{i,j}$ in $\mathbf{L}$ is the similarity measured by dot products between elements $i$ and $j$. Hence, we can verify that a DPP places higher probabilities on diverse sets because the more orthogonal the feature vectors are, the larger the volume parallelepiped spanned by the feature vector is. In this work, these feature embeddings represent class embeddings, which are derived using either a pre-trained Prototypical Network \citep{snell2017prototypical} model or the model being trained as discussed in Sec. \ref{task_sampling}.  

In a DPP, the cardinality of a sampled subset, $|A|$, is random in general. A $k$-DPP \cite{kuhn2003higher} is an extension of the DPP where the cardinality of subsets are fixed as $k$ (i.e., $|A|=k$). In this work, we use $k$-DPPs as an off-the-shelf implementation to retrieve classes that represent a task used in the meta-learning step.

\subsection{Task Sampling}
\label{task_sampling}

In this work, we experiment with eight distinct task samplers, each offering a different level of task diversity. To illustrate the task samplers, we use a 2-way classification problem, and denote each class with a unique alphabet from the Omniglot dataset \citep{lake2011one}. To make our study more theoretically sound and less heuristic in nature, we create a more formal definition of task diversity and discuss it in more detail in Section~\ref{td}.

\paragraph{\textbf{Uniform Sampler}} This is the most widely used Sampler used in the setting of meta-learning (with mutually-exclusive tasks \citep{yin2019meta}). The Sampler creates a new task by sampling uniformly classes. An illustration of this Sampler is shown in Figure~\ref{fig:task_sampling_low_diversity}.

% \begin{align}
% \label{Uniform_sampler}
% \underset{batch\ 1}{\boxed{\underset{task\ 1}{\boxed{R, K}}\quad \underset{task\ 2}{\boxed{T, D}}}}\quad \underset{batch\ 2}{\boxed{\underset{task\ 1}{\boxed{Y, B}}\quad \underset{task\ 2}{\boxed{A, F}}}}\quad
% ...\quad
% \underset{batch\ n}{\boxed{\underset{task\ 1}{\boxed{Z, K}}\quad \underset{task\ 2}{\boxed{L, D}}}}
% \end{align}

\paragraph{\textbf{No Diversity Task Sampler}} In this setting, we uniformly sample one set of the task at the beginning and propagate the same task across all batches and meta-batches. Note that repeating the same class over and over again does not simply repeat the same images/inputs as we episodically retrieve different images for each class. An illustration of this Sampler is shown in Figure~\ref{fig:task_sampling_low_diversity}.

% \begin{align}
% \label{ndt_sampler}
% \underset{batch\ 1}{\boxed{\underset{task\ 1}{\boxed{M, M}}\quad \underset{task\ 2}{\boxed{M, M}}}}\quad \underset{batch\ 2}{\boxed{\underset{task\ 1}{\boxed{M, M}}\quad \underset{task\ 2}{\boxed{M, M}}}}\quad
% ...\quad
% \underset{batch\ n}{\boxed{\underset{task\ 1}{\boxed{M, M}}\quad \underset{task\ 2}{\boxed{M, M}}}}
% \end{align}
\paragraph{\textbf{No Diversity Batch Sampler}} In this setting, we uniformly sample one set of tasks for batch one and propagate the same tasks across all other batches. Furthermore, we shuffle the labels, as in the No Diversity Task Sampler, to prevent the model from overfitting. An illustration of this Sampler is shown in Figure~\ref{fig:task_sampling_low_diversity}.

% \begin{align}
% \label{ndb_sampler}
% \underset{batch\ 1}{\boxed{\underset{task\ 1}{\boxed{A, Z}}\quad \underset{task\ 2}{\boxed{B, F}}}}\quad \underset{batch\ 2}{\boxed{\underset{task\ 1}{\boxed{Z, A}}\quad \underset{task\ 2}{\boxed{B, F}}}}\quad
% ...\quad
% \underset{batch\ n}{\boxed{\underset{task\ 1}{\boxed{Z, A}}\quad \underset{task\ 2}{\boxed{F, B}}}}
% \end{align}

\begin{figure}[h]
    \centering
    \includegraphics[width=0.99\linewidth]{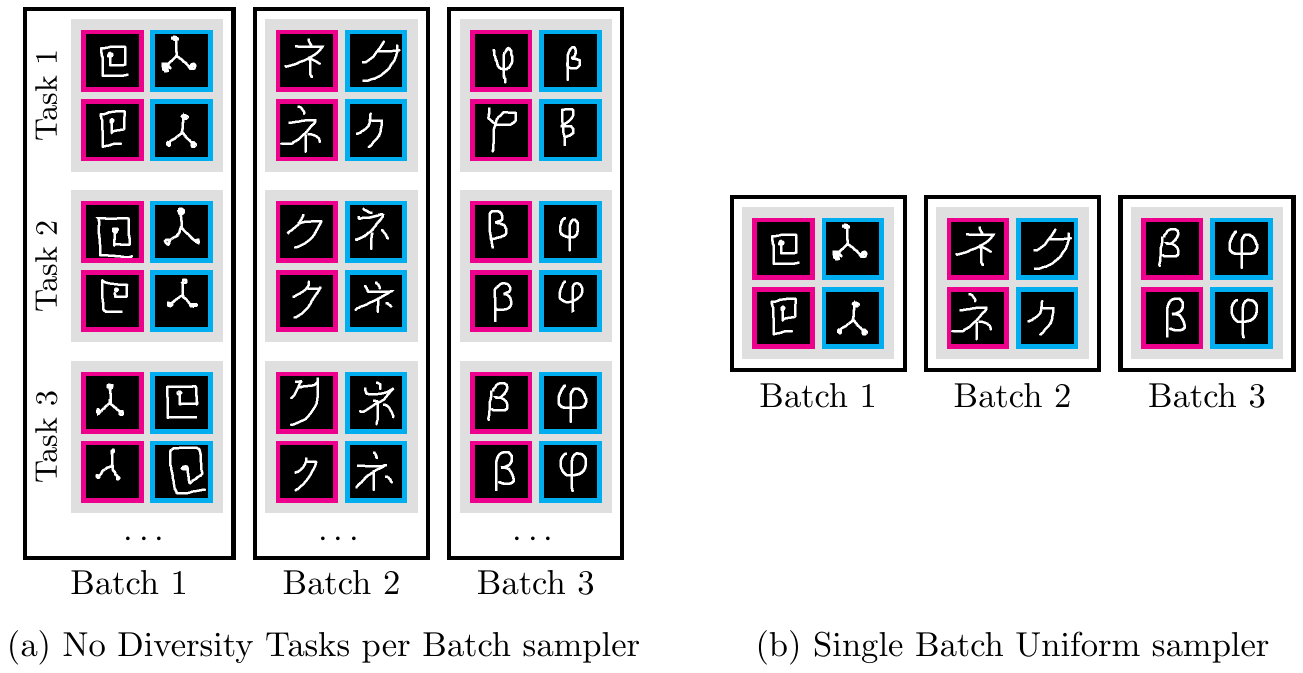}
    \caption{Illustration of (a) the No Diversity Task per Batch Sampler, and (b) the Single Batch Uniform Sampler.}
    \label{fig:task_sampling_single}
\end{figure}

\paragraph{\textbf{No Diversity Tasks per Batch Sampler}} In this setting, we uniformly sample one set of tasks for a given batch and propagate the same tasks for all meta-batches. We then repeat this same principle for sampling the next batch. Similar to the Samplers above, we also shuffle the labels to reduce overfitting. An illustration of this Sampler is shown in Figure~\ref{fig:task_sampling_single}.

% \begin{align}
% \label{ndtb_sampler}
% \underset{batch\ 1}{\boxed{\underset{task\ 1}{\boxed{D, T}}\quad \underset{task\ 2}{\boxed{T, D}}}}\quad \underset{batch\ 2}{\boxed{\underset{task\ 1}{\boxed{B, Y}}\quad \underset{task\ 2}{\boxed{Y, B}}}}\quad
% ...\quad
% \underset{batch\ n}{\boxed{\underset{task\ 1}{\boxed{K, R}}\quad \underset{task\ 2}{\boxed{R, K}}}}
% \end{align}

\paragraph{\textbf{Single Batch Uniform Sampler}} In this setting, we set the meta-batch size to one. This Sampler is intuitively the same as the No Diversity Task per Batch Sampler, without the repetition of tasks inside a meta-batch. This Sampler would be an ideal ablation study for the repetition of tasks in the meta-learning setting. An illustration of this Sampler is shown in Figure~\ref{fig:task_sampling_single}.

% \begin{align}
% \label{sbu_sampler}
% \underset{batch\ 1}{\boxed{\underset{task\ 1}{\boxed{D, T}}}}\quad \underset{batch\ 2}{\boxed{\underset{task\ 1}{\boxed{B, Y}}}}\quad
% ...\quad
% \underset{batch\ n}{\boxed{\underset{task\ 1}{\boxed{K, R}}}}
% \end{align}

\begin{figure}[h]
    \centering
    \begin{adjustbox}{center}
    \includegraphics[width=0.99\linewidth]{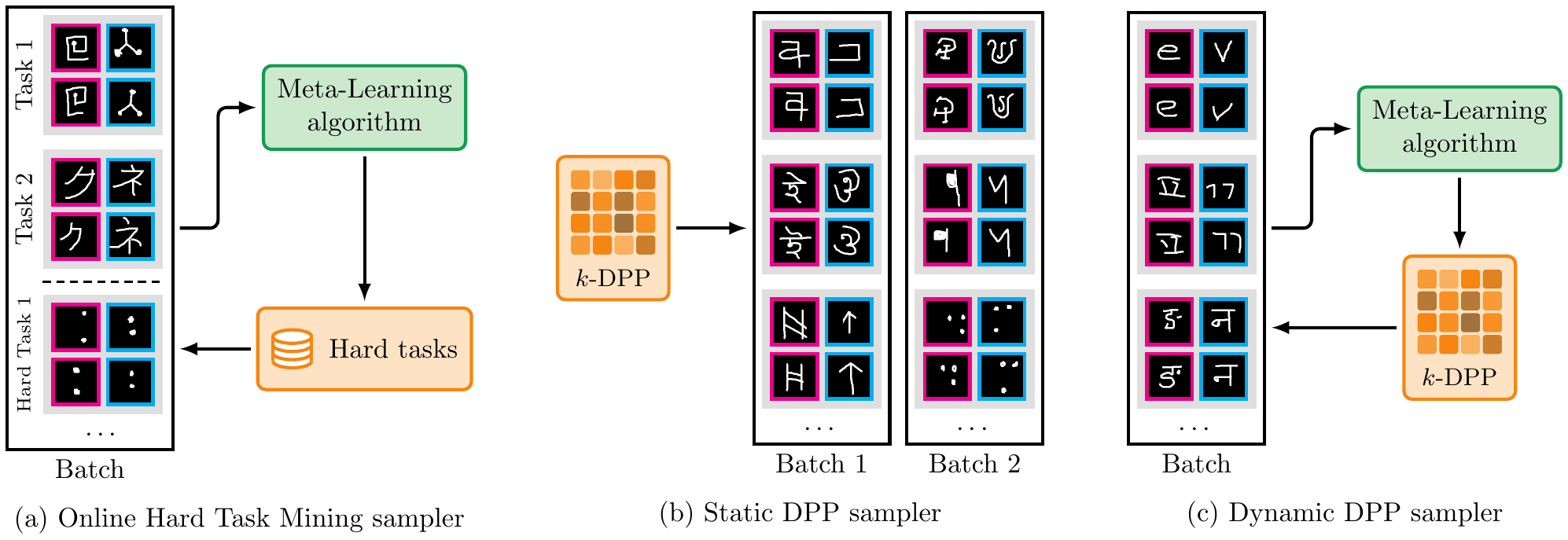}
    \end{adjustbox}
    \caption{Illustration of (a) Online Hard Task Mining Sampler, (b) the Static DPP Sampler, and (c) the Dynamic DPP Sampler.}
    \label{fig:task_sampling_high_diversity}
\end{figure}

\paragraph{\textbf{Online Hard Task Mining Sampler}} This setting is inspired by the works of \cite{shrivastava2016training}, where they proposed OHEM, which yielded significant boosts in detection performance on benchmarks like PASCAL VOC 2007 and 2012. OHEM sampler samples the hardest tasks from a pool of tasks previously seen. However, to reproduce OHEM for meta-learning, we only apply the OHEM sampler for half the meta-batch size and Uniform Sampler for the remaining half. This approach would involve many tasks and not restrict us to only known tasks. Furthermore, to avoid OHEM in the initial stages, we sample tasks with a uniform sampler until the buffer of tasks seen by the model becomes sufficiently big, say 50 tasks in our case. An illustration of this Sampler is shown in Figure~\ref{fig:task_sampling_high_diversity}.

% \begin{align}
% \label{ohtm_sampler}
% \underset{batch\ 1}{\boxed{\underset{task\ 1}{\boxed{D, T}}\quad \underset{task\ 2}{\boxed{K, R}}}}\quad \underset{batch\ 2}{\boxed{\underset{task\ 1}{\boxed{\mathit{OHEM}(\omega)}}\quad \underset{task\ 2}{\boxed{Y, B}}}}\quad
% ...\quad
% \underset{batch\ n}{\boxed{\underset{task\ 1}{\boxed{\mathit{OHEM}(\omega)}}\quad \underset{task\ 2}{\boxed{L, I}}}}
% \end{align}

\paragraph{\textbf{Static DPP Sampler}} Determinantal Point Processes (DPP) have been used for several machine learning problems \cite{kulesza2012determinantal}. They have also been used in other problems such as the active learning settings \cite{biyik2019batch} and mini-batch sampling problems \cite{zhang2019active}. These algorithms have also inspired other works in active learning in the batch mode setting \cite{ravi2018meta}. In this setting, we use DPP as an off-the-shelf implementation to sample tasks based on their class embeddings. These class embeddings are generated using our pre-trained Protonet model. The DPP instance is used to sample the most diverse tasks based on these embeddings and used for meta-learning. An illustration of this Sampler is shown in Figure~\ref{fig:task_sampling_high_diversity}.

% \begin{align}
% \label{sdpp_sampler}
% \underset{batch\ 1}{\boxed{\underset{task\ 1}{\boxed{\mathit{sDPP}(\omega')}}\quad \underset{task\ 2}{\boxed{\mathit{sDPP}(\omega')}}}}\quad \underset{batch\ 2}{\boxed{\underset{task\ 1}{\boxed{\mathit{sDPP}(\omega')}}\quad \underset{task\ 2}{\boxed{\mathit{sDPP}(\omega')}}}}\quad
% ...\quad
% \underset{batch\ n}{\boxed{\underset{task\ 1}{\boxed{\mathit{sDPP}(\omega')}}\quad \underset{task\ 2}{\boxed{\mathit{sDPP}(\omega')}}}}
% \end{align}

\paragraph{\textbf{Dynamic DPP Sampler}} In this setting, we extend the previous sDPP setting such that the model in training generates the class embeddings. The Sampler is motivated by the intuition that selecting the most diverse tasks for a given model will help learn better. Furthermore, to bootstrap the model in the initial stages of meta-training, we sample tasks with a uniform sampler until the model becomes sufficiently trained, say 500 batches in our case. An illustration of this Sampler is shown in Figure~\ref{fig:task_sampling_high_diversity}.

% \begin{align}
% \label{ddpp_sampler}
% \underset{batch\ 1}{\boxed{\underset{task\ 1}{\boxed{K, R}}\quad \underset{task\ 2}{\boxed{M, P}}}}\quad \underset{batch\ 2}{\boxed{\underset{task\ 1}{\boxed{\mathit{dDPP}(\omega^{0})}}\quad \underset{task\ 2}{\boxed{\mathit{dDPP}(\omega^{0})}}}}\quad
% ...\quad
% \underset{batch\ n}{\boxed{\underset{task\ 1}{\boxed{\mathit{dDPP}(\omega^{t})}}\quad \underset{task\ 2}{\boxed{\mathit{dDPP}(\omega^{t})}}}}
% \end{align}

\section{Study of Diversity}
\label{td}
\subsection{Preliminaries}

Before giving a more formal definition of task diversity, we set a few more fundamental ideas required to better understand our metric. In the domain of meta-learning, there have been no previous proposed definition of Task Diversity, and has remained highly heuristic and intuitive. Our definition could be used to serve as an established notion of ``Task Diversity'' to be used in future works. In this work, we consider the volume parallelopiped definition as discussed briefly below. Although simple, this definition is very intuitive to our concept of diversity in meta-learning. Our definition is highly robust and does consider diversity across various modalities such as classes, tasks, and batches. In this work, we compute the embedding from  a pre-trained protonet model. It would also be possible to compute these embeddings from another neural network approximation function, such as ResNet, VGG, etc., trained on ILSVRC as is commonly used to compare the difference between two images in the computer vision domain. Below we briefly introduce the proposed definition of ``Diversity'' in the meta-learning domain.

\paragraph{Task Diversity} We define task diversity as the diversity among classes within a task. This diversity is defined as the volume of parallelepiped spanned by the embeddings of each of these classes.
\begin{equation*}
    \mathcal{TD} \propto \left [\mathrm{vol}(\mathcal{T})  \right ]^2 
\end{equation*}
where $\mathcal{T}$ is defined as $\left \{ c_1,\ldots,c_N \right \}$, where $N$ is the number of ways, and $c_i$ is the feature embedding of the $i^{th}$ class. These feature embeddings are pre-computed using our pre-trained Protonet model, similar to the one used in sDPP. This value is analogous to the probability of selecting a task of the following classes.

\paragraph{Task Embedding} We define the task embedding as the mean embedding of class features within that task. The task embedding is computed such that:
\begin{equation*}
    \mathcal{TE} = \frac{1}{m}\sum_{i=0}^m c_i
\end{equation*}
where the task is defined as $\left \{ c_1,\ldots,c_N \right \}$, where $N$ is the number of ways, and $c_i$ is the embedding of the $i^{th}$ class. 

\paragraph{Batch Diversity} We define batch diversity as the diversity among tasks within a mini-batch. This diversity is defined as the volume of parallelepiped spanned by the task embeddings of each of these tasks within a mini-batch:
\begin{equation*}
    \mathcal{BD} \propto \left [\mathrm{vol}(\mathcal{B})  \right ]^2 
\end{equation*}
where $\mathcal{B}$ is defined as $\left \{ t_1,\ldots,t_m \right \}$, where $m$ is the number of tasks within a mini-batch, and $t_i$ is the feature embedding of the $i^{th}$ task, $\mathcal{TE}_i$. 

\paragraph{Batch Embedding} We define batch embeddings $\mathcal{BE}$ as the expected value of the embedding where the probability of each batch is proportional to the volume of the embeddings parallelepiped. This definition of probability is analogous to the one used in traditional DPPs.
\begin{equation*}
\begin{aligned}
\mathcal{BE} ={}& \mathcal{BD} \sum_{i} \pi(t_i) \mathcal{TE}_i
\end{aligned}
\end{equation*}
where $\pi(.)$ is the distribution derived from normalized task diversity $\mathcal{TD}$.
By definition, the batch embeddings $\mathcal{BE}$ have been defined such that the embedding is biased towards the most diverse samplers. To compute the overall diversity of our Sampler, we compute the volume of the parallelepiped spanned by the batch embeddings. However, we make a slight modification, such that the length of each batch embeddings is proportional to the average batch diversity, as defined earlier. This is useful when computing the volume, since we would like samplers that result in high batch diversity to encompass a higher volume.

Our process of computing the volume of parallelepiped spanned by the vectors is discussed in Appendix~\ref{vol}.

\subsection{Definition of Diversity}

We define the diversity of the Sampler as the volume of the parallelepiped spanned by the batch embeddings.
\begin{equation*}
    \mathcal{OD} \propto \left [\mathrm{vol}(\mathcal{BE})  \right ]^2.
\end{equation*}
With this definition, the volume spanned will be reduced if the Sampler has low diversity within a batch. Furthermore, the batch embeddings would be very similar if the model has low diversity across batches, thus reducing the practical volume spanned. 

With the following definition in place, we computed the average batch diversity across five batches, with a batch size of 8 with three different seeds. For samplers such as d-DPP and OHTM, we evaluate on the Protonet model since the embeddings would be similar and in the same latent space as those obtained from the other samplers which use the pre-trained Protonet model. Intuitively, $\mathcal{OD}$ measures the volume the embeddings cover in the latent space. The higher the value, the more volume has been covered in the latent space.

The average task diversity on the Omniglot dataset, scaled such that the Uniform Sampler has a diversity of 1, has been reported in Table~\ref{Results_std}. We confirm and show rigorously that our samplers can be broadly divided into three categories:
\begin{itemize}
    \item \textbf{Low-Diversity Task Samplers:} These samplers include those with an overall diversity score less than 1. These include NDT, NDB, NDTB, and SBU Samplers.
    \item \textbf{Standard Sampler:} This serves as our baseline and is the standard Sampler used in the community - the Uniform Sampler.
    \item \textbf{High-Diversity Task Samplers:} These samplers include those with an overall diversity score greater than 1. These include OHTM, sDPP, and dDPP Samplers.
\end{itemize}

Furthermore, Our approach does have its advantages over other trivial alternatives such as pairwise-distance metrics. Our proposed formulation is agnostic of the batch size. This property is much desired in the meta-learning setting since the meta-training objectives also work with batch averages. Furthermore, our proposed formulation is more computationally efficient in terms of both time and space than other simpler alternatives. The formulation also offers modularity in its approach, and we can study the diversity at each level, be it tasks, meta-batches, or batches, something not possible with other metrics such as pairwise-distance metrics. 

\begin{table}[t]
\caption{Overall Diversity of Task Samplers.}
\label{Results_std}
\vskip 0.15in
\begin{center}
\begin{small}
\begin{sc}
\begin{adjustbox}{width=0.5\linewidth,center}
\begin{tabular}{lcr}
\toprule
\textbf{Sampler} & \textit{Diversity}\\
\midrule
No Diversity Task Sampler        & \textcolor{red}{$0.00$}\\
No Diversity Batch Sampler        & \textcolor{red}{$0.00$}\\
Single Batch Uniform Sampler            & \textcolor{red}{$0.00$} \\
No Diversity Tasks per Batch Sampler            & \textcolor{red}{$\approx 0.00$}\\ \midrule 
Uniform Sampler              & 1.00 \\ \midrule
OHTM Sampler            & \textcolor{darkgreen}{$1.69$} \\
d-DPP Sampler           & \textcolor{darkgreen}{$12.40$} \\
s-DPP Sampler            & \textcolor{darkgreen}{$12.86$} \\
\bottomrule
\end{tabular}
\end{adjustbox}
\end{sc}
\end{small}
\end{center}
\vskip -0.1in
\end{table}

\section{Experiments}
\label{exp}
The experiment aims to answer the following questions: (a) How does limiting task diversity affect meta-learning? (b) Do sophisticated samplers such as OHEM or DPP  that improve diversity offer any significant boost in performance? (c) What does our finding imply about the efficacy of the current meta-learning models?

To make an exhaustive study on the effect of task diversity in meta-learning, we train on four datasets: Omniglot \cite{lake2011one}, \textit{mini}Imagenet \cite{ravi2016optimization}, \textit{tiered}ImageNet \cite{ren2018meta}, and Meta-Dataset \cite{triantafillou2019meta}. With this selection of datasets, we cover both simple datasets, such as Omniglot and \textit{mini}ImageNet, as well as the most difficult ones, such as \textit{tiered}ImageNet and Meta-Dataset. We train three broad classes of meta-learning models on these datasets: Metric-based (i.e., Protonet \citep{snell2017prototypical}, Matching Networks \citep{vinyals2016matching}), Optimization-based (i.e., MAML \citep{finn2017model}, Reptile \citep{nichol2018first}, and MetaOptNet \citep{lee2019meta}), and Bayesian meta-learning models (i.e., CNAPs \citep{requeima2019fast}). More details about the datasets which were used in our experiments are discussed in Appendix~\ref{datasets}. More details about the models and their hyperparameters are discussed in Appendix~\ref{models}. We created a common pool of 1024 randomly sampled held-out tasks to test every algorithm in our experiments to make an accurate comparison. We assessed the statistical significance of our results for all experiments based on a paired-difference t-test, with a p-value $p=0.05$.

% The code and implementation of all our experiments are publicly available at \url{https://github.com/RamnathKumar181/Task-Diversity-meta-learning}.

\subsection{Results}

\begin{figure}[t]
    \centering
    \includegraphics[width=0.99\linewidth]{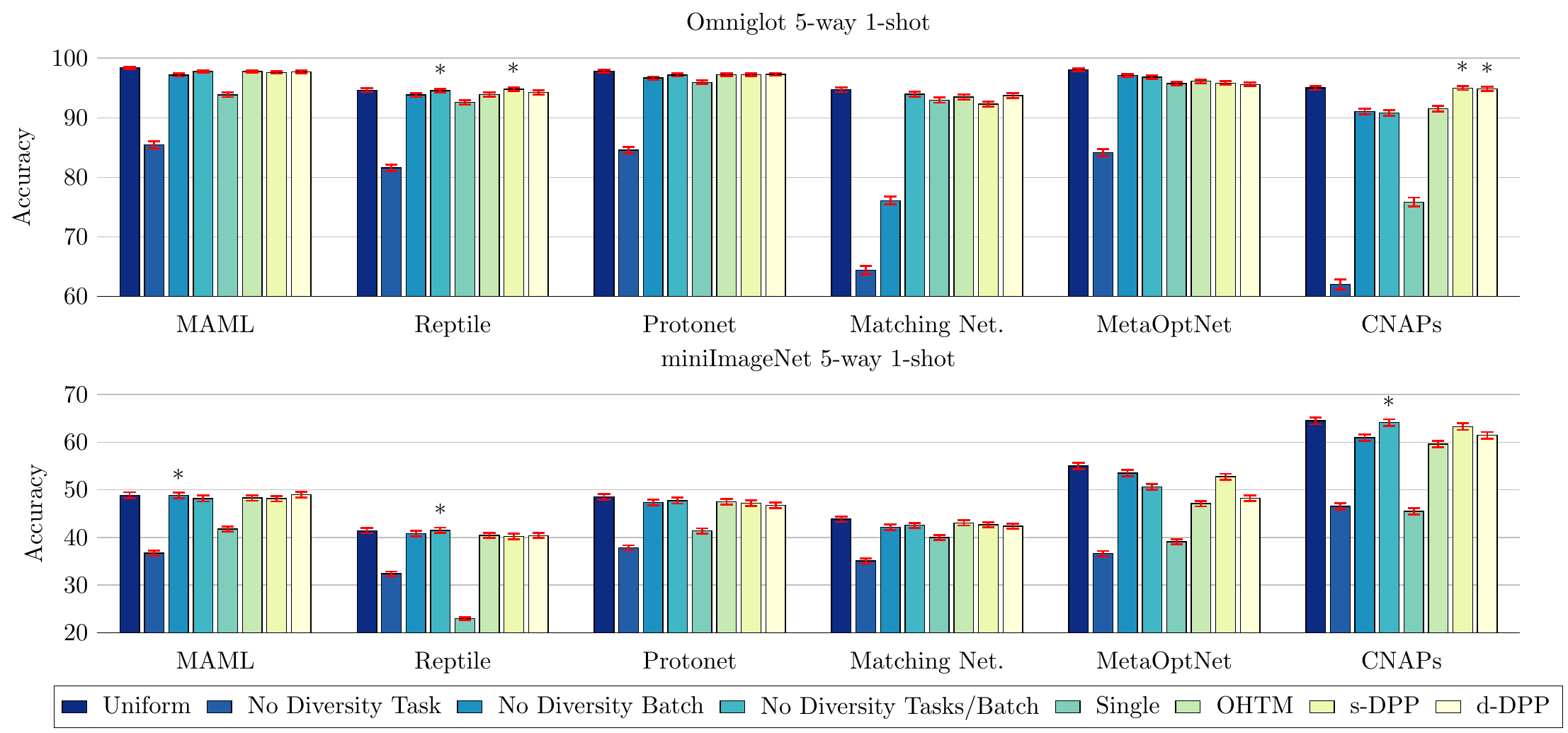}
    \caption{Average accuracy on Omniglot 5-way 1-shot \& \textit{mini}ImageNet 5-way 1-shot, with 95\% confidence interval. All samplers are poorer than the Uniform Sampler and are statistically significant (with a p-value $p=0.05$). We use the symbol $*$ to represent the instances where the results are not statistically significant and similar to the performance achieved by Uniform Sampler.}
    \label{fig:results_omniglot_miniimagenet}
\end{figure}

\begin{figure}[t]
    \centering
    \includegraphics[width=0.99\linewidth]{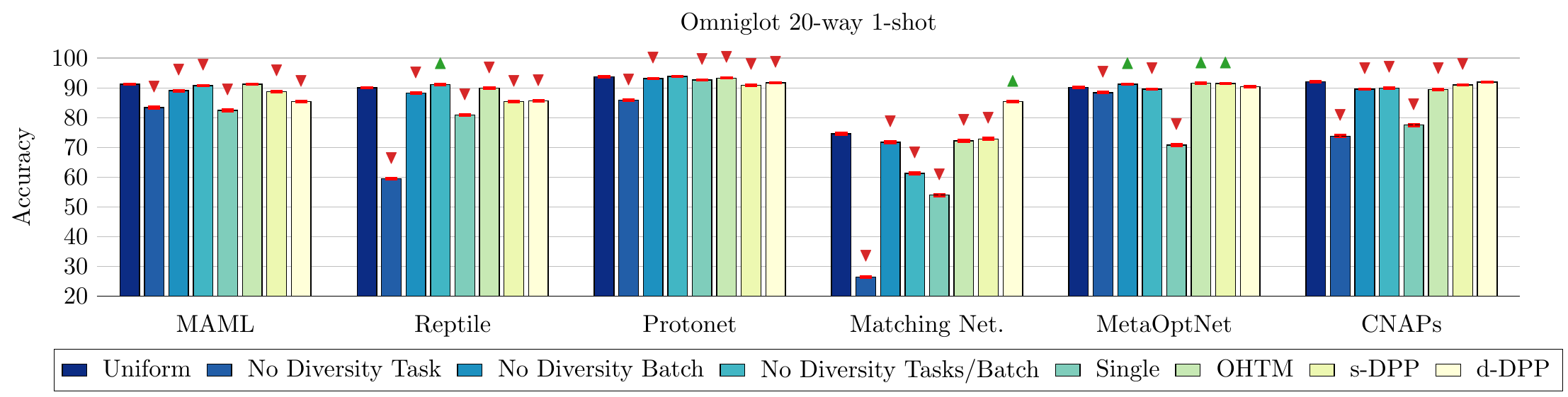}
    \caption{Average accuracy on Omniglot 20-way 1-shot, with a 95\% confidence interval. We denote all samplers that are worse than the Uniform Sampler and are statistically significant (with a p-value $p=0.05$) with \textcolor{tableau10_C3}{$\blacktriangledown$}, and those that are significantly better than the Uniform Sampler with \textcolor{tableau10_C2}{$\blacktriangle$}.}
    \label{fig:results_omniglot_20}
\end{figure}

% \begin{figure}[t]
%     \centering
%     \includegraphics[width=\linewidth]{figures/results_tiered.pdf}
%     \caption{Average accuracy on \textit{tiered}ImageNet 5-way 1-shot, with a 95\% confidence interval. We denote all samplers that are worse than the Uniform Sampler and are statistically significant (with a p-value $p=0.05$) with \textcolor{tableau10_C3}{$\blacktriangledown$}, and those that are significantly better than the Uniform Sampler with \textcolor{tableau10_C2}{$\blacktriangle$}.}
%     \label{fig:results_tiered}
% \end{figure}

\begin{figure}[t]
    \centering
    \includegraphics[width=0.99\linewidth]{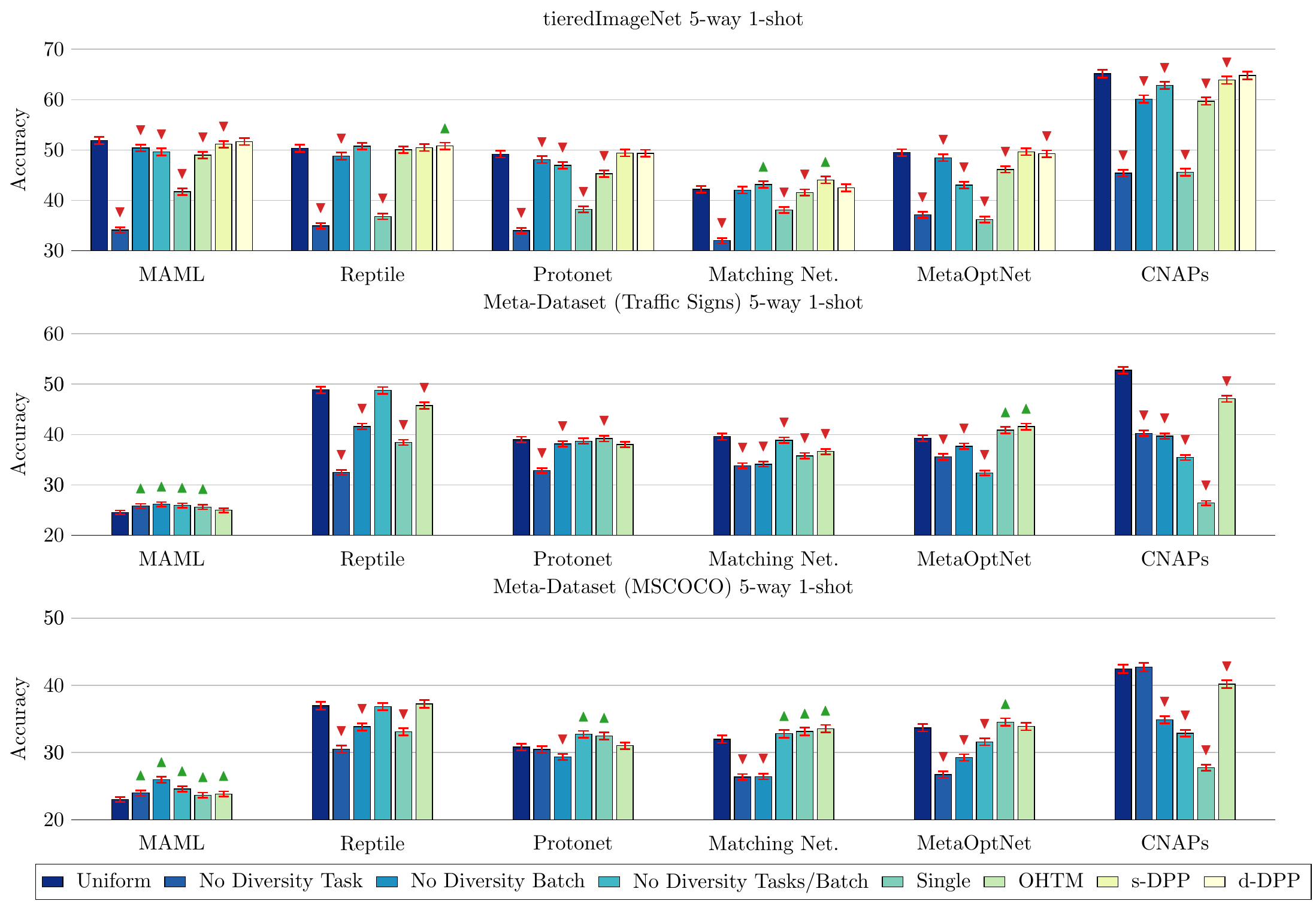}
    \caption{Average accuracy on \textit{tiered}ImageNet 5-way 1-shot, Meta-Dataset Traffic Sign 5-way 1-shot \& Meta-Dataset MSCOCO 5-way 1-shot, with a 95\% confidence interval. We denote all samplers that are worse than the Uniform Sampler and are statistically significant (with a p-value $p=0.05$) with \textcolor{tableau10_C3}{$\blacktriangledown$}, and those that are significantly better than the Uniform Sampler with \textcolor{tableau10_C2}{$\blacktriangle$}.}
    \label{fig:results_tiered_meta}
\end{figure}

In this section, we present the results of our experiments. Figure~\ref{fig:results_omniglot_miniimagenet} presents the performance of the six models on the Omniglot and \textit{mini}ImageNet under different task samplers in the 5-way 1-shot setting. Table~\ref{Results_5} in the Appendix presents the same results with higher precision.

We also reproduce our experiments on the 20-way 1-shot setting on the Omniglot dataset to establish that these trends are shared across different settings. Figure~\ref{fig:results_omniglot_20} presents our performance of the models under this setting. Furthermore, the results on the 20-way 1-shot experiments are presented in Table~\ref{Results_20} in the Appendix with higher precision. We also extend the same to the meta-regression setting and observe similar trends as further discussed in Appendix~\ref{add_results}.
To further establish our findings, we also present our results on notoriously harder datasets such as \textit{tiered}ImageNet and Meta-Dataset. Figure~\ref{fig:results_tiered_meta} presents the performance of the models on the \textit{tiered}ImageNet. Again, Table~\ref{Results_5} in the Appendix presents the same results with higher precision. 

Figure~\ref{fig:results_tiered_meta} presents the performance of the models on the Meta-Dataset Traffic Sign and Meta-Dataset MSCOCO datasets. We only present the results on Traffic Sign and MSCOCO of the Meta-Dataset, as these two sub-datasets are exclusively used for testing and accurately represent the generalization power of the models when trained with different levels of task diversity.  Other results on the Meta-Dataset are presented in Table~\ref{Results_meta}. We empirically show that task diversity does not lead to any significant boost in the performance of the models. In the subsequent section, we discuss some of the other key findings from our work.

\section{Discussion}
\label{disc}

From our experiments, we show a similar trend on easy meta-classification tasks (Omniglot and \textit{mini}ImageNet as depicted in Figure~\ref{fig:results_omniglot_miniimagenet}), as well as harder tasks (\textit{tiered}ImageNet and Meta-Dataset as depicted in Figure~\ref{fig:results_tiered_meta}) in the 5-way 1-shot setting. We also extended our study to the 20-way 1-shot setting with the Omniglot dataset (Figure~\ref{fig:results_omniglot_20}). To test the effect of diversity when the number of shots increases, we turn to the meta-regression domain as depicted in Table~\ref{Results_Regression_5}. Furthermore, to study the effect of diversity in the OOD setting, we turn back to our results on Traffic Sign and MSCOCO datasets from Meta-Dataset (Figure~\ref{fig:results_tiered_meta}). Across all our experiments, we notice a general trend, and we discuss this briefly below.

\paragraph{Disparity between Single Batch Uniform and NDTB Sampler} An exciting result is the Disparity between Single Batch Uniform Sampler and No Diversity Tasks per Batch Sampler. The only difference between the two samplers is that tasks are repeated in the latter. However, this repetition seems to offer a great deal of information to the model and allows the model to perform on par with the Uniform Sampler. One might hypothesize that the Single Batch Uniform Sampler obtains the performance observed by the No Diversity Tasks per Batch Sampler if trained for enough epochs. This scenario has been considered and refuted by our experiments in Appendix~\ref{add_results}.

\paragraph{Difference between ``Task Difficulty'' and ``Task Diversity''} Prior works have studied the effects of task difficulty on the performance of the model. Classifying diverse classes would be easier for metric-based networks and harder for optimization-based networks (while testing, it might be difficult to reach very different latent spaces after the inner loop optimization). Thus, the concept of diversity and its connection to the difficulty of the tasks becomes model-dependent and not suitable as a robust metric for analogous understanding. It is important that throughout this work, we do not use the concept of difficulty as a definition or analogy for diversity. Furthermore, although we notice OHTM sampler to perform sparingly well in some cases, note that these models are pre-dominantly metric-based methods. Thus, the OHTM sampler in this scenario would be sampling less diverse tasks and is in coherence with our overall result.

\paragraph{Comparison between NDTB, NDB, and Uniform Sampler} From our experiments, we also notice that the No Diversity Tasks per Batch Sampler and No Diversity Batch Sampler are pretty similar to the Uniform Sampler in terms of performance. This observation would suggest that the model trained on only a data fragment can perform similarly to that trained on the Uniform Sampler. This phenomenon questions the improvement/addition the additional amount of data has brought.

\paragraph{\textbf{Declining performance in d-DPP Methodology}} The performance may degrade over epochs for d-DPP due to the non-stationarity of the sampling process (the DPP gets updated along with the model). This effect may be evident for metric-based methods (matching nets, protonet) since class embeddings directly impact the sampling process through the DPP and the model's performance. One notable exception is MetaOptNet, which explains our highlight in the paper. This behavior hypothesizes that the SVM may be more robust to changes in the embeddings (induced by this non-stationary process) due to the max-margin classifier. We present the convergence graph of the MetaOptNet model on Omniglot 5-way 1-shot run in Figure~\ref{fig:metaoptnet} in the Appendix with an added smoothing factor of 1.

\paragraph{\textbf{Theoretical Analysis}} Neural networks in theory are capable of mapping any function given the width of the network is sufficiently large. However, in practice two scenarios could occur derailing the network to a sub-optimal solution: (i) The network is shallow/small and not expressive enough for the optimal solution or (ii) model is expressive enough, but SGD is not able to find the global optima, either due to saddle points, low learning rate etc. Under the assumption that we have a well-defined model, we can intuitively understand why increasing diversity does not help the model better. When the data points are close to each other, the learning of features from one could easily transfer to the other points and achieve a good fit. The diverse data distribution might not be as straightforward since the model would have to learn multiple disjoint concepts to classify these points. This is the crux of Simpson's Paradox. This visualization would be easier to understand in a generic regression setting. We expand on our theoretical analysis further in Appendix~\ref{theory}.

% \paragraph{General Trend} From our experiments, we notice that there are generally two classes of samplers: High Performing Samplers and Low Performing Samplers. The High Performing Samplers include No Diversity Batch, No Diversity Tasks per Batch, Uniform, OHTM, and s-DPP Sampler. The Low Performing Samplers include No Diversity Task, Single Batch Uniform, and d-DPP Sampler. This trend is shared across all datasets and models. There are some perturbations in ranking within the two classes, but the High Performing Samplers tend to perform better than the Low Performing Samplers by their very definition.

\section{Related Works}

Meta-learning formulations typically rely on episodic training, wherein an algorithm adapts to a task, given its support set, to minimize the loss incurred on the query set. Meta-learning methods differ in terms of the algorithms they learn, and can be broadly classified under four prominent classes: \textit{Metric-based}, \textit{Model-based}, \textit{Optimization-based} and \textit{Bayesian-based} approaches. A more detailed overview of these methods is discussed in Appendix~\ref{metalearning_method}.

Although research in meta-learning has attracted much attention recently, the effect of task diversity in the domain of meta-learning is still an open question. However, task sampling and task diversity have been more extensively studied in other closely related problems such as active learning. Active learning involves selecting unlabeled data items to improve an existing classifier. Although most of the approaches in this domain are based on heuristics, there are few approaches to sample a batch of samples for active learning. \cite{ravi2018meta} proposed an approach to sample a batch of samples using a Prototypical Network \citep{snell2017prototypical} as the backbone architecture. The model maximizes the query set, given support set and unlabeled data. Other works such as CACTUs \cite{hsu2018unsupervised}, proposed a framework that samples tasks/examples using relatively simple task construction mechanisms such as clustering embeddings. The unsupervised representations learned via these samples perform well on various downstream human-specified tasks.

Although nascent, a few recent works aim to improve meta-learning by explicitly looking at the task structure and relationships. Among these, \cite{yin2019meta} proposed an approach to handle the lack of mutual exclusiveness among different tasks through an information-theoretic regularized objective.  In addition, several popular meta-learning methods \cite{lee2019meta, snell2017prototypical} improve the meta-test performance by changing the number of ways or shots of the sampled meta-training tasks, thus increasing the complexity and diversity of the tasks. \cite{liu2020adaptive} proposed an approach to sample classes using class-pair-based sampling and class-based sampling. The Class-pair based Sampler selects pairs of classes that confuse the model the most, and the Class-based Sampler samples each class independently and does not consider the difficulty of a task as a whole. Our OHTM sampler is similar to the Class-pair based Sampler. \cite{liu2020task} propose to augment the set of possible tasks by augmenting the pre-defined set of classes that generate the tasks with varying degrees of rotated inputs as new classes. Closer to our work, \cite{setlur2020support} study a specific sampler by limiting the pool of tasks. Our work, however, has remained fundamentally different, and we expand on this briefly in Appendix~\ref{add_results}. To the best of our knowledge, our work is the first to study the full range of the effect of task diversity in meta-learning.

\section{Conclusions and Future Work}

In this paper, we have studied the effect of task diversity in meta-learning. We have empirically shown task diversity's effects in the meta-learning domain. We notice two important findings from our research: (i) Limiting task diversity and repeating the same tasks over the training phase allows us to obtain similar performances to the Uniform Sampler without any significant adverse effects. Our experiments using the NDTB and NDB empirically show that a model trained on even a tiny data fragment can perform similarly to a model trained using Uniform Sampler. This is a crucial finding since this questions the need to increase the support set pool to improve the models' performance. (ii) We also show that sophisticated samplers such as OHTM or DPP samplers do not offer any significant boost in performance. In contradiction, we notice that increasing task diversity using the d-DPP Sampler hampers the performance of the meta-learning model.  

We believe that the experiments and task diversity definition we performed and defined lay the roadwork to further research on the effect of task diversity domain in meta-learning and encourage more in-depth studies into the efficacy of our meta-learning methods.

\section*{Reproducibility Statement}

In this paper, we work with four different datasets - Omniglot, \textit{mini}ImageNet, \textit{tiered}ImageNet and Meta-Dataset. Additional details about setting up these datasets is available in Appendix \ref{datasets}. Furthermore, we experimented with six models - MAML, Reptile, Protonet, Matching Networks, MetaOptNet, and CNAPs. All these models were run after reproducing from their open-source codes. Additional details about setting up these models are available in Appendix \ref{models}. 
Our source code is made available for additional reference \footnote{\url{https://github.com/RamnathKumar181/Task-Diversity-meta-learning}}.

\section*{Ethical Statement}
Our work studies the effect of task diversity in the meta-learning setting. It helps us understand the efficacy of our models and better design samplers in the future. To the best of our knowledge, this work poses no negative impacts on society.

\section*{Acknowledgments}
We would like to thank Sony Corporation for funding this research through the Sony Research Award Program. We would also like to thank Dheeraj Mysore Nagaraj from Google Research, India, for his valuable discussions and ideas, which led to the conclusions presented in the theoretical analysis section.

\bibliography{main.bib}
\bibliographystyle{plain}

\newpage

\appendix

\section*{Appendix}

\section{Datasets}
\label{datasets}

Omniglot, \textit{mini}ImageNet and \textit{tiered}ImageNet were extracted using pytorch-meta. The Meta-Dataset was downloaded using the setup information from the official repository: \url{https://github.com/google-research/meta-dataset}. 

\paragraph{\textbf{Omniglot}} Omniglot is a benchmark dataset proposed by \cite{lake2011one} for few-shot image classification tasks. Omniglot dataset consists of 20 instances and 1623 characters from 50 different alphabets. We experiment with both 5-way 1-shot and 2-way 1-shot in this work.

\paragraph{\textit{mini}ImageNet} \textit{mini}ImageNet is another benchmark dataset proposed by \cite{ravi2016optimization} for few-shot image classification tasks. The \textit{mini}ImageNet dataset involves 64 training classes, 12 validation classes, and 24 test classes. We run under the setting 5-way 1-shot for this experiment.

\paragraph{\textit{tiered}ImageNet} \textit{tiered}ImageNet is notorious as a difficult benchmark dataset proposed by \cite{ren2018meta}. The \textit{tiered}ImageNet dataset is a larger subset of ILSVRC-12 with 608 classes (779,165 images) grouped into 34 higher-level nodes in the ImageNet human-curated hierarchy. This set of nodes is partitioned into 20, 6, and 8 disjoint sets of training, validation, and testing nodes, and the corresponding classes form the respective meta-sets. As argued in \cite{ren2018meta}, this split near the root of the ImageNet hierarchy results in a more challenging yet realistic regime with test classes that are less similar to training classes. We run under the setting 5-way 1-shot for this experiment.

\paragraph{\textbf{Meta-Dataset}} Meta-Dataset is notorious as a difficult benchmark dataset proposed by \cite{triantafillou2019meta}. The Meta-Dataset is much larger than any previous benchmark and comprises multiple existing datasets. This invites research into how diverse data sources can be exploited by a meta-learner and allows us to evaluate a more challenging generalization problem to new datasets. Specifically, Meta-Dataset leverages data from the following 10 datasets: ILSVRC-2012 (\cite{russakovsky2015imagenet}), Omniglot (\cite{lake2011one}), Aircraft (\cite{maji2013fine}), CUB-200-2011 (\cite{wah2011caltech}), Describable Textures (\cite{cimpoi2014describing}), Quick Draw (\cite{jongejan2016quick}), Fungi (\cite{schroeder2018fgvcx}), VGG Flower (\cite{nilsback2008automated}), Traffic Signs (\cite{houben2013detection}) and MSCOCO (\cite{lin2014microsoft}). There exist few classes with fewer image samples than 16. This becomes an issue since we need 1 for training and 15 for testing. We repeat some of the support images for these classes to make up for the lack of examples. Since the number of such classes is minimal, we justify this use as this solution cannot significantly increase the model's performance.
Furthermore, unlike previous experiments on this dataset which use a variable number of ways and shots during training, we train with a fixed number of ways and shots. Furthermore, unlike the works in \cite{triantafillou2019meta}, we do not perform any pre-training to help aid the model. We believe these are the main attributions for the disparity in our performance. However, since we are focused on the relative performance of the samplers for a given model, this discrepancy would not affect our study of task diversity. We again run under the setting 5-way 1-shot for this experiment.

\section{Models}
\label{models}
This section describes some of the models we used for our experiments and the hyperparameters used for their training. Before we get into the exact models, we describe the different classes of meta-learning models and the algorithms that come under these classes.

\subsection{Overview of meta-learning models}
\label{metalearning_method}

\textit{Metric-based methods} such as \cite{koch2015siamese, vinyals2016matching, snell2017prototypical, sung2018learning} operate on the core idea similar to nearest neighbors algorithm and kernel density estimation. These methods are also called non-parametric approaches. \textit{Model-based methods} such as \cite{santoro2016meta, munkhdalai2017meta} depend on a model designed specifically for fast learning, which updates its parameters rapidly with a few training steps, achieved by its internal architecture or controlled by another meta-learned model. Generic deep learning models learn through backpropagation of gradients, which are neither designed to cope with a small number of training samples nor converge within a few optimization steps. To address this, \textit{Optimization-based methods} such as \cite{ravi2016optimization, finn2017model, nichol2018first} were proposed, which were better suited to learn from a small number of samples. However, all the above approaches are deterministic and are not the most suited for few-shot problems that are generally ambiguous. Hence, \textit{Bayesian-based methods} such as \cite{yoon2018bayesian, requeima2019fast} were proposed which helped address the above issue.

\subsection{Experimental Setup}

\subsubsection{Prototypical Networks} Prototypical Networks proposed by \cite{snell2017prototypical} constructs a prototype for each class and then classifies each query example as the class whose prototype is 'nearest' to it under Euclidean distance. More concretely, the probability that a query example $x^*$ belongs to class $k$ is defined as:
\begin{equation}
    p(y^*=k|x^*,\mathcal{S}) = \frac{\exp(-\left \| g(x^*)-c_k \right \|_2^2)}{\sum_{k'\in \left \{ 1,...,N \right \}}\exp(-\left \| g(x^*)-c_{k'} \right \|_2^2)}
\end{equation}
Where $c_k$ is the 'prototype' for class $k$: the average embeddings of class $k$'s support examples. 

\subparagraph{Hyperparameters}
In our experiments on Omniglot and \textit{mini}ImageNet, and \textit{tiered}ImageNet under a 5-way, 1-shot setting, we run the model for 100 epochs with a batch size of 32 and a meta-learning rate of 0.001. We use an Adam optimizer to make gradient steps and a StepLR scheduler with step size 0.4 and gamma 0.5. The same hyperparameters are used for training our model on Omniglot under a 20-way 1-shot setting. 

We use the same parameters as the \textit{mini}ImageNet to train our model on the Meta-Dataset under a 5-way 1-shot setting. However, we run with a batch size of 16 rather than 32 to accommodate the extensive training period and memory constraints.

\subsubsection{Matching Networks} Matching Networks proposed by \cite{vinyals2016matching} labels each query example as a cosine distance-weighted linear combination of the support labels:
\begin{equation}
    p(y^*=k|x^*,\mathcal{S}) = \sum _{i=1}^{|\mathcal{S}|}\alpha(x^*, x_i)\Phi  _{y_i=k},
\end{equation}
where $\Phi_A$ is the indicator function and $\alpha (x^*, x_i)$ is the cosine similarity between $g(x^*)$ and $g(x_i)$, softmax normalized over all support examples $x_i$, where $1\leq i \leq |\mathcal{S}|$. 

We had trouble reproducing the results from matching networks using cosine distance since the convergence seemed to be slow and the final performance dependent on the random initialization. 
This is similar to what is observed by other repositories such as \url{https://github.com/oscarknagg/few-shot}. Since we are focused on the relative performance of the samplers for a given model, this discrepancy would not affect our study of task diversity.

\subparagraph{Hyperparameters}
In our experiments on Omniglot, \textit{mini}ImageNet, and \textit{tiered}ImageNet under a 5-way, 1-shot setting, we run the model for 100 epochs with a batch size of 32 and an Adam optimizer with a meta-learning rate of 0.001 and a weight decay of 0.0001. The same hyperparameters are used for training our model on Omniglot under a 20-way 1-shot setting.

We use the same parameters as the \textit{mini}ImageNet to train our model on the Meta-Dataset under a 5-way 1-shot setting. However, we run with a batch size of 16 rather than 32 to accommodate the extensive training period and memory constraints.

\subsubsection{MAML} MAML proposed by \cite{finn2017model} uses a linear layer parametrized by \textbf{W} and \textbf{b} on top of the embedding function $g(.;\theta)$ and classifies a query example as:
\begin{equation}
    p(y^*|x^*,\mathcal{S}) = \mathit{softmax}(\mathbf{b}'+\mathbf{W}'g(x^*;\theta ')) ,
\end{equation}
where the output layer parameters \textbf{W'} and \textbf{b'} and the embedding function parameters $\theta '$ are obtained by performing a small number of within-episode training steps on the support set $\mathcal{S}$, starting from initial parameter values $(\mathbf{b}, \mathbf{W},\theta)$. 

\subparagraph{Hyperparameters}
In our experiments on Omniglot, \textit{mini}ImageNet, and \textit{tiered}ImageNet under a 5-way, 1-shot setting, we run the model for 150 epochs with a batch size of 32, with the Adam optimizer with a meta-learning rate of 0.001, number of inner adaptations as 1, and step size 0.4. For our experiments on Omniglot under the 20-way 1-shot setting, we set the step size of 0.1 and the number of inner adaptations to 5, batch size of 16, and kept all other hyperparameters constant.

We use the same parameters as the \textit{mini}ImageNet to train our model on the Meta-Dataset under a 5-way 1-shot setting. However, we ran with a batch size of 16 rather than 32 and only trained for 100 epochs to accommodate the extensive training period and memory constraints.

\subsubsection{Reptile} Like MAML, Reptile proposed by \cite{nichol2018first} learns an initialization for the parameters of a neural network model, such that when we optimize these parameters at test time, learning is fast - i.e., the model generalizes from a small number of test tasks. Reptile converges towards a solution $\phi$ that is close (in Euclidean distance) to each task $\tau$'s manifold of optimal solutions. Let $\phi$ denote the network initialization, and $W = \phi + \Delta \phi$ denote the network weights after performing some sort of update. Let $\mathcal{W}_{\tau}^*$ denote the set of optimal network weights for task $\tau$. We want to find $\phi$ such that the distance $D(\phi, \mathcal{W}_{\tau}^*)$ is small for all tasks:
\begin{equation}
    \min_{\phi}\mathbb{E}_{\tau}[\frac{1}{2}D(\phi , \mathcal{W}_{\tau}^*)^2]
\end{equation}

The official repository seems to train the model with a 5-way 15-shot and test the model on a 5-way 1-shot. However, we do not consider this to be an accurate study of the effect of task diversity. In our work, we train and test the model in a 5-way 1-shot setting to ensure a fair and accurate comparison with other models. We believe this to be the source of discrepancy in our performance scores. Since we are focused on the relative performance of the samplers for a given model, this discrepancy would not affect our study of task diversity.

\subparagraph{Hyperparameters}
In our experiments on Omniglot and \textit{mini}ImageNet under a 5-way, 1-shot setting, we run the model for 150 epochs with a batch size of 32, a learning rate of 0.01, a meta-learning rate of 0.001, and a number of inner adaptations as 5. We use the SGD optimizer of inner steps and Adam optimizer for the outer steps. For our experiments on \textit{tiered}ImageNet, we change the number of inner adaptations to 10, keeping all other hyperparameters constant. For our experiments on Omniglot under the 20-way 1-shot setting, we set the meta-learning rate to 0.0005 and the number of inner adaptations to 10 and kept all other hyperparameters constant. Furthermore, we only run the model for 50 epochs due to the very high training time.

We use the same parameters as the \textit{tiered}ImageNet to train our model on the Meta-Dataset under a 5-way 1-shot setting. However, we run with a batch size of 16 rather than 32 to accommodate the extensive training period and memory constraints.

\subsubsection{CNAPs} Conditional Neural Adaptive Processes proposed by \cite{requeima2019fast} is able to efficiently solve new multi-class classification problems after an initial training phase. The proposed approach, based on Conditional Neural Processes (CNPs) mentioned in \cite{garnelo2018conditional}, adapts a small number of task-specific parameters for each new task encountered at test time. These parameters are conditioned on a set of training examples for the new task. They do not require additional tuning to adapt both the final classification layer and feature extraction process. This allows the model to handle various input distributions. The CNPs construct predictive distributions given $x^*$ as:
\begin{equation}
    p(y^*|x^*,\theta, D^{\tau}) = p(y^*|x^*,\theta, \Psi^{\tau} = \Psi_{\phi}(D^{\tau})),
\end{equation}
where $\theta$ are global classifier parameters shared across tasks, $\Psi^{\tau}$ are local task-specific parameters, produced by a function $\Psi_{\phi}(.)$ that acts of $D^{\tau}$. $\Psi_{\phi}(.)$ has another set of global parameters $\phi$ called \textit{adaptation network parameters}. $\theta$ and $\phi$ are the learnable parameters in the model. 

\subparagraph{Hyperparameters}
In all our experiments with CNAPs, we run the model for ten epochs with a batch size of 16 and a meta-learning rate of 0.01.

\subsubsection{MetaOptNet} MetaOptNet proposed by \cite{lee2019meta} proposes a linear classifier as the base learner for a meta-learning based approach for few-shot learning. The approach uses a linear support vector machine (SVM) to learn a classifier given a set of labeled training examples. The generalization error is computed on a novel set of examples from the same task. The objective is to learn an embedding model $\phi$ that minimizes generalization (or test) error across tasks given a base learner $\mathcal{A}$. Formally, the learning objective is:
\begin{equation}
    \min_{\phi}\mathbb{E}_{\mathcal{T}}[\mathcal{L}^{\mathit{meta}}(\mathcal{D}^{\mathit{test}};\theta, \phi), \textsf{ where } \theta = \mathcal{A}(\mathcal{D}^{\mathit{test}};\phi)].
\end{equation}
The choice of base learner $\mathcal{A}$ has a significant impact on the above equation. The base learner that computes $\theta = \mathcal{A}(\mathcal{D}^{\mathit{test}};\phi)$ has to be efficient since the expectation has to be computed over a distribution of tasks. This work considers base learners based on multi-class linear classifiers such as SVM, where the base learner's object is convex. Thus, the base learner can be simplified as follows:
\begin{equation}
  \begin{array}{l}
     \theta = \mathcal{A}(\mathcal{D}^{\mathit{test}};\phi) = \underset{\{\mathbf{w}_k\}}{\arg \min} \ \underset{\xi _i}{\min} \frac{1}{2}\sum _k \left \| \mathbf{w}_k \right \|_2^2 + C\sum _n\xi _n ; \textnormal{ subject to:} \\
    \mathbf{w}_{y_n}.f_{\phi}(x_n)-\mathbf{w}_{k}.f_{\phi}(x_n) \geq 1-\delta _{y_n,k} - \xi _n, \forall n,k
  \end{array}
\end{equation}
where $\mathcal{D}^{\mathit{train}} = \{(x_n,y_n)\}, C$ is the regularization parameter and $\delta_{.,.}$ is the Kronecker delta function. 

Furthermore, the official repository seems to train the model with a 5-way 15-shot and test the model on a 5-way 1-shot. However, we do not consider this to be an accurate study of the effect of task diversity. In our work, we train and test the model in a 5-way 1-shot setting to ensure a fair and accurate comparison with other models. We believe this to be the source of discrepancy in our performance scores. Since we are focused on the relative performance of the samplers for a given model, this discrepancy would not affect our study of task diversity in any manner.

\subparagraph{Hyperparameters}
In our experiments on Omniglot, \textit{mini}ImageNet and \textit{tiered}ImageNet under a 5-way, 1-shot setting, we run the model for 60 epochs with a batch size of 32 and a meta-learning rate of 0.01. We use an SGD optimizer with a momentum of 0.9 and a weight decay of 0.0001 to make gradient steps. We also use a LambdaLR scheduler to train our model. The same hyperparameters are used for training our model on Omniglot under a 20-way 1-shot setting.

We use the same parameters as the \textit{mini}ImageNet to train our model on the Meta-Dataset under a 5-way 1-shot setting. However, we run with a batch size of 16 rather than 32 to accommodate the extensive training period and memory constraints.

\section{Computing Diversity of Latents}
\label{vol}
\textbf{Theorem.} \textit{Let $\Pi$ be an $m$-dimensional parallelotope defined by edge vectors $\mathcal{B}= \left \{ v_1,v_2,...,v_m \right \}$, where $v_i \in \mathbb{R}^n$ for $n\geq m$. That is, we are looking at an $m$-dimensional parallelotope embedded inside $n$-dimensional space. Suppose $\mathcal{A}$ is the $m\times n$ matrix with row vectors $\mathcal{B}$ given by:}

\begin{equation*}
    \mathcal{A} = \begin{pmatrix}
v_1^T\\ 
\vdots\\ 
v_m^T
\end{pmatrix}
\end{equation*}
\textit{Then the $m$-dimensional volume of the paralleletope is given by:}
\begin{equation*}
\left [\text{vol}(\pi)  \right ]^2 = \det(AA^T)
\end{equation*}
\textbf{Proof.} Note that $AA^T$ is an $m\times m$ square matrix. Suppose that $m=1$, then:
\begin{equation*}
\det(AA^T)= \det(v_1v_1^T) = v_1\cdot  v_1 = \left \| v_1 \right \|^2 = \left [\text{vol}_1(v_1)  \right ]^2
\end{equation*}
so the proposition holds for $m=1$. From this base equation, we prove the above theorem by induction. Now, we induct on $m$.

Let us assume the proposition holds for $m^{'}$ such $m^{'}\geq 1$. If we can also prove that the proposition holds for $m^{'}+1$, we would have proved the above theorem. Letting $A_{m^{'}}$ denote the matrix containing rows $v_1$ to $v_{m^{'}}$, we can write $A = A_{m^{'}+1}$ as:
\begin{equation*}
A = \begin{pmatrix}
A_{m^{'}}\\ 
v_{m^{'}+1}^T
\end{pmatrix}
\end{equation*}

We may decompose $v_{m^{'}+1}$ orthogonally as:
\begin{equation*}
v_{m^{'}+1} = v_{\perp } + v_{\parallel }
\end{equation*}
where $v_{\perp}$ lies in the orthogonally complement of the base (i.e., the height of our parallelepipe), and $v_{\perp}\cdot v_i=0$, $\forall$ $1\leq i \leq m^{'}$. Furthermore, $v_{\parallel}$ must be in the span of vectors $\left \{ v_1,v_2,...,v_{m^{'}} \right \}$, such that:
\begin{equation*}
v_{\parallel} = c_1v_1 + ...  c_{m^{'}}v_{m^{'}}
\end{equation*}
We apply a sequence of elementary row operations to $A$, adding a multiple $-c_i$ of row $i$ to row $m^{'}+1$, $\forall$ $1\leq i \leq m^{'}$. We can then write the resulting matrix $B$ as:

\begin{equation*}
    B = \begin{pmatrix}
A_{m^{'}}\\ 
 \\
v_{\perp}^T
\end{pmatrix} = E_{m^{'}}...E_1 A,
\end{equation*}

where each $E_i$ is an elementary matrix adding a multiple of one row to another. Notice that the above operation corresponds to shearing the parallelotope so that the last edge is perpendicular to the base. We see that these operations do not change the determinant as:

\begin{equation*}
\det(BB^T) = \det(E_{m^{'}}...E_1 (AA^T) E_{1}^T...E_{m^{'}}^T) = \det(AA^T)
\end{equation*}

Through block multiplication, we can obtain $BB^T$ as follows:
\begin{equation*}
\begin{aligned}
BB^T ={}& \bigl(\begin{smallmatrix}
 A_{m^{'}}\\ 
 \\
v_{\perp}^T
\end{smallmatrix}\bigr) \bigl(\begin{smallmatrix}
 A_{m^{'}}^T & &
v_{\perp}
\end{smallmatrix}\bigr) \\
={}& \bigl(\begin{smallmatrix}
 A_{m^{'}}A_{m^{'}}^T & & A_{m^{'}}v_{\perp}\\ 
 \\
v_{\perp}^TA_{m^{'}}^T & & v_{\perp}^Tv_{\perp}
\end{smallmatrix}\bigr)\\
={}& \bigl(\begin{smallmatrix}
 A_{m^{'}}A_{m^{'}}^T & & A_{m^{'}}v_{\perp}\\ 
 \\
(A_{m^{'}}v_{\perp})^T & & \left \| v_{\perp} \right \|^2\end{smallmatrix}\bigr)\\
\end{aligned}
\end{equation*}

Furthermore, notice that
\begin{equation*}
\begin{aligned}
A_{m^{'}}v_{\perp} ={}&\begin{pmatrix}
v_1^T\\ 
\vdots\\ 
v_{m^{'}}^T
\end{pmatrix}v_{\perp} =0\\
\end{aligned}
\end{equation*}

Therefore, we have 
\begin{equation*}
\begin{aligned}
BB^T ={}& \bigl(\begin{smallmatrix}
 A_{m^{'}}A_{m^{'}}^T & & 0\\ 
 \\
0^T & & \left \| v_{\perp} \right \|^2\end{smallmatrix}\bigr)\\
\end{aligned}
\end{equation*}

Taking the determinant, we can simplify $\det(BB^T)$ as 
\begin{equation*}
\begin{aligned}
\det(BB^T) ={}&  \left \| v_{\perp} \right \|^2 \det( A_{m^{'}}A_{m^{'}}^T)\\
\end{aligned}
\end{equation*}

By definition, $\left \| v_{\perp} \right \|$ is the height of the parallelotope, and by the induction hypothesis, $\det(A_{m^{'}},A_{m^{'}}^T)$ is the square of the base. Therefore, we have proved the above theorem by induction.

\section{Additional Results}
\label{add_results}

\subsection{Theoretical Analysis}
\label{theory}

In theory, there are two broad classifications of models: (i) Well-specified models and (ii) Misspecified models. For well-specified models, we assume that the model has sufficient capacity, and the data behaves well such that they can be perfectly captured by the model (neural network in our case). This scenario has been discussed in detail in Appendix~\ref{ws_models}. Appendix~\ref{mis_models} considers the scenario of Misspecificed models, where the model does not perfectly fit the data, either due to lack of capacity or due to falling in local minima. We consider the model misspecified when the local optima reached by the neural network is misspecified.

\subsubsection{Well-specified Models}
\label{ws_models}

Let us consider a simple linear regression problem modeled as follows: 

\begin{equation*}
    y_i = k^Tx_i + \eta_i
\end{equation*}

where $k$ is a constant, $x_i$ is the $i^{th}$ input, $y_i$ is output for the $i^{th}$ input, and $\eta$ is the Gaussian noise which is independent of $x$. The class of well-specified models guarantees that the model we choose to fit will have sufficient capacity to capture the ``true model''. Let us assume that we try to learn a linear regression model that best models this ``true model'' of the abovementioned system.

\paragraph{Null Hypothesis:} Let us assume that the sampling schemes of the data distribution used represents the true distribution of data. Under such as assumption, a change in the data distribution would not affect the model learned. Suppose we have a model whose weights are initialized at $\theta$, and we know that the optimal weight for the model is $\theta^*$. Furthermore, $p_1$ and $p_2$ are two sampling mechanisms over the dataset $\mathcal{D}$. Then, a model training mechanism $\mathbb{A}$ which leads to the learnt model $\theta_{\mathsf{final}} = \mathbb{A}(\mathcal{D}, \theta)$ will follow the property of convergence to the optimal solution where:

\begin{equation}
    \theta^* = \mathbb{A}(p_1(\mathcal{D})) = \mathbb{A}(p_2(\mathcal{D}))
\end{equation}

\begin{figure}[hbt!]
\centering
\subfigure[True Data distribution]{%
\includegraphics[width=0.6\linewidth]{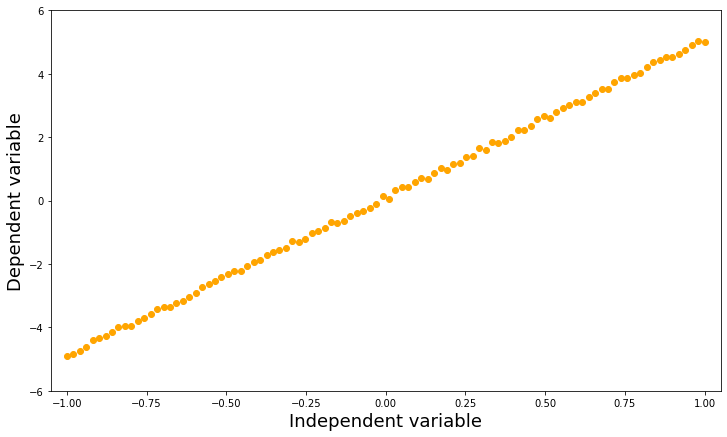}
\label{fig:wd_models_true}}
\quad
\subfigure[Sampling Mechanism 1]{%
\includegraphics[width=0.45\linewidth]{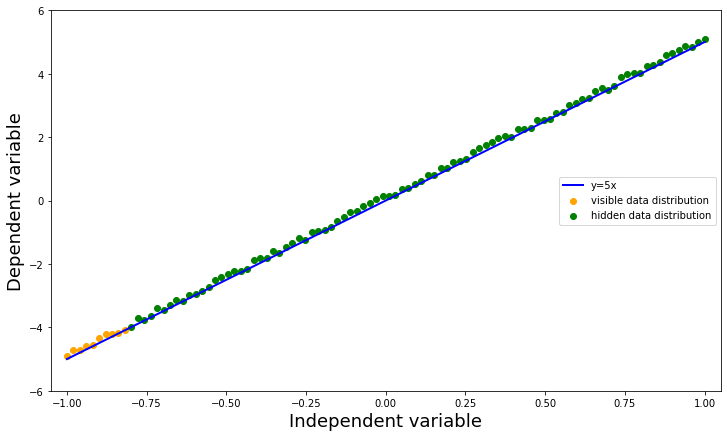}
\label{fig:wd_models_samp1}}
\quad
\subfigure[Sampling Mechanism 2]{%
\includegraphics[width=0.45\linewidth]{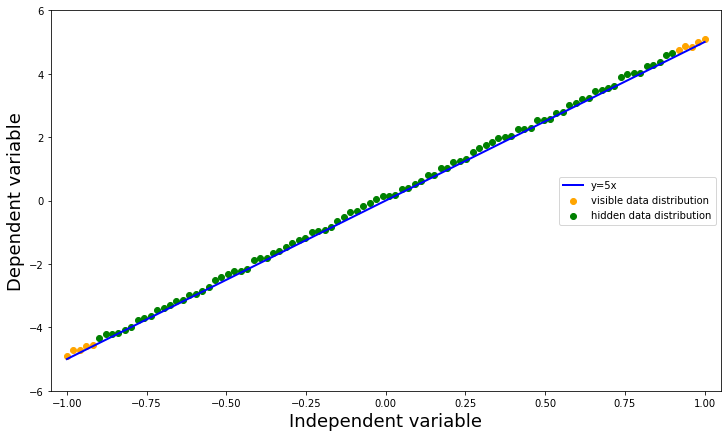}
\label{fig:wd_models_samp2}}

\caption{Well defined models scneario where the model is invariant to the sampling scheme of dataset. Figure~\ref{fig:wd_models_samp1}, and Figure~\ref{fig:wd_models_samp2} depict the scenario where a linear regression model is fit to the data sampled using two different sampling schemes (similar and diverse schemes).}
\label{fig:wm_fig}
\end{figure}

Such a condition can be easily visualized in a 1D regression domain, as highlighted in Figure~\ref{fig:wm_fig}.

\begin{figure}[hbt!]
\centering
\subfigure[True Data distribution]{%
\includegraphics[width=0.6\linewidth]{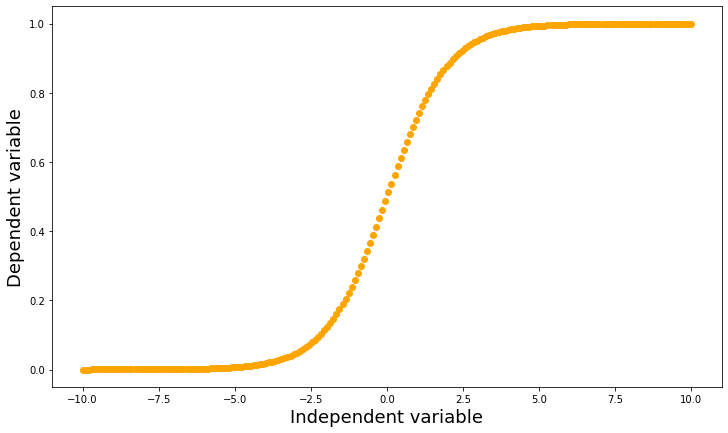}
\label{fig:sig_wd_models_true}}
\quad
\subfigure[Sampling Mechanism 1]{%
\includegraphics[width=0.45\linewidth]{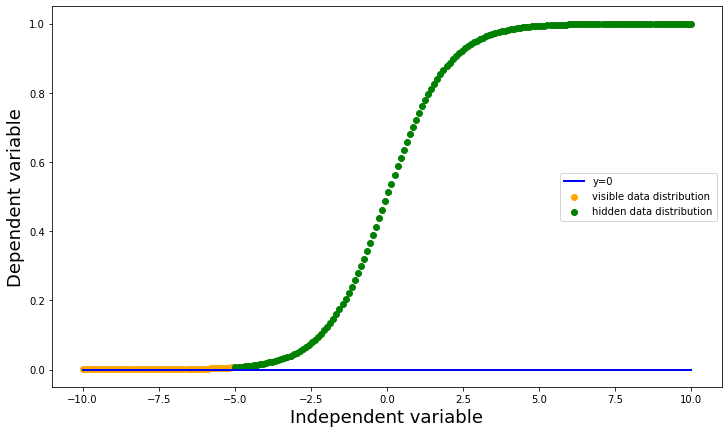}
\label{fig:sig_wd_models_samp1}}
\quad
\subfigure[Sampling Mechanism 2]{%
\includegraphics[width=0.45\linewidth]{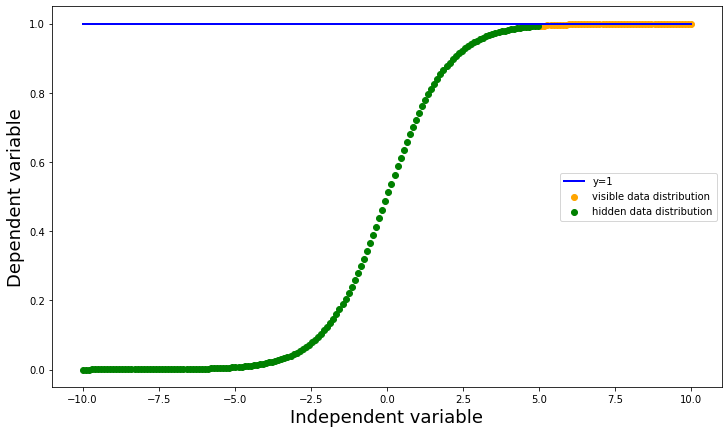}
\label{fig:sig_wd_models_samp2}}

\subfigure[Sampling Mechanism 3]{%
\includegraphics[width=0.45\linewidth]{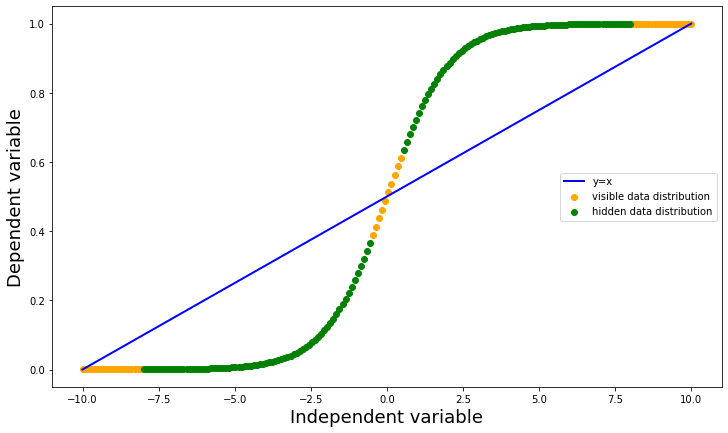}
\label{fig:sig_wd_models_samp3}}

\caption{Well-defined models scenario where the model is not invariant to the sampling scheme of dataset. Figure~\ref{fig:sig_wd_models_true} denotes the true data distribution, and we assume that our neural network is able to successfully capture this ``true model''. Figure~\ref{fig:sig_wd_models_samp1} and Figure~\ref{fig:sig_wd_models_samp2} depict the model learnt when an sub-optimal sampling strategy is used which no longer denotes the true distribution of the ``true model''. The blue line denotes the neural network model trained on this biased data. Figure~\ref{fig:sig_wd_models_samp3} depicts a diverse sampling scheme mechanism, which does depict a closer representation of the ``true model'' than previous versions.}
\label{fig:sig_ws_models}
\end{figure}

\paragraph{Alternative Hypothesis:} In this section, we assume the scenario when the sampling distribution no longer conforms to the actual distribution of the model. We notice that even neural networks (function approximations capable of representing any function given enough capacity) could fail to learn an optimal model in this case. For this scenario, let us consider a sigmoid model as the actual function represented such that:

\begin{equation*}
    y_i = \sigma (x_i)
\end{equation*}

where $x_i$ is the $i^{th}$ input, $y_i$ is output for the $i^{th}$ input as defined above. Let us assume we try to learn a neural network approximation of this model with enough capacity. Figure~\ref{fig:sig_ws_models} denotes our study on such a scenario. Although uniform does guarantee a better depiction of the ``true model'' if sampled for enough epochs, a single sampling step might lead to sub-optimal representation. It is under this intuition that conventional wisdom flourished.

\subsubsection{Misspecified Models}
\label{mis_models}

\begin{figure}[hbt!]
\centering
\subfigure[Individual distributions]{%
\includegraphics[width=0.6\linewidth]{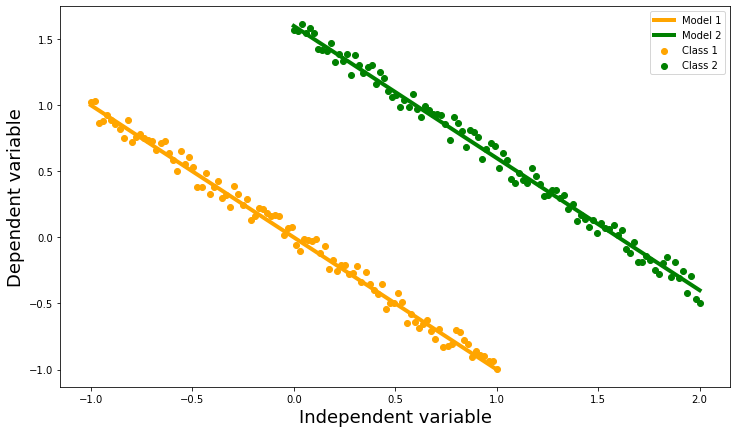}
\label{fig:sp_ind}}
\quad
\subfigure[Sampling Mechanism 1]{%
\includegraphics[width=0.45\linewidth]{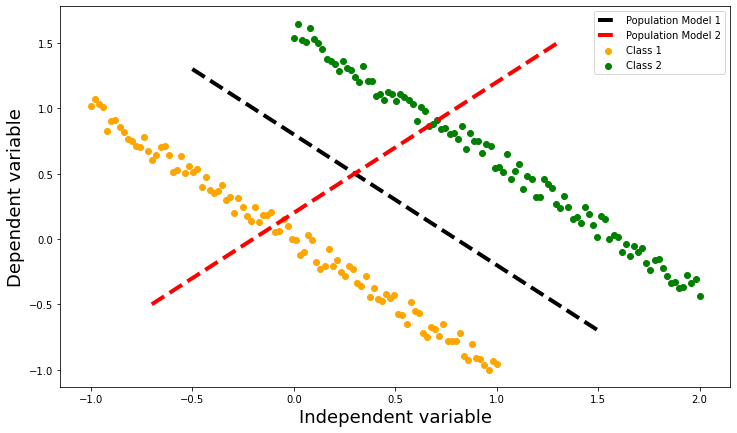}
\label{fig:sp_similar}}
\quad
\subfigure[Sampling Mechanism 2]{%
\includegraphics[width=0.45\linewidth]{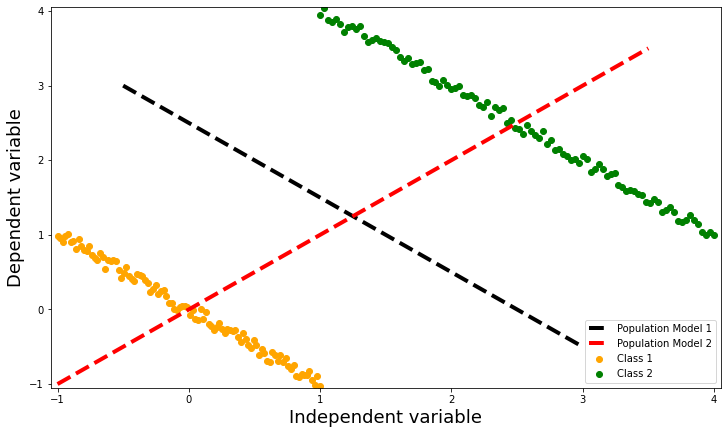}
\label{fig:sp_diverse}}

\caption{Misspecified models. Figure~\ref{fig:sp_ind} depicts the scenario when we fit two different models to each class. Under this scenario, the model is well-specified. However, when we try to model the entire population, we consider two scenarios: (i) Similar classes (Figure~\ref{fig:sp_similar}) and (ii) Diverse classes (Figure~\ref{fig:sp_diverse}). Note that under the similar classes scenario, the model parallel to both classes (black line) is optimal and minimizes the MSE loss of the problem. However, we enter the Simpson Paradox scenario when the points become diverse. It is more suitable for the model to learn the perpendicular line (red line) as the true model instead, as this has a lower MSE than the parallel candidate solution.}
\label{fig:simpson_paradox}
\end{figure}

Under the scenario of misspecified models, the network might either have a low capacity or could be propagated towards a local optimum by the SGD optimizer. This section mainly delves into the Simpson's paradox, a phenomenon in probability and statistics in which a trend appears in several data groups but disappears or reverses when the groups are combined. This phenomenon is very closely related to the problem of understanding the latent confounders associated with the problem. Figure~\ref{fig:simpson_paradox} depicts this intuition better.

\subsubsection{Relation to Empirical Results}

Although Appendix~\ref{ws_models} suggests that the conventional wisdom holds in specific scenarios and models, meta-learning, in general, seems to follow the misspecified model scenario as discussed in Appendix~\ref{mis_models}. Hence, increasing diversity seems to fall into the Simpson's paradox scenario and warrants additional research to mitigate this scenario. This explanation is also in line with our explanation for the robustness of MetaOptNet due to its SVM usage. This work also shows that using the NDTB and NDB sampler and training on a tiny data fragment can perform similarly to a model trained using Uniform Sampler. Again, this could be closely linked to Simpson's paradox, and our previous explanation follows. Furthermore, we notice that the current pool of models and methods are either not expressive enough or are not successful at mitigating these confounders that appear with ``diverse sampling schemes''. We believe this is an important research domain that needs to be highlighted to the research community. Research on this problem would further strengthen the robustness and understanding of our methods.

\subsection{Results on Regression}

\begin{figure}[ht]
\centering
\subfigure[Uniform]{%
\includegraphics[width=0.3\linewidth]{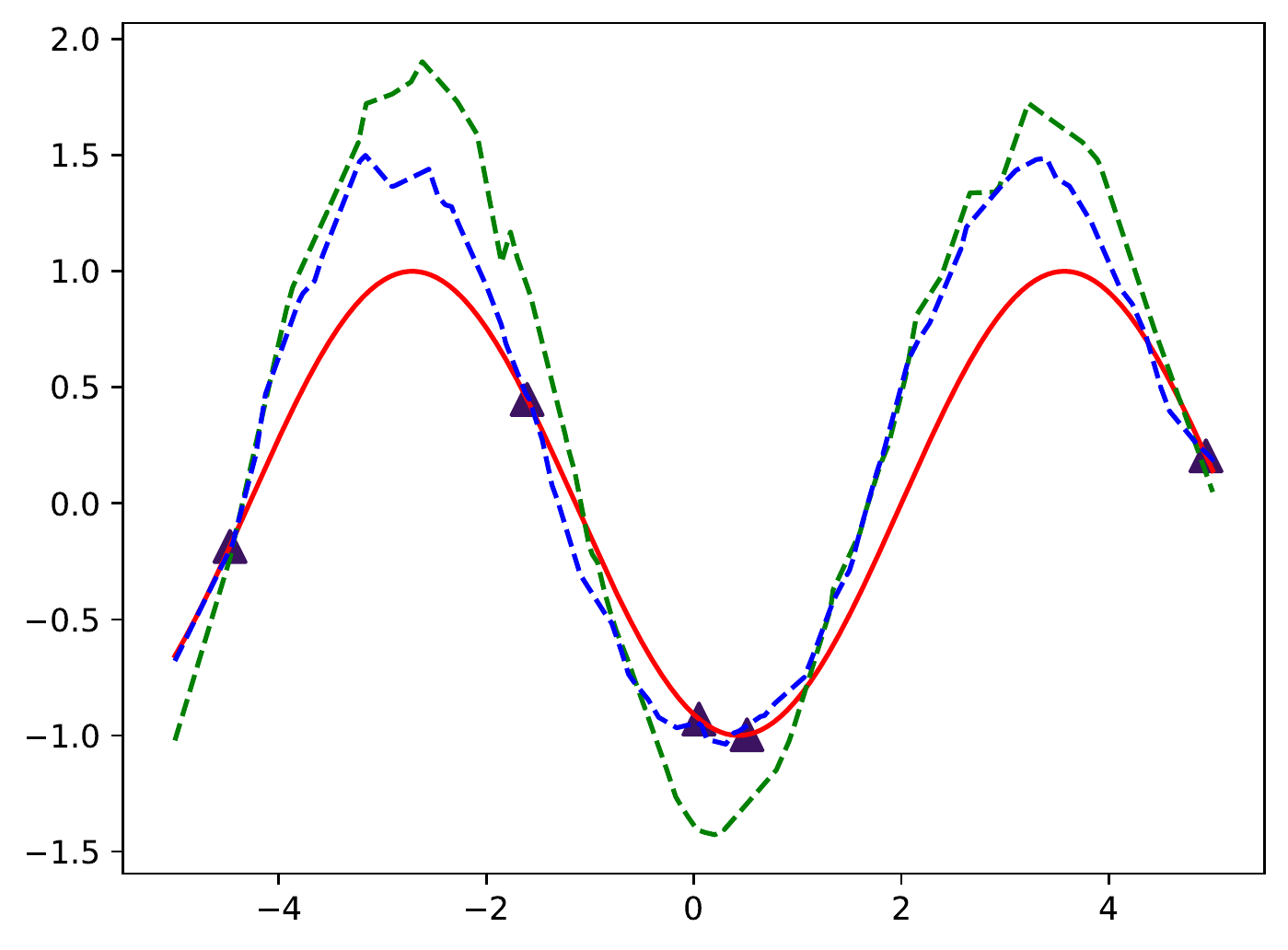}
\label{fig:sinusoid5_1-1}}
\quad
\subfigure[No Diversity Task]{%
\includegraphics[width=0.3\linewidth]{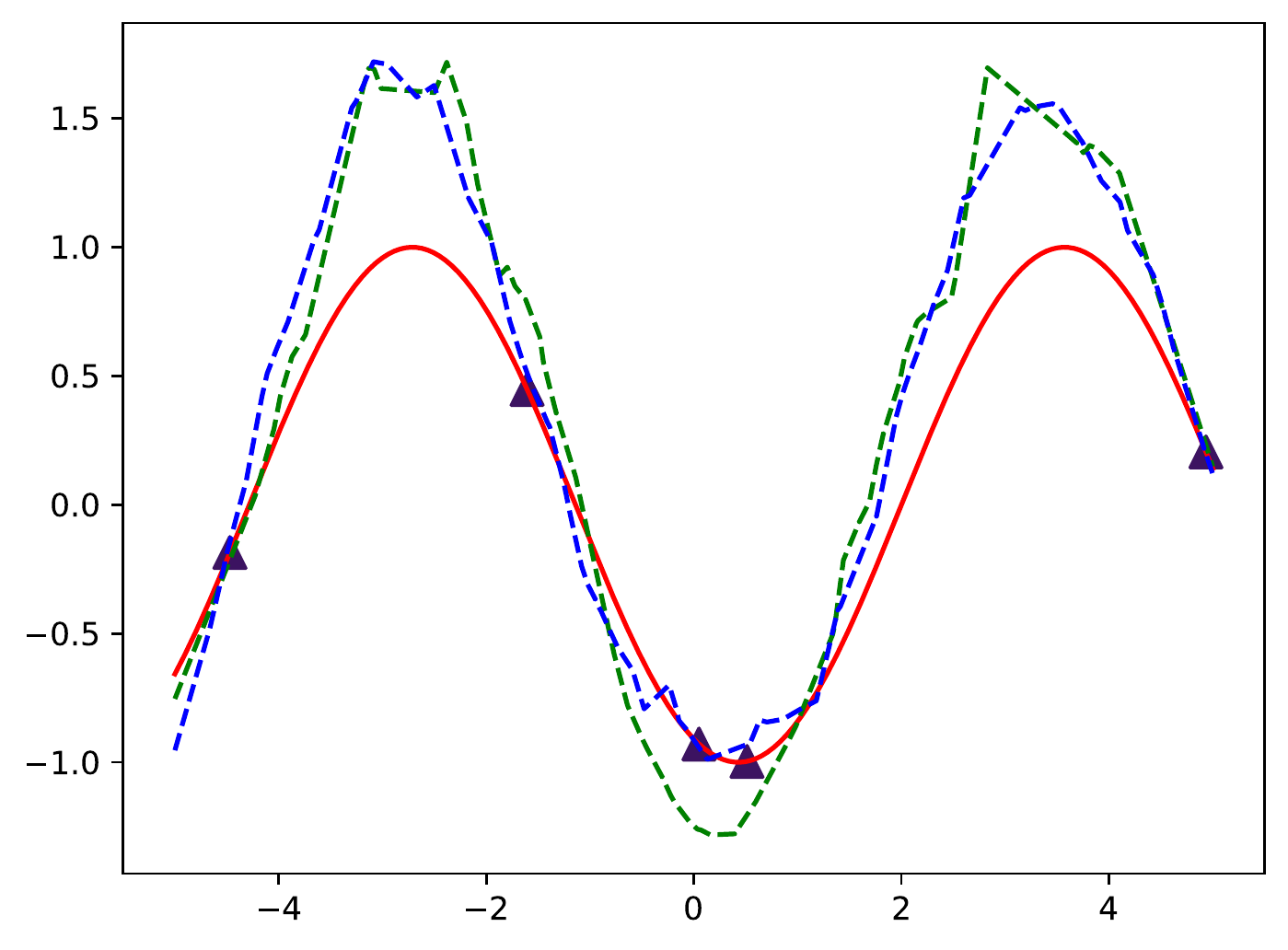}
\label{fig:sinusoid5_1-2}}
\quad
\subfigure[No Diversity Batch]{%
\includegraphics[width=0.3\linewidth]{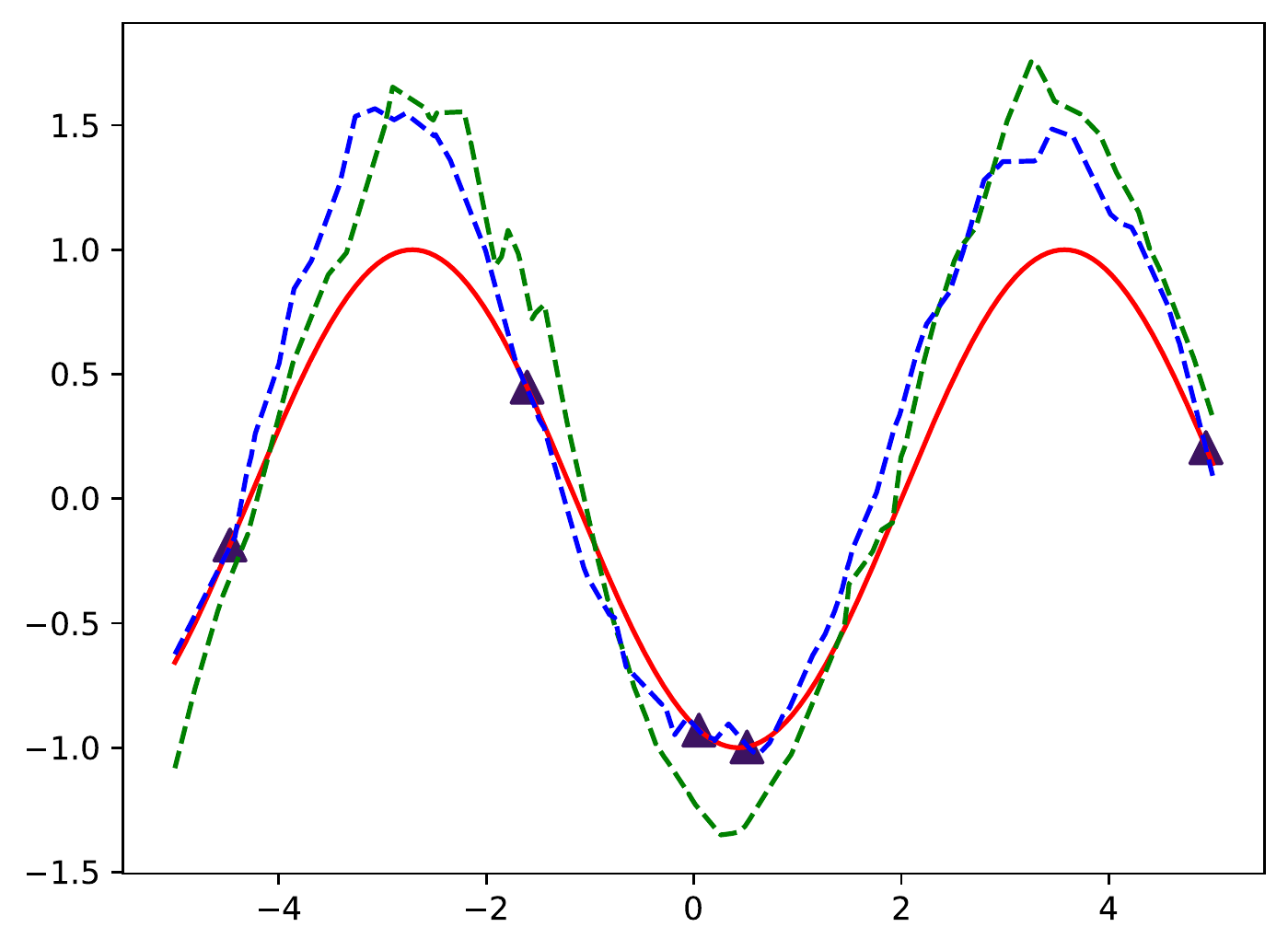}
\label{fig:sinusoid5_1-3}}

\subfigure[No Diversity Tasks/Batch]{%
\includegraphics[width=0.3\linewidth]{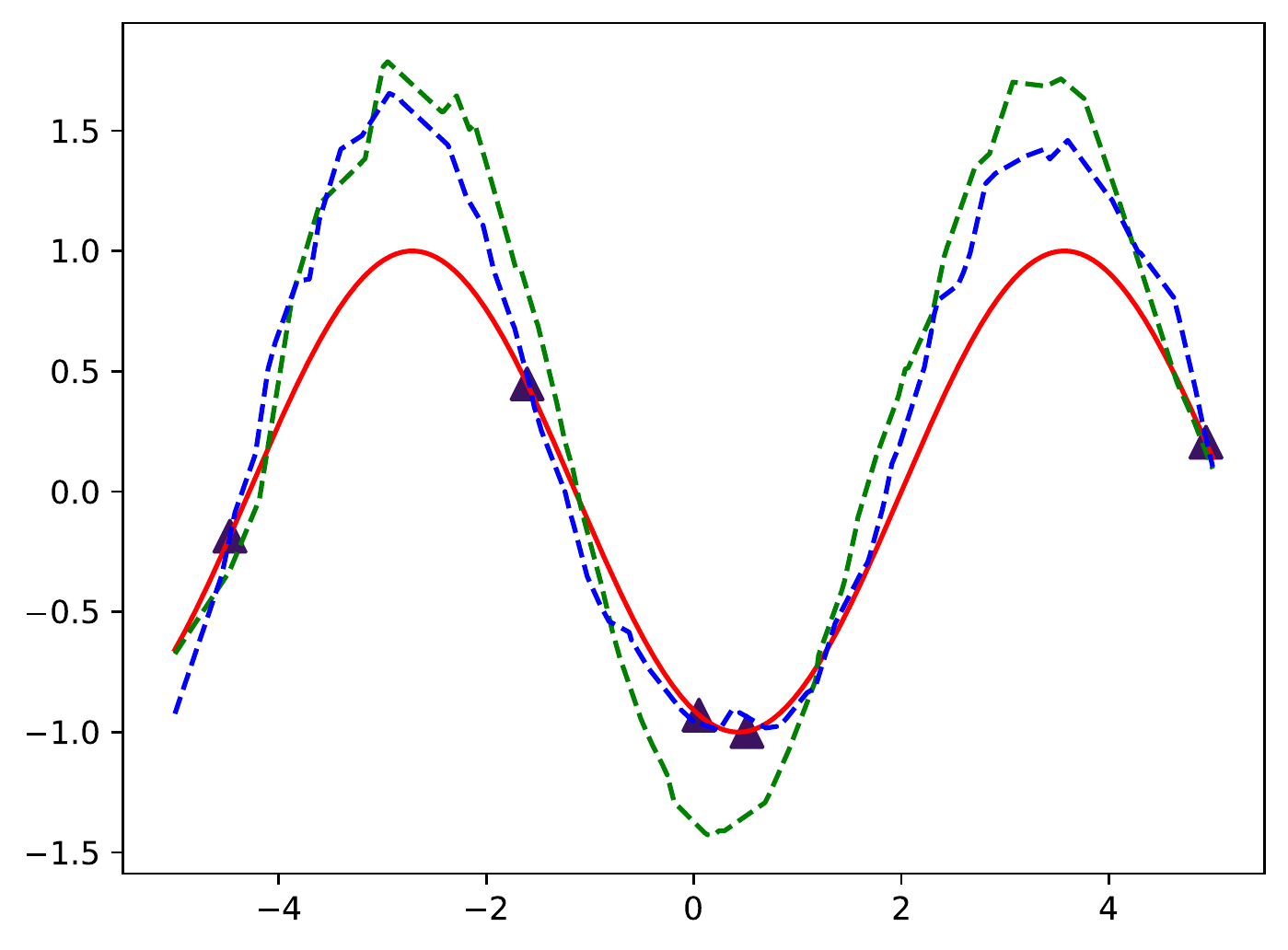}
\label{fig:sinusoid5_1-4}}
\quad
\subfigure[Single]{%
\includegraphics[width=0.3\linewidth]{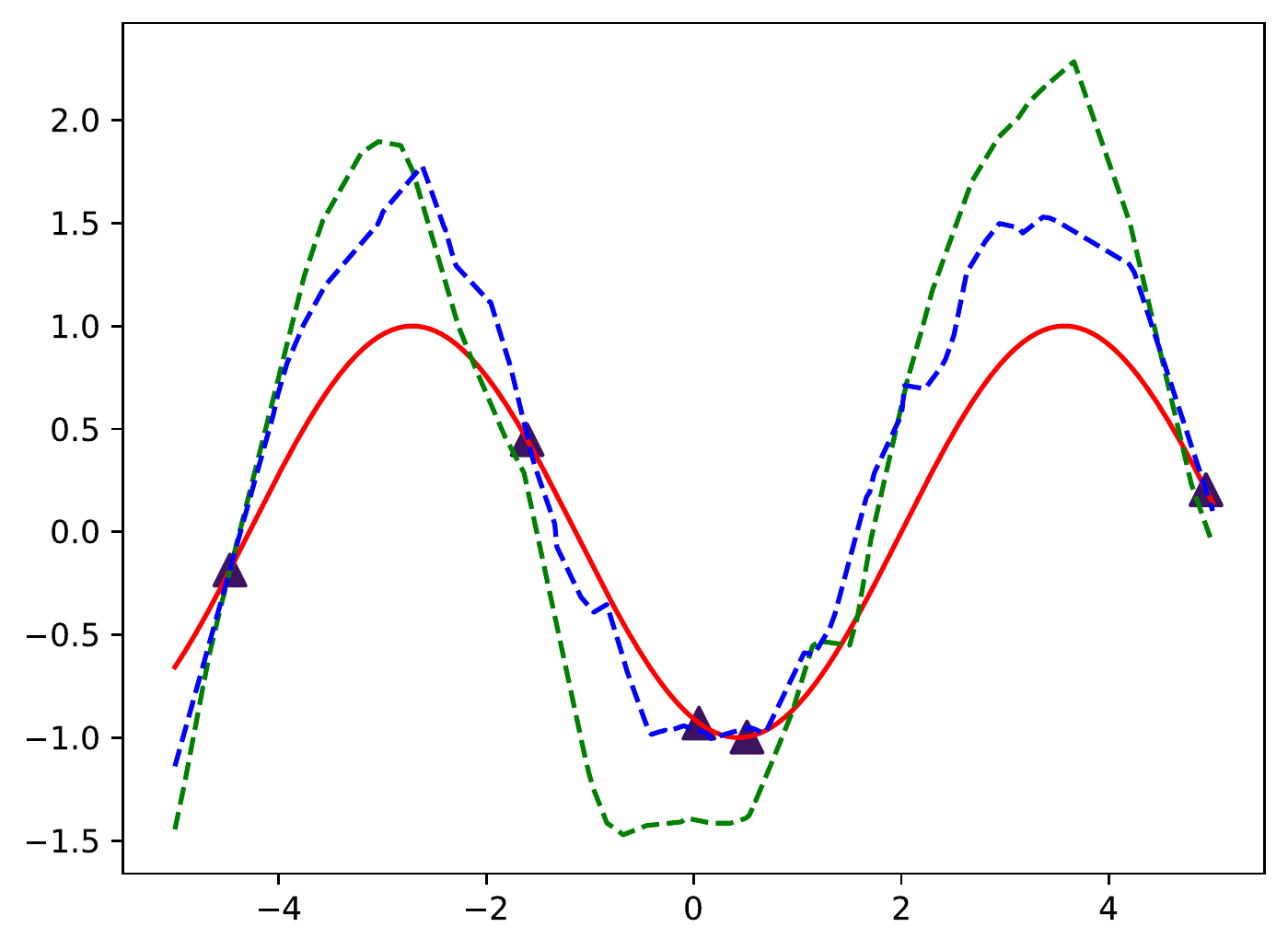}
\label{fig:sinusoid5_1-5}}
\quad
\subfigure[OHTM]{%
\includegraphics[width=0.3\linewidth]{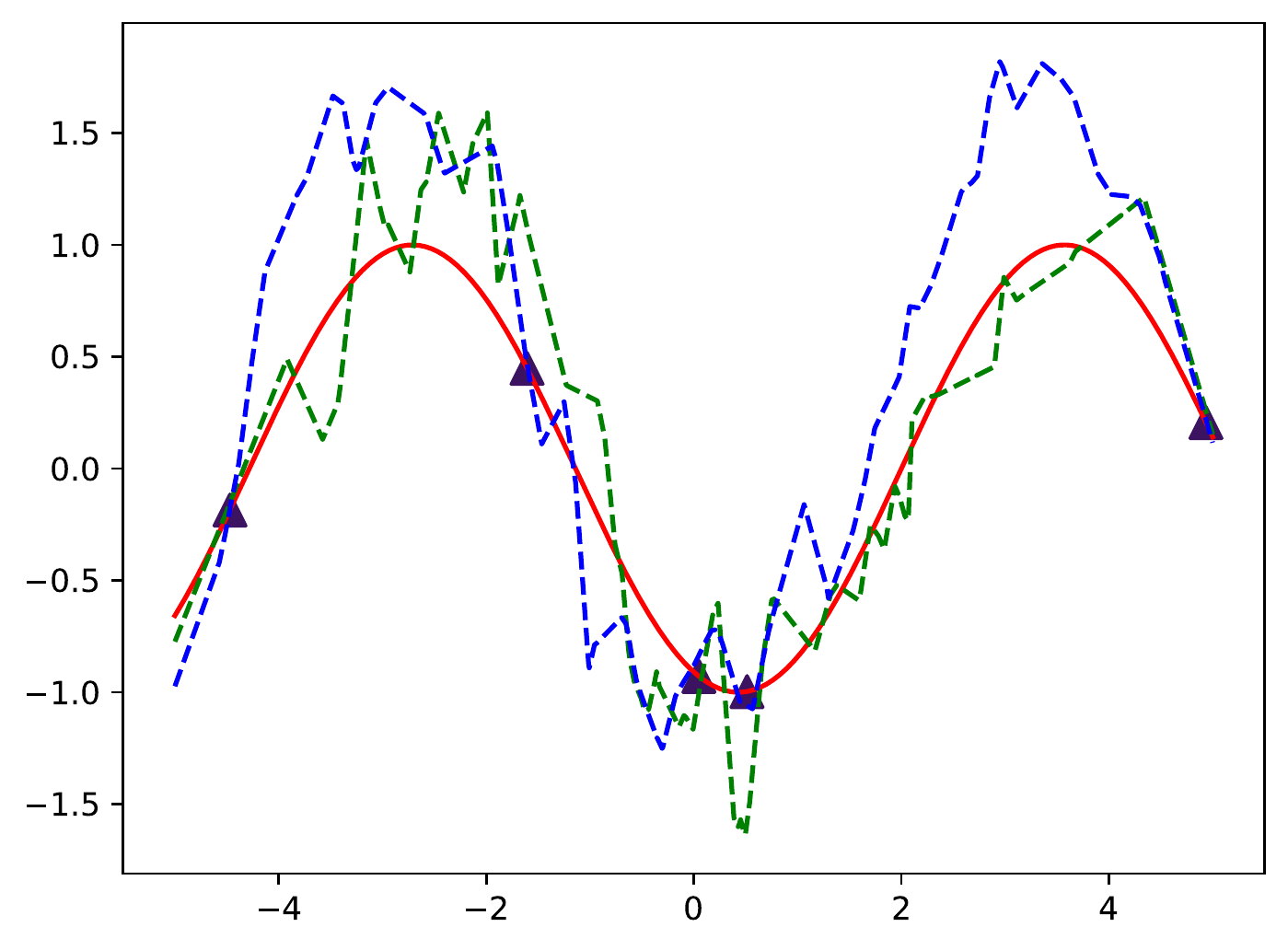}
\label{fig:sinusoid5_1-6}}

\caption{Few-shot adaptation for the simple regression task on Sinusoid 5-shot dataset. The ground truth is denoted by "\protect\tikz \protect\draw[red,thick] (0,0) -- (0.4,0);". The predicted output of MAML and Reptile are denoted by "\protect\tikz \protect\draw[green,densely dashed] (0,0) -- (0.4,0);" and "\protect\tikz \protect\draw[blue,densely dashed] (0,0) -- (0.4,0);" respectively. The training points used for computing gradients is denoted by \textcolor{purple}{$\blacktriangle$}.}
\label{fig:regression_sinusoid_5}
\end{figure}

\begin{figure}[ht]
\centering
\subfigure[Uniform]{%
\includegraphics[width=0.3\linewidth]{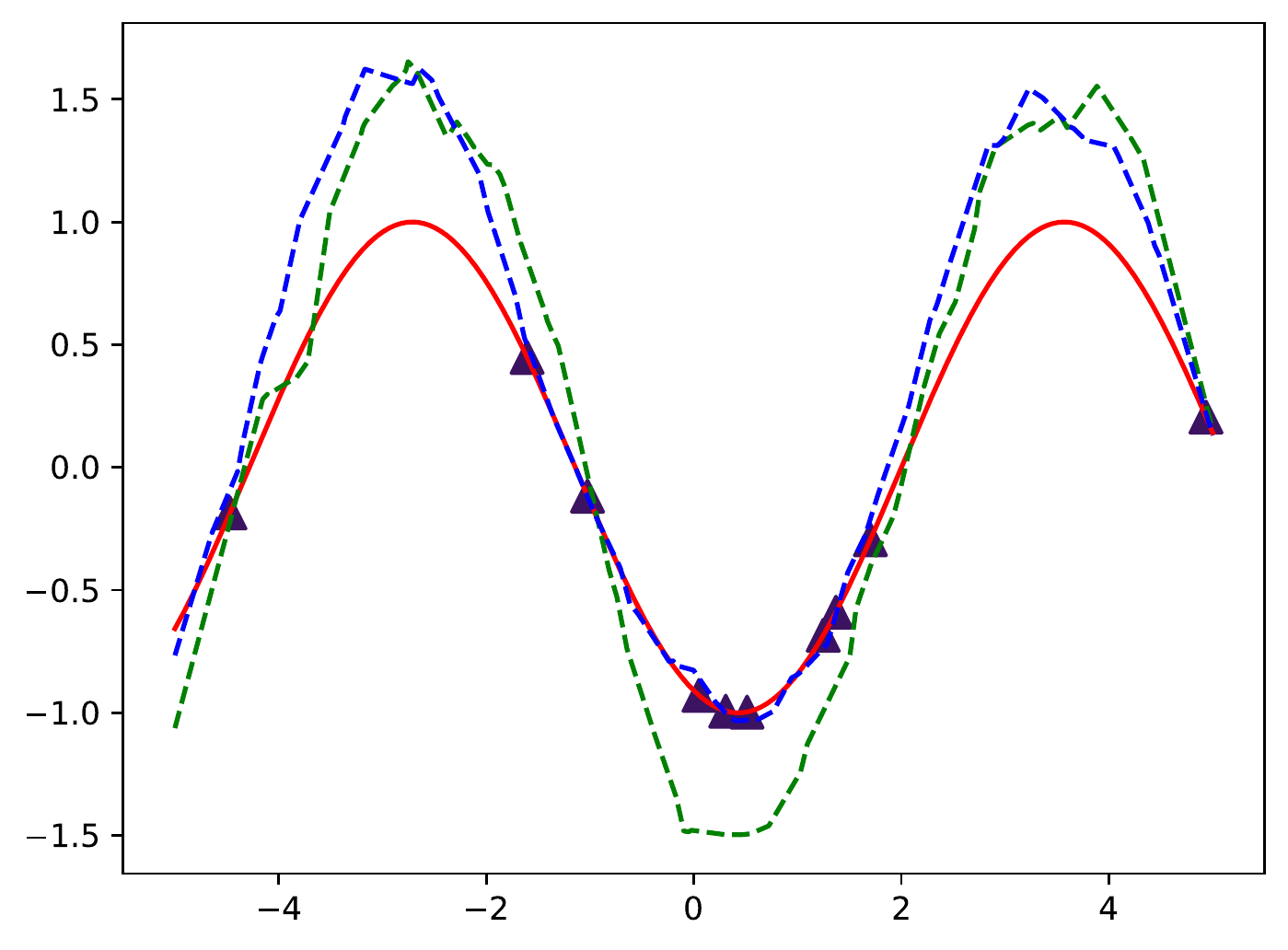}
\label{fig:sinusoid10_1-1}}
\quad
\subfigure[No Diversity Task]{%
\includegraphics[width=0.3\linewidth]{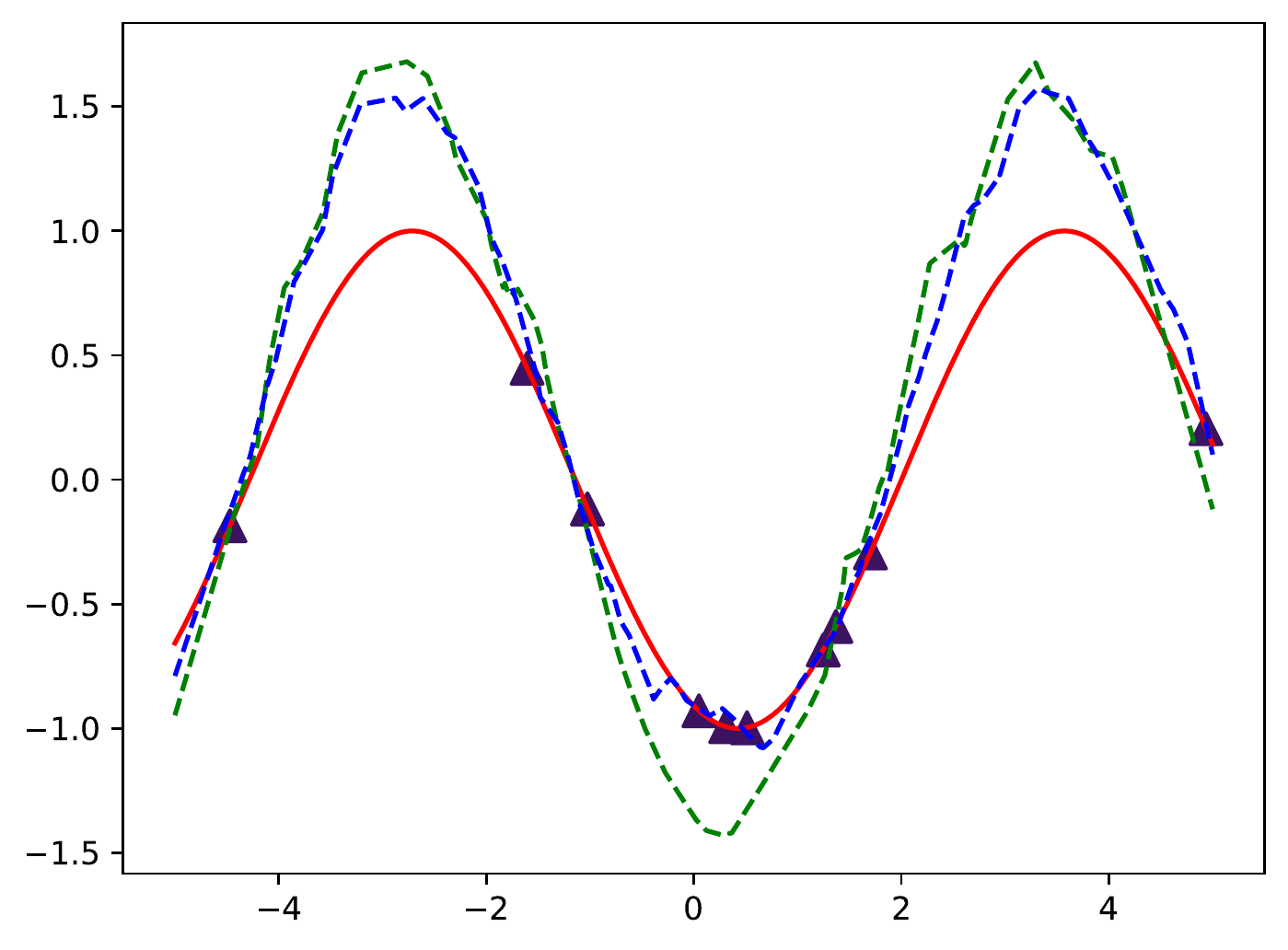}
\label{fig:sinusoid10_1-2}}
\quad
\subfigure[No Diversity Batch]{%
\includegraphics[width=0.3\linewidth]{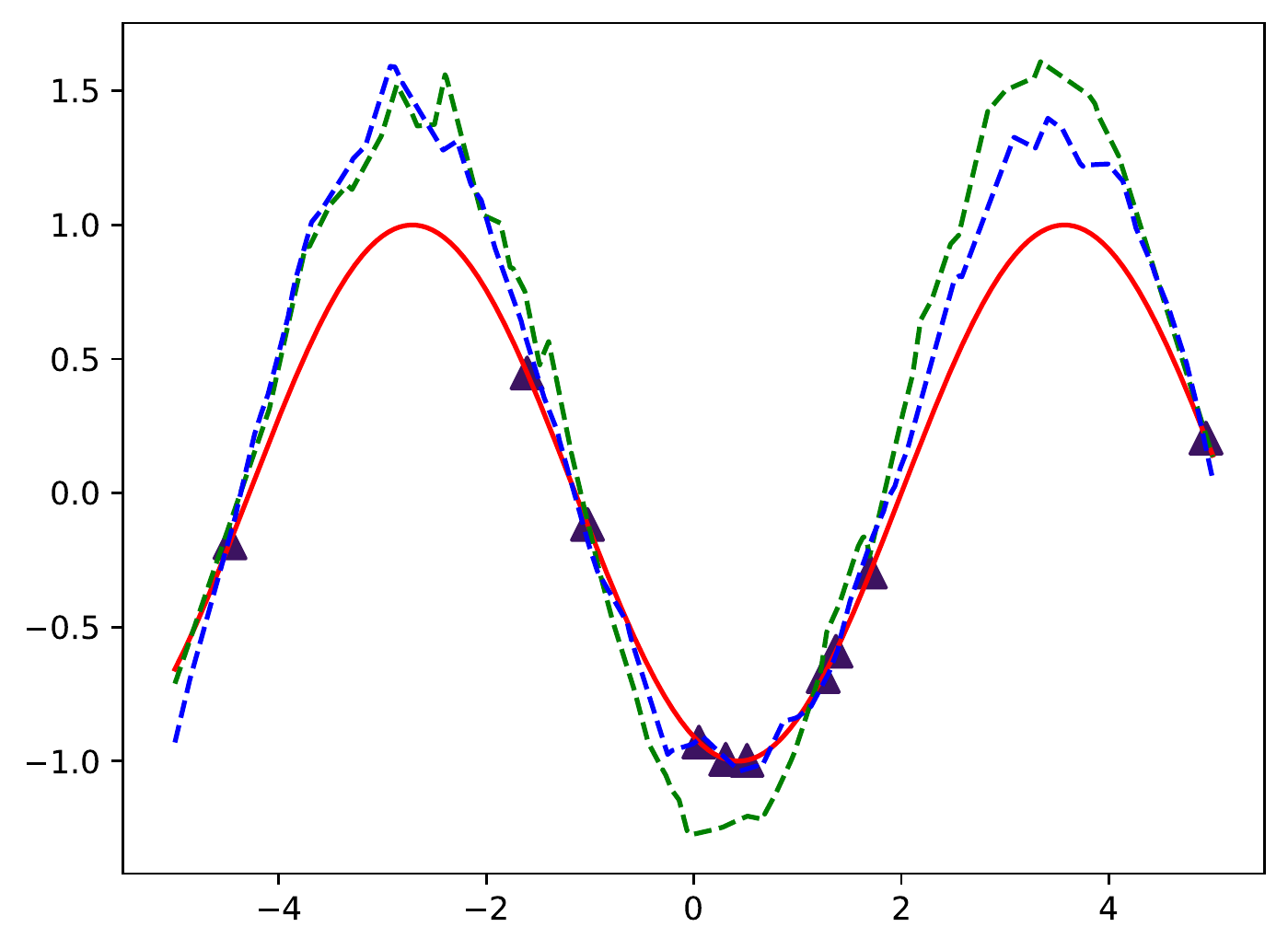}
\label{fig:sinusoid10_1-3}}

\subfigure[No Diversity Tasks/Batch]{%
\includegraphics[width=0.3\linewidth]{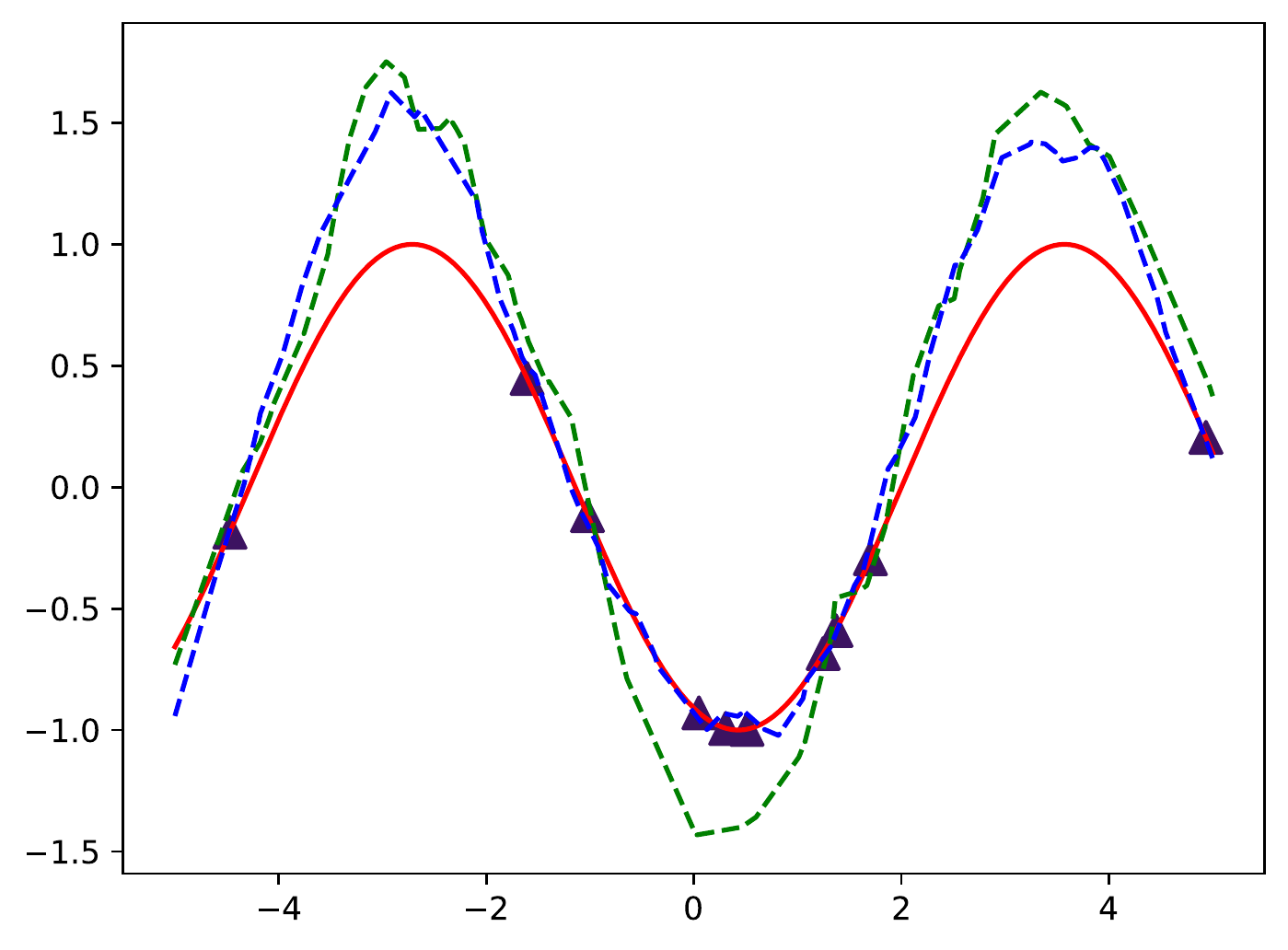}
\label{fig:sinusoid10_2-4}}
\quad
\subfigure[Single]{%
\includegraphics[width=0.3\linewidth]{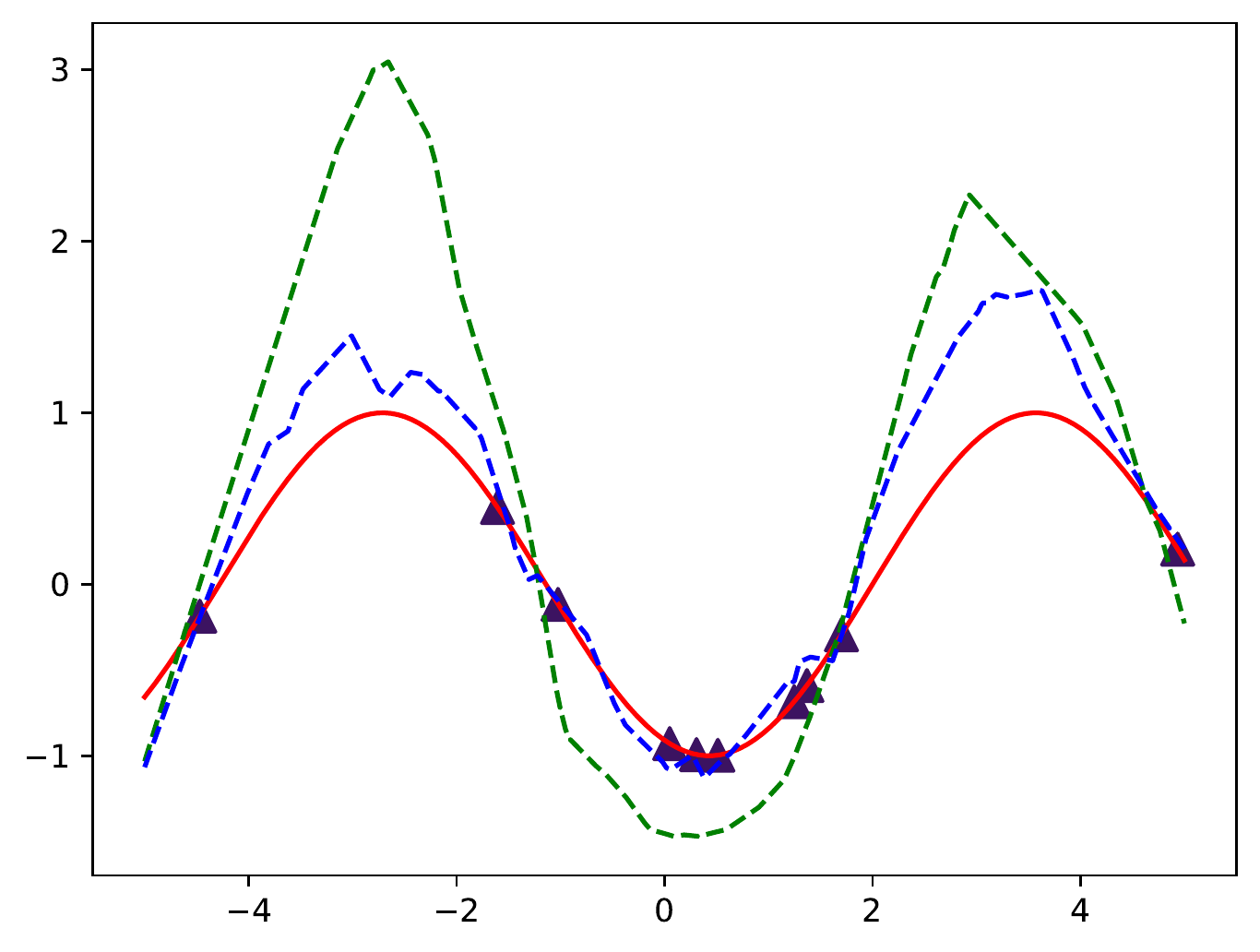}
\label{fig:sinusoid10_2-5}}
\quad
\subfigure[OHTM]{%
\includegraphics[width=0.3\linewidth]{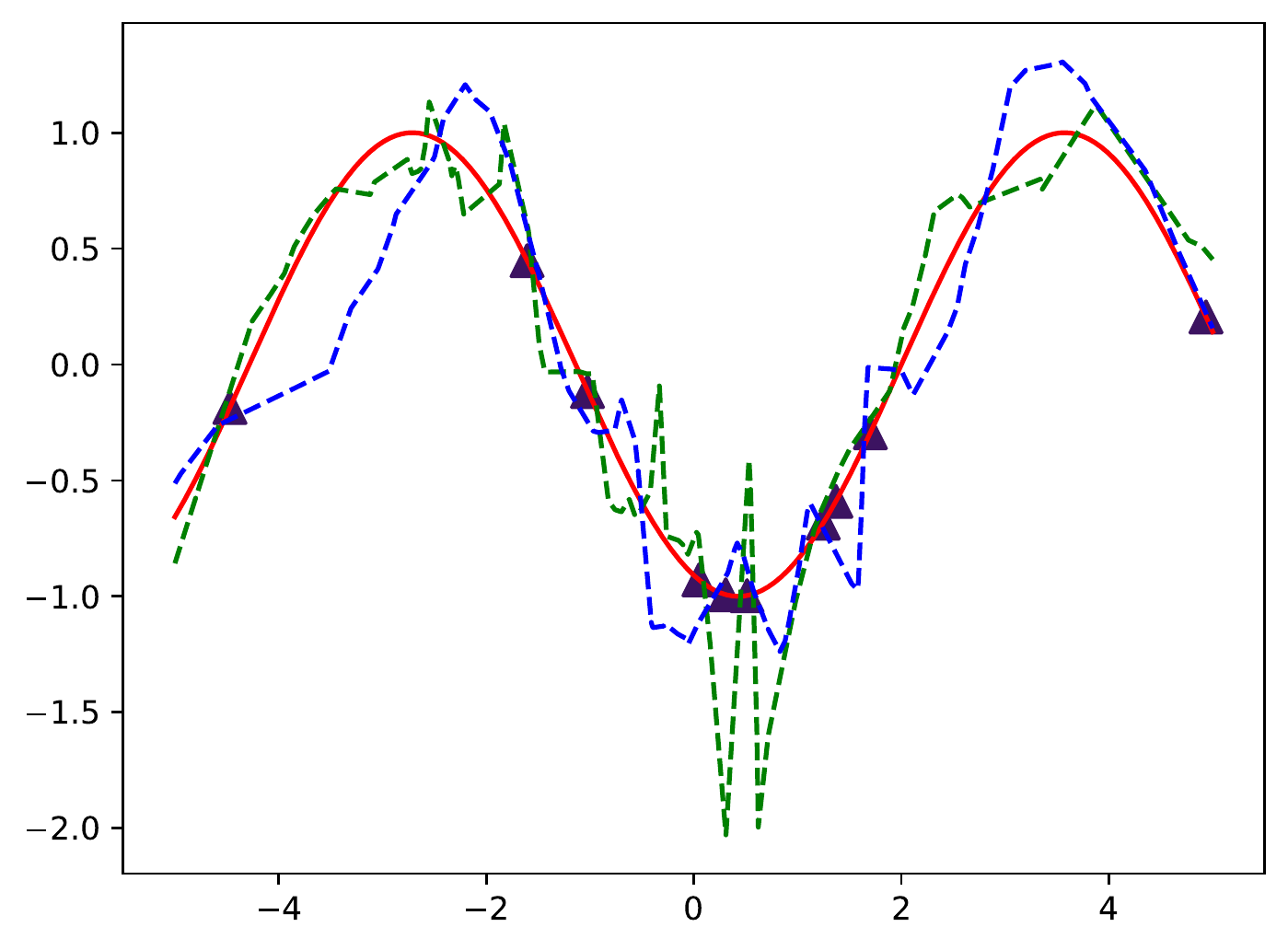}
\label{fig:sinusoid10_2-6}}

\caption{Few-shot adaptation for the simple regression task on Sinusoid 10-shot dataset. The ground truth is denoted by "\protect\tikz \protect\draw[red,thick] (0,0) -- (0.4,0);". The predicted output of MAML and Reptile are denoted by "\protect\tikz \protect\draw[green,densely dashed] (0,0) -- (0.4,0);" and "\protect\tikz \protect\draw[blue,densely dashed] (0,0) -- (0.4,0);" respectively. The training points used for computing gradients is denoted by \textcolor{purple}{$\blacktriangle$}.}
\label{fig:regression_sinusoid_10}
\end{figure}

We further experiment on the few-shot regression regime to maintain our findings on classification. For this problem, we trained on three few-shot regression datasets: (i) Sinusoid (\cite{finn2017model}), (ii)Sinusoid \& Line (\cite{finn2018probabilistic}), and (iii) Harmonic (\cite{lacoste2018uncertainty}. Table~\ref{Results_Regression_5} presents our results using the mean square error metric on the 5-shot and 10-shot settings for all three datasets. Figure~\ref{fig:regression_sinusoid_5} and Figure~\ref{fig:regression_sinusoid_10} presents our results on the Sinusoid dataset specifically for easier comparison. We use the same network for both MAML and Reptile, with two hidden layers, each of size 40.

\subparagraph{Hyperparameters} For both MAML and Reptile, we train the model for 150 epochs and use the same hyperparameters as Omniglot. Only in the case of MAML, we use a step size of 0.001 for the Sinusoid and Sinusoid \& Line dataset instead of the traditional 0.4.

\paragraph{Results} We again observe that samplers that limit diversity, such as NDB, NDT/B samplers achieve performances similar to the uniform sampler with no adverse effects. Astonishingly, we also notice that the NDT sampler seems to outperform all other samplers in the regression domain. Furthermore, similar to our previous findings, samplers that increase diversity, such as the OHTM sampler, do not perform very well in this domain either, as shown in Figure~\ref{fig:regression_sinusoid_5} and Figure~\ref{fig:regression_sinusoid_10}. We cannot train using DPP-oriented samplers for regression since the data is continuous, and computing features or representations for the entire range would be theoretically impossible.

\begin{table*}[!htbp]
\centering
\resizebox{0.7\textwidth}{!}{%
\begin{tabular}{c|c|c|c|c|c}
\hline
&&\multicolumn{2}{|c|}{\textbf{5-shot}}&\multicolumn{2}{|c|}{\textbf{10-shot}} \\ \hline
\textbf{Dataset} &\textbf{Sampler} & {\textit{\textbf{MAML}}} & {\textit{\textbf{Reptile}}} & {\textit{\textbf{MAML}}} & {\textit{\textbf{Reptile}}} \\ \hline
\multirow{6}{*}{\textbf{Sinusoid}}
& Uniform Sampler           &  0.94 {\footnotesize $\pm$ 0.06} & 0.37 {\footnotesize $\pm$ 0.04} &  \textbf{0.50 {\footnotesize $\pm$ 0.03}} & 0.12 {\footnotesize $\pm$ 0.01} \\ \cline{2-6}
& No Diversity Task Sampler & 0.79 {\footnotesize $\pm$ 0.06} \textcolor{green}{\ddag} & \textbf{0.35 {\footnotesize $\pm$ 0.35}} &  0.54 {\footnotesize $\pm$ 0.04} \textcolor{red}{\dag} & \textbf{0.12 {\footnotesize $\pm$ 0.01}} \textcolor{red}{\dag}\\ \cline{2-6}
& No Diversity Batch Sampler               & \textbf{0.77 {\footnotesize $\pm$ 0.05}} \textcolor{green}{\ddag} & 0.37 {\footnotesize $\pm$ 0.04} \textcolor{red}{\dag} &  0.56 {\footnotesize $\pm$ 0.04} \textcolor{green}{\ddag} & 0.13 {\footnotesize $\pm$ 0.01} \textcolor{red}{\dag}\\ \cline{2-6}
& No Diversity Tasks per Batch Sampler              & 0.92 {\footnotesize $\pm$ 0.06} \textcolor{red}{\dag} & 0.37 {\footnotesize $\pm$ 0.04} \textcolor{red}{\dag}  &  0.52 {\footnotesize $\pm$ 0.04} \textcolor{red}{\dag} & 0.12 {\footnotesize $\pm$ 0.01} \textcolor{red}{\dag}\\ \cline{2-6}
& Single Batch Uniform Sampler        & 1.94 {\footnotesize $\pm$ 0.11} & 0.71 {\footnotesize $\pm$ 0.06}  &  1.39 {\footnotesize $\pm$ 0.07} & 0.25 {\footnotesize $\pm$ 0.02} \\ \cline{2-6}
& OHTM Sampler              & 1.17 {\footnotesize $\pm$ 0.08} & 0.84 {\footnotesize $\pm$ 0.30}  &  0.79 {\footnotesize $\pm$ 0.05} & 0.38 {\footnotesize $\pm$ 0.03} \\ \hline \hline

\multirow{6}{*}{\textbf{Sinusoid and Line}}
& Uniform Sampler           &  4.08 {\footnotesize $\pm$ 0.32} & 3.74 {\footnotesize $\pm$ 3.07}  &  2.78 {\footnotesize $\pm$ 0.22} & 0.66 {\footnotesize $\pm$ 0.13} \\ \cline{2-6}
& No Diversity Task Sampler & \textbf{3.91 {\footnotesize $\pm$ 0.32}} \textcolor{red}{\dag} & \textbf{2.80 {\footnotesize $\pm$ 0.30}} \textcolor{red}{\dag} &  2.70 {\footnotesize $\pm$ 0.23}  \textcolor{red}{\dag} & 0.70 {\footnotesize $\pm$ 0.13}  \textcolor{red}{\dag}\\ \cline{2-6}
& No Diversity Batch Sampler               & 4.07 {\footnotesize $\pm$ 0.34} \textcolor{red}{\dag} & 2.18 {\footnotesize $\pm$ 0.27} \textcolor{red}{\dag}  &  \textbf{2.66 {\footnotesize $\pm$ 0.22}} \textcolor{red}{\dag} & 0.69 {\footnotesize $\pm$ 0.11}  \textcolor{red}{\dag}\\ \cline{2-6}
& No Diversity Tasks per Batch Sampler              & 4.17 {\footnotesize $\pm$ 0.37} \textcolor{red}{\dag} & 2.82 {\footnotesize $\pm$ 0.31} \textcolor{red}{\dag}  & 2.73 {\footnotesize $\pm$ 0.22} \textcolor{red}{\dag} & \textbf{0.55 {\footnotesize $\pm$ 0.08}}  \textcolor{red}{\dag}\\ \cline{2-6}
& Single Batch Uniform Sampler        & 6.62 {\footnotesize $\pm$ 0.57} \textcolor{red}{\dag} & 5.52 {\footnotesize $\pm$ 4.45} &  8.48 {\footnotesize $\pm$ 0.94} & 1.19 {\footnotesize $\pm$ 0.15} \\ \cline{2-6}
& OHTM Sampler              & 4.43 {\footnotesize $\pm$ 0.33} & 3.67 {\footnotesize $\pm$ 0.51}  &  3.36 {\footnotesize $\pm$ 0.28} & 1.19 {\footnotesize $\pm$ 0.15} \\ \hline \hline

\multirow{6}{*}{\textbf{Harmonic}}
& Uniform Sampler           & \textbf{1.07 {\footnotesize $\pm$ 0.07}} & 1.20 {\footnotesize $\pm$ 0.08} &  1.07 {\footnotesize $\pm$ 0.07} & 1.05 {\footnotesize $\pm$ 0.07} \\ \cline{2-6}
& No Diversity Task Sampler & 1.12 {\footnotesize $\pm$ 0.07} \textcolor{red}{\dag} & 1.22 {\footnotesize $\pm$ 0.09} \textcolor{red}{\dag} &  1.07 {\footnotesize $\pm$ 0.07} \textcolor{red}{\dag} & 1.07 {\footnotesize $\pm$ 0.07} \textcolor{red}{\dag}\\ \cline{2-6}
& No Diversity Batch Sampler               & 1.11 {\footnotesize $\pm$ 0.07} \textcolor{red}{\dag} & 1.18 {\footnotesize $\pm$ 0.09} \textcolor{red}{\dag} &  \textbf{1.03 {\footnotesize $\pm$ 0.06}} \textcolor{red}{\dag} & 1.08 {\footnotesize $\pm$ 0.07} \textcolor{red}{\dag} \\ \cline{2-6}
& No Diversity Tasks per Batch Sampler              & 1.14 {\footnotesize $\pm$ 0.07} \textcolor{red}{\dag} & 1.18 {\footnotesize $\pm$ 0.08} \textcolor{red}{\dag} &  1.05 {\footnotesize $\pm$ 0.07} \textcolor{red}{\dag} & \textbf{1.05 {\footnotesize $\pm$ 0.07}} \textcolor{red}{\dag} \\ \cline{2-6}
& Single Batch Uniform Sampler        & 1.19 {\footnotesize $\pm$ 0.08} & \textbf{1.10 {\footnotesize $\pm$ 0.08}} \textcolor{red}{\dag} &  1.06 {\footnotesize $\pm$ 0.07} & 1.08 {\footnotesize $\pm$ 0.07} \textcolor{red}{\dag} \\ \cline{2-6}
& OHTM Sampler              & 1.13 {\footnotesize $\pm$ 0.08} & 1.18 {\footnotesize $\pm$ 0.08} \textcolor{red}{\dag} &  1.42 {\footnotesize $\pm$ 0.15} & 1.10 {\footnotesize $\pm$ 0.07} \textcolor{red}{\dag} \\ \hline \hline

\end{tabular}%
}
\caption{Performance metric of our models on different task samplers in the few-shot regression setting.}
\label{Results_Regression_5}
\end{table*}

\subsection{Results on Classification}
\begin{figure}[!htbp]
    \centering
    \begin{adjustbox}{center}
    \includegraphics[height=0.4\textheight, keepaspectratio]{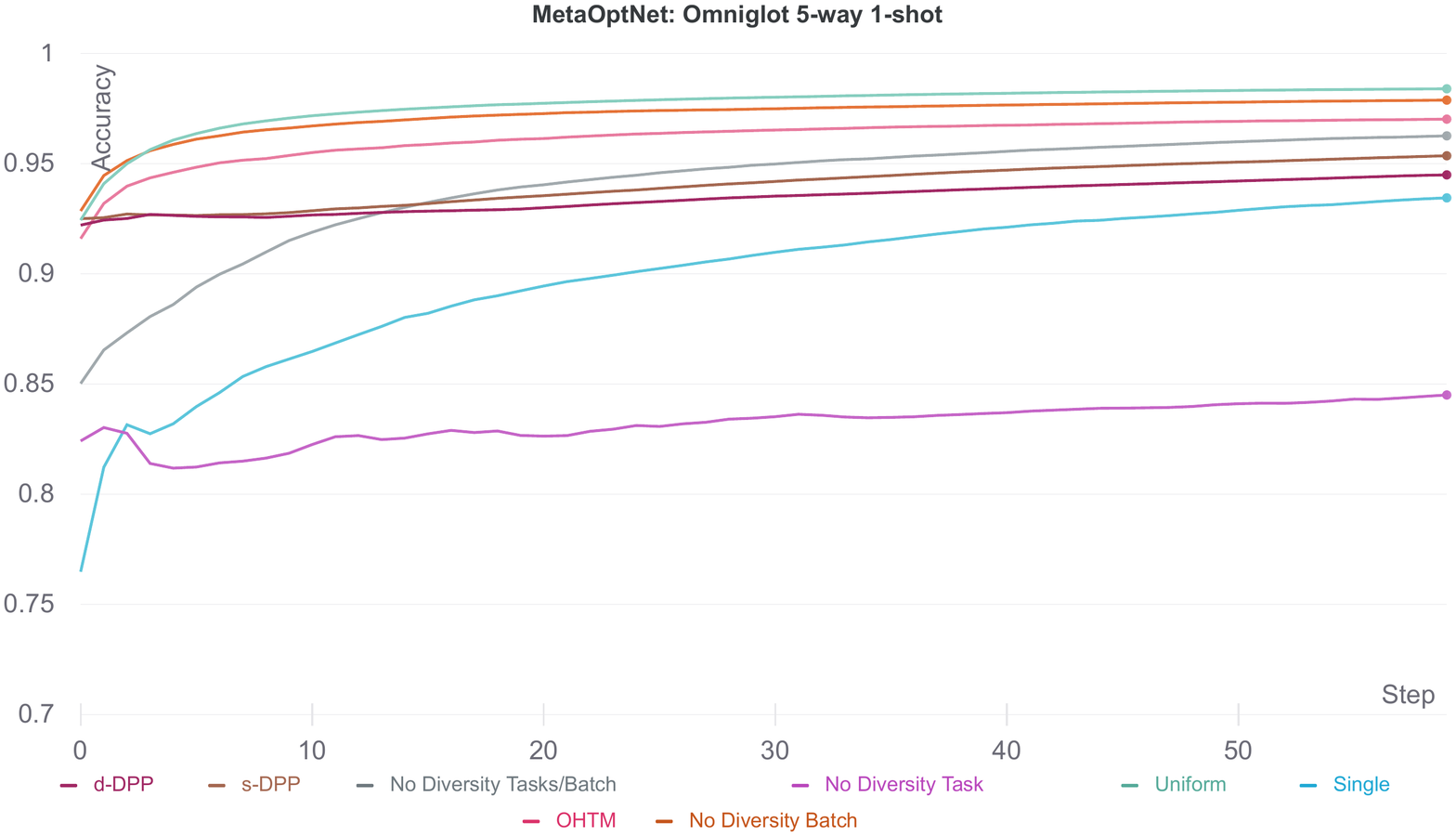}
    \end{adjustbox}
    \caption{Convergence curve of MetaOptNet model on Omniglot 5-way 1-shot.}
    \label{fig:metaoptnet}
\end{figure}

In this section, we present the results with higher precision from our earlier experiments in a Table~\ref{Results_5} and Table~\ref{Results_20}. Subsequently, we also plot convergence curves to aid better visualizations of findings mentioned earlier in Figure~\ref{fig:metaoptnet}. We also present the results of our models on the Meta-Dataset in Table~\ref{Results_meta}.

\paragraph{Statistical Results} We compare the performance of different models to the Uniform Sampler. All samplers are poorer than the Uniform Sampler and are statistically significant with a confidence interval of 95\%. We use the symbol \textcolor{red}{\dag} to represent the instances where the results are not statistically significant and similar to the performance achieved by the uniform Sampler. We only observe a few cases where a sampler performs significantly better than the Uniform Sampler, which we represent using the symbol \textcolor{green}{\ddag}. To assess statistical significance, we used a paired-difference t-test, with a p-value $p=0.05$.

\begin{figure}[t]
    \centering
    \includegraphics[width=\linewidth]{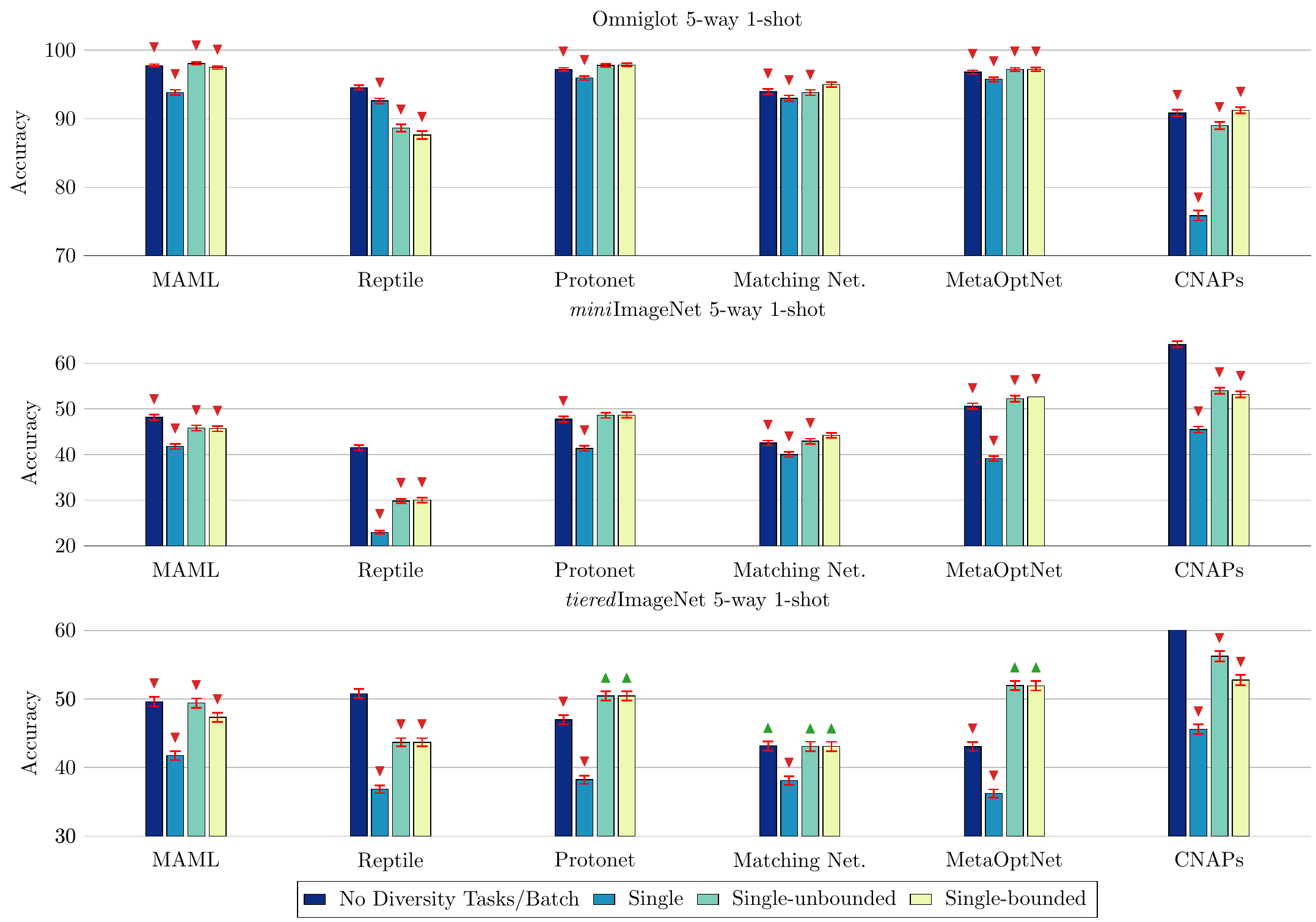}
    \caption{Average accuracy on Omniglot 5-way 1-shot, \textit{mini}ImageNet 5-way 1-shot, \& \textit{tiered}ImageNet 5-way 1-shot with a 95\% confidence interval. We denote all samplers that are worse than the Uniform Sampler and are statistically significant (with a p-value $p=0.05$) with \textcolor{tableau10_C3}{$\blacktriangledown$}, and those that are significantly better than the Uniform Sampler with \textcolor{tableau10_C2}{$\blacktriangle$}.}
    \label{fig:results_sbu_comparison}
\end{figure}

\paragraph{Comparisons with SBU Sampler} Previously, we compared the performance of NDTB and SBU samplers. However, due to the possibility of unfair comparisons, limited by the number of iterations of the SBU sampler, we extend the SBU Sampler to propose two new samplers: (1) \textbf{SBU-unbounded sampler} - This sampler is similar to the SBU sampler but is run for more iterations. However, this sampler allows the model to be trained on more classes than our traditional SBU sampler. Since this remains an unfair comparison, we use this as a higher bound/ideal performance and propose a more appropriate sampler. (2) \textbf{SBU-bounded sampler} - This sampler is again similar to the SBU sampler but ran for more iterations with a bounded pool of tasks, similar to the NDTB sampler or the SBU sampler. Our results from these experiments are presented in Figure~\ref{fig:results_sbu_comparison}. Table \ref{Results_Single} shows the same results with higher precision.

We notice that training for more iterations using samplers such as SBU-unbounded and SBU-bounded leads to a boost in performance. However, we also observe that the NDTB sampler performs better in most cases. With this, we maintain our previous findings. Furthermore, we also notice a peculiar behavior where the SBU-bounded sampler performs better than the SBU-unbounded sampler in most cases. This finding is interesting because the SBU-unbounded sampler has access to more data, and the model being trained should lead to better representations and performance. However, we notice that repeating the tasks and fixing the number of classes achieves the same performance with little to no adverse effects.

\paragraph{\textbf{Comparison with Setlur et. al. \cite{setlur2020support}}} Our goal in this paper has not been to promote a new task sampler but to study its effects. To this end, our goal is different from the works of Setlur et. al, and we are fundamentally different. Furthermore, we believe, we do bring some additional and vital findings which are crucial to the meta-learning community in general:
\begin{itemize}
    \item Against conventional wisdom, increasing task diversity (especially more diverse tasks than Uniform sampling, a setting not studied in \cite{setlur2020support}) does not significantly boost performance.
    \item Furthermore, we also show that the same batch repeated multiple times is sufficient to achieve performances similar to Uniform Sampler. This constraint is even harder than the Setlur constraint, limiting the number of tasks in the pool to a significantly larger number. This has great practical aspects on the application-side where users need not focus on scraping for more data and could instead employ one of our samplers such as NDB, NDTB, etc.
    \item Our empirical analysis is more exhaustive, covering 6 meta-learning methods (as opposed to only being limited to 3 metric-based methods in Setlur), 5 datasets across a wide range of difficulty (e.g., Meta-Dataset, while Setlur focuses on miniImageNet \& CIFAR-FS), and 8 samplers.
    \item  Finally, we point toward an essential question to the readers. If our meta-learning model has seen more information (using Uniform Sampler), why has it not been able to take advantage of this and perform significantly better than those with marginal data (NDTB, NDB, etc.)? This is a severe limitation faced by all meta-learning models, as we have highlighted in our paper. This limitation needs to be addressed to ensure that our model has learned efficiently.
\end{itemize}

\begin{table*}[!htbp]
\centering
\resizebox{0.95\textwidth}{!}{%
\begin{tabular}{c|c|c|c|c|c|c|c}
\hline
\textbf{Dataset} &\textbf{Sampler} & {\textit{\textbf{MAML}}} & {\textit{\textbf{Reptile}}}  & {\textit{\textbf{Protonet}}}  & {\textit{\textbf{Matching Networks}}}  & {\textit{\textbf{MetaOptNet}}}  & {\textit{\textbf{CNAPs}}}\\ \hline
\multirow{8}{*}{\textbf{Omniglot}}
& Uniform Sampler           & \textbf{98.38 {\footnotesize $\pm$ 0.17}} & 94.64 {\footnotesize $\pm$ 0.32} & \textbf{97.82 {\footnotesize $\pm$ 0.23}} & \textbf{94.71 {\footnotesize $\pm$ 0.39}} & \textbf{98.04 {\footnotesize $\pm$ 0.22}} & \textbf{95.01 {\footnotesize $\pm$ 0.35}}\\ \cline{2-8}
& No Diversity Task Sampler & 85.46 {\footnotesize $\pm$ 0.59} & 81.59 {\footnotesize $\pm$ 0.57} & 84.55 {\footnotesize $\pm$ 0.56} & 64.41 {\footnotesize $\pm$ 0.74} & 84.15 {\footnotesize $\pm$ 0.57} & 62.06 {\footnotesize $\pm$ 0.83}\\ \cline{2-8}
& No Diversity Batch Sampler               & 97.17 {\footnotesize $\pm$ 0.25} & 93.83 {\footnotesize $\pm$ 0.34} & 96.67 {\footnotesize $\pm$ 0.27} & 76.10 {\footnotesize $\pm$ 0.65} & 97.11 {\footnotesize $\pm$ 0.26} & 91.07 {\footnotesize $\pm$ 0.46}\\ \cline{2-8}
& No Diversity Tasks per Batch Sampler              & 97.76 {\footnotesize $\pm$ 0.20} & 94.55 {\footnotesize $\pm$ 0.31} \textcolor{red}{\dag} & 97.18 {\footnotesize $\pm$ 0.25} & 93.97 {\footnotesize $\pm$ 0.40} & 96.80 {\footnotesize $\pm$ 0.27} & 90.84 {\footnotesize $\pm$ 0.47}\\ \cline{2-8}
& Single Batch Uniform Sampler        & 93.84 {\footnotesize $\pm$ 0.37} & 92.60 {\footnotesize $\pm$ 0.38} & 95.95 {\footnotesize $\pm$ 0.31} & 92.98 {\footnotesize $\pm$ 0.44} & 95.76 {\footnotesize $\pm$ 0.31} & 75.86 {\footnotesize $\pm$ 0.73}\\ \cline{2-8}
& OHTM Sampler              & 97.74 {\footnotesize $\pm$ 0.20} & 93.89 {\footnotesize $\pm$ 0.34} & 97.22 {\footnotesize $\pm$ 0.25} & 93.48 {\footnotesize $\pm$ 0.43} & 96.12 {\footnotesize $\pm$ 0.29} & 91.51 {\footnotesize $\pm$ 0.47} \\ \cline{2-8}
& s-DPP Sampler            & 97.61 {\footnotesize $\pm$ 0.21} & \textbf{94.79 {\footnotesize $\pm$ 0.30}} \textcolor{red}{\dag} & 97.22 {\footnotesize $\pm$ 0.24} & 92.29 {\footnotesize $\pm$ 0.44} & 95.83 {\footnotesize $\pm$ 0.30} & 95.00 {\footnotesize $\pm$ 0.33} \textcolor{red}{\dag} \\ \cline{2-8}
& d-DPP Sampler            & 97.69 {\footnotesize $\pm$ 0.21} & 94.25 {\footnotesize $\pm$ 0.33} & 97.28 {\footnotesize $\pm$ 0.24} & 93.71 {\footnotesize $\pm$ 0.40} & 95.59 {\footnotesize $\pm$ 0.30} & 94.84 {\footnotesize $\pm$ 0.34} \textcolor{red}{\dag} \\ \hline \hline

\multirow{8}{*}{\textbf{MiniImagenet}}
& Uniform Sampler           & \textbf{48.86 {\footnotesize $\pm$ 0.62}} & 41.42 {\footnotesize $\pm$ 0.56} & \textbf{48.56 {\footnotesize $\pm$ 0.60}} & \textbf{43.84 {\footnotesize $\pm$ 0.58}} & \textbf{55.02 {\footnotesize $\pm$ 0.66}} & \textbf{64.48 {\footnotesize $\pm$ 0.71}}\\ \cline{2-8}
& No Diversity Task Sampler               & 36.70 {\footnotesize $\pm$ 0.53} & 32.38 {\footnotesize $\pm$ 0.48} & 37.83 {\footnotesize $\pm$ 0.53} & 35.08 {\footnotesize $\pm$ 0.53} & 36.62 {\footnotesize $\pm$ 0.55} & 46.51 {\footnotesize $\pm$ 0.63}\\ \cline{2-8}
& No Diversity Batch Sampler               & 48.78 {\footnotesize $\pm$ 0.60} \textcolor{red}{\dag} & 40.80 {\footnotesize $\pm$ 0.54} & 47.32 {\footnotesize $\pm$ 0.62} & 42.15 {\footnotesize $\pm$ 0.58} & 53.50 {\footnotesize $\pm$ 0.63} & 60.92 {\footnotesize $\pm$ 0.68}\\ \cline{2-8}
& No Diversity Tasks per Batch Sampler              & 48.17 {\footnotesize $\pm$ 0.62} & \textbf{41.49 {\footnotesize $\pm$ 0.56}} \textcolor{red}{\dag} & 47.73 {\footnotesize $\pm$ 0.60} & 42.54 {\footnotesize $\pm$ 0.53} & 50.60 {\footnotesize $\pm$ 0.62} & 64.11 {\footnotesize $\pm$ 0.68} \textcolor{red}{\dag} \\ \cline{2-8}
& Single Batch Uniform Sampler        & 41.76 {\footnotesize $\pm$ 0.56} & 22.96 {\footnotesize $\pm$ 0.33} & 41.35 {\footnotesize $\pm$ 0.56} & 40.00 {\footnotesize $\pm$ 0.54} & 39.10 {\footnotesize $\pm$ 0.54} & 45.47 {\footnotesize $\pm$ 0.67}\\ \cline{2-8}
& OHTM Sampler              & 48.30 {\footnotesize $\pm$ 0.58} & 40.44 {\footnotesize $\pm$ 0.54} & 47.45 {\footnotesize $\pm$ 0.59} & 43.05 {\footnotesize $\pm$ 0.55} & 47.11 {\footnotesize $\pm$ 0.58} & 59.62 {\footnotesize $\pm$ 0.69}\\ \cline{2-8}
& s-DPP Sampler            & 48.14 {\footnotesize $\pm$ 0.59} & 40.19 {\footnotesize $\pm$ 0.56} & 47.22 {\footnotesize $\pm$ 0.58} & 42.66 {\footnotesize $\pm$ 0.56} & 52.74 {\footnotesize $\pm$ 0.63} & 63.26 {\footnotesize $\pm$ 0.69}\\ \cline{2-8}
& d-DPP Sampler             & 48.99 {\footnotesize $\pm$ 0.60} & 40.40 {\footnotesize $\pm$ 0.54} & 46.73 {\footnotesize $\pm$ 0.60} & 42.37 {\footnotesize $\pm$ 0.56} & 48.26 {\footnotesize $\pm$ 0.60} & 61.44 {\footnotesize $\pm$ 0.67}\\ \hline \hline

\multirow{8}{*}{\textbf{Tiered-Imagenet}} & Uniform Sampler & \textbf{51.89 {\footnotesize $\pm$ 0.68}} & 50.35 {\footnotesize $\pm$ 0.69} & 49.18 {\footnotesize $\pm$ 0.68} & 42.18 {\footnotesize $\pm$ 0.66} & 49.51 {\footnotesize $\pm$ 0.67} & \textbf{65.16 {\footnotesize $\pm$ 0.75}}\\ \cline{2-8}
& No Diversity Task Sampler               & 34.13 {\footnotesize $\pm$ 0.51} & 34.94 {\footnotesize $\pm$ 0.54} & 34.01 {\footnotesize $\pm$ 0.54} & 32.01 {\footnotesize $\pm$ 0.51} & 37.12 {\footnotesize $\pm$ 0.59} & 45.40 {\footnotesize $\pm$ 0.68}\\ \cline{2-8}
& No Diversity Batch Sampler              & 50.40 {\footnotesize $\pm$ 0.68} & 48.80 {\footnotesize $\pm$ 0.67} & 48.09 {\footnotesize $\pm$ 0.67} & 42.03 {\footnotesize $\pm$ 0.67} \textcolor{red}{\dag} & 48.47 {\footnotesize $\pm$ 0.67} & 60.13 {\footnotesize $\pm$ 0.75}\\ \cline{2-8}
& No Diversity Tasks per Batch Sampler    & 49.62 {\footnotesize $\pm$ 0.69} & \textbf{50.77 {\footnotesize $\pm$ 0.67}}\textcolor{red}{\dag} & 46.96 {\footnotesize $\pm$ 0.67} & 43.15 {\footnotesize $\pm$ 0.66} \textcolor{green}{\ddag} & 43.05 {\footnotesize $\pm$ 0.63} & 62.81 {\footnotesize $\pm$ 0.73}\\ \cline{2-8}
& Single Batch Uniform Sampler            & 41.74 {\footnotesize $\pm$ 0.66} & 36.82 {\footnotesize $\pm$ 0.56} & 38.22 {\footnotesize $\pm$ 0.62} & 38.08 {\footnotesize $\pm$ 0.61} & 36.21 {\footnotesize $\pm$ 0.60} & 45.56 {\footnotesize $\pm$ 0.72}\\ \cline{2-8}
& OHTM Sampler                            & 49.00 {\footnotesize $\pm$ 0.67} & 50.06 {\footnotesize $\pm$ 0.67} \textcolor{red}{\dag} & 45.29 {\footnotesize $\pm$ 0.66} & 41.56 {\footnotesize $\pm$ 0.65} & 46.14 {\footnotesize $\pm$ 0.62} & 59.70 {\footnotesize $\pm$ 0.74}\\ \cline{2-8}
& s-DPP Sampler                           & 51.15 {\footnotesize $\pm$ 0.66} & 50.50 {\footnotesize $\pm$ 0.68} \textcolor{red}{\dag} & \textbf{49.42 {\footnotesize $\pm$ 0.68}} \textcolor{red}{\dag} & \textbf{44.08 {\footnotesize $\pm$ 0.68}} \textcolor{green}{\ddag} & \textbf{49.65 {\footnotesize $\pm$ 0.68}} \textcolor{red}{\dag} & 63.88 {\footnotesize $\pm$ 0.75}\\ \cline{2-8}
& d-DPP Sampler                           & 51.68 {\footnotesize $\pm$ 0.67} \textcolor{red}{\dag} & 50.81 {\footnotesize $\pm$ 0.67} \textcolor{green}{\ddag} & 49.35 {\footnotesize $\pm$ 0.69} \textcolor{red}{\dag} & 42.50 {\footnotesize $\pm$ 0.68} \textcolor{red}{\dag} & 49.26 {\footnotesize $\pm$ 0.67} & 64.80 {\footnotesize $\pm$ 0.75} \textcolor{red}{\dag}\\ \hline \hline

\end{tabular}%
}
\caption{Performance metric of our models on different task samplers in the 5-way 1-shot setting.}
\label{Results_5}
\end{table*}

\begin{table*}[!htbp]
\centering
\resizebox{0.95\textwidth}{!}{%
\begin{tabular}{c|c|c|c|c|c|c|c}
\hline
\textbf{Dataset} &\textbf{Sampler} & {\textit{\textbf{MAML}}} & {\textit{\textbf{Reptile}}}  & {\textit{\textbf{Protonet}}}  & {\textit{\textbf{Matching Networks}}}  & {\textit{\textbf{MetaOptNet}}}  & {\textit{\textbf{CNAPs}}}\\ \hline

\multirow{8}{*}{\textbf{Omniglot}} 
& Uniform Sampler & \textbf{91.28 {\footnotesize $\pm$ 0.22}} & 90.09 {\footnotesize $\pm$ 0.22} & 93.72 {\footnotesize $\pm$ 0.20} & \textbf{74.62 {\footnotesize $\pm$ 0.38}} & 90.20 {\footnotesize $\pm$ 0.23} & \textbf{92.09 {\footnotesize $\pm$ 0.22}}\\ \cline{2-8}
& No Diversity Task Sampler               & 83.39 {\footnotesize $\pm$ 0.29} & 59.49 {\footnotesize $\pm$ 0.33} & 85.84 {\footnotesize $\pm$ 0.27} & 26.50 {\footnotesize $\pm$ 0.32} & 88.40 {\footnotesize $\pm$ 0.26} & 73.82 {\footnotesize $\pm$ 0.39}\\ \cline{2-8}
& No Diversity Batch Sampler              & 89.07 {\footnotesize $\pm$ 0.25} & 88.23 {\footnotesize $\pm$ 0.23} & 93.18 {\footnotesize $\pm$ 0.20} & 71.77 {\footnotesize $\pm$ 0.38} & 91.24 {\footnotesize $\pm$ 0.22} \textcolor{green}{\ddag} & 89.56 {\footnotesize $\pm$ 0.24}\\ \cline{2-8}
& No Diversity Tasks per Batch Sampler    & 90.77 {\footnotesize $\pm$ 0.23} & \textbf{91.15 {\footnotesize $\pm$ 0.21}}\textcolor{green}{\ddag} & \textbf{93.85 {\footnotesize $\pm$ 0.19}} \textcolor{red}{\dag} & 61.31 {\footnotesize $\pm$ 0.41} & 89.59 {\footnotesize $\pm$ 0.24} & 89.99 {\footnotesize $\pm$ 0.24}\\ \cline{2-8}
& Single Batch Uniform Sampler            & 82.45 {\footnotesize $\pm$ 0.31} & 80.89 {\footnotesize $\pm$ 0.27} & 92.67 {\footnotesize $\pm$ 0.20} & 54.01 {\footnotesize $\pm$ 0.40} & 70.81 {\footnotesize $\pm$ 0.35} & 77.54 {\footnotesize $\pm$ 0.37}\\ \cline{2-8}
& OHTM Sampler                            & 91.25 {\footnotesize $\pm$ 0.22} \textcolor{red}{\dag} & 89.92 {\footnotesize $\pm$ 0.22} & 93.33 {\footnotesize $\pm$ 0.20} & 72.20 {\footnotesize $\pm$ 0.38} & \textbf{91.56 {\footnotesize $\pm$ 0.23}} \textcolor{green}{\ddag} & 89.51 {\footnotesize $\pm$ 0.25} \\ \cline{2-8}
& s-DPP Sampler                           & 88.79 {\footnotesize $\pm$ 0.24} & 85.40 {\footnotesize $\pm$ 0.25} & 90.90 {\footnotesize $\pm$ 0.22} & 72.86 {\footnotesize $\pm$ 0.37} & 91.47 {\footnotesize $\pm$ 0.22} \textcolor{green}{\ddag} & 90.98 {\footnotesize $\pm$ 0.22}\\ \cline{2-8}
& d-DPP Sampler                           & 85.36 {\footnotesize $\pm$ 0.30} & 85.60 {\footnotesize $\pm$ 0.25} & 91.74 {\footnotesize $\pm$ 0.22} & 85.36 {\footnotesize $\pm$ 0.30} \textcolor{green}{\ddag} & 90.40 {\footnotesize $\pm$ 0.24} \textcolor{red}{\dag} & 91.95 {\footnotesize $\pm$ 0.22} \textcolor{red}{\dag}\\ \hline \hline

\end{tabular}%
}
\caption{Performance metric of our models on different task samplers in the 20-way 1-shot setting.}
\label{Results_20}
\end{table*}

\begin{table*}[!htbp]
\centering
\resizebox{0.95\textwidth}{!}{%
\begin{tabular}{c|c|c|c|c|c|c|c}
\hline
\textbf{Dataset} &\textbf{Sampler} & {\textit{\textbf{MAML}}} & {\textit{\textbf{Reptile}}}  & {\textit{\textbf{Protonet}}}  & {\textit{\textbf{Matching Networks}}}  & {\textit{\textbf{MetaOptNet}}}  & {\textit{\textbf{CNAPs}}}\\ \hline

\multirow{6}{*}{\textbf{Meta-Dataset (ILSVRC)}} & Uniform Sampler & 22.24 {\footnotesize $\pm$ 0.33} & 31.32 {\footnotesize $\pm$ 0.47} & 25.41 {\footnotesize $\pm$ 0.35} & 27.52 {\footnotesize $\pm$ 0.45} & 28.74 {\footnotesize $\pm$ 0.43} & 40.11 {\footnotesize $\pm$ 0.59}\\ \cline{2-8}
& No Diversity Task Sampler               & 23.02 {\footnotesize $\pm$ 0.35} \textcolor{green}{\ddag} & 25.43 {\footnotesize $\pm$ 0.40} & 26.68 {\footnotesize $\pm$ 0.44} \textcolor{green}{\ddag} & 24.32 {\footnotesize $\pm$ 0.37} & 23.85 {\footnotesize $\pm$ 0.36} & \textbf{42.23 {\footnotesize $\pm$ 0.64}} \textcolor{green}{\ddag}\\ \cline{2-8}
& No Diversity Batch Sampler              & \textbf{23.24 {\footnotesize $\pm$ 0.36}} \textcolor{green}{\ddag} & 28.70 {\footnotesize $\pm$ 0.44} & 24.71 {\footnotesize $\pm$ 0.34} & 24.25 {\footnotesize $\pm$ 0.37} & 25.64 {\footnotesize $\pm$ 0.40} & 34.07 {\footnotesize $\pm$ 0.52}\\ \cline{2-8}
& No Diversity Tasks per Batch Sampler    & 23.14 {\footnotesize $\pm$ 0.37} \textcolor{green}{\ddag} & 31.25 {\footnotesize $\pm$ 0.51} \textcolor{red}{\dag} & 27.04 {\footnotesize $\pm$ 0.42} \textcolor{green}{\ddag} & 26.47 {\footnotesize $\pm$ 0.44} & 27.01 {\footnotesize $\pm$ 0.43} & 33.00 {\footnotesize $\pm$ 0.53}\\ \cline{2-8}
& Single Batch Uniform Sampler            & 22.99 {\footnotesize $\pm$ 0.34} \textcolor{green}{\ddag} & 26.43 {\footnotesize $\pm$ 0.41} & \textbf{27.05 {\footnotesize $\pm$ 0.42}} \textcolor{green}{\ddag} & \textbf{27.79 {\footnotesize $\pm$ 0.44}} \textcolor{red}{\dag} & 28.00 {\footnotesize $\pm$ 0.45} & 26.18 {\footnotesize $\pm$ 0.44}\\ \cline{2-8}
& OHTM Sampler                            & 22.66 {\footnotesize $\pm$ 0.35} \textcolor{red}{\dag} & \textbf{31.62 {\footnotesize $\pm$ 0.48}} \textcolor{green}{\ddag} & 26.48 {\footnotesize $\pm$ 0.39} \textcolor{green}{\ddag} & 27.05 {\footnotesize $\pm$ 0.43} \textcolor{red}{\dag} & \textbf{29.03 {\footnotesize $\pm$ 0.44}} \textcolor{red}{\dag} & 38.12 {\footnotesize $\pm$ 0.57} \\ \hline \hline

\multirow{6}{*}{\textbf{Meta-Dataset (Omniglot)}} & Uniform Sampler & 30.93 {\footnotesize $\pm$ 0.69} & 86.07 {\footnotesize $\pm$ 0.50} & \textbf{77.77 {\footnotesize $\pm$ 0.65}} & \textbf{77.61 {\footnotesize $\pm$ 0.64}} & \textbf{81.82 {\footnotesize $\pm$ 0.60}} & \textbf{87.08 {\footnotesize $\pm$ 0.52}}\\ \cline{2-8}
& No Diversity Task Sampler               & 31.17 {\footnotesize $\pm$ 0.65} \textcolor{red}{\dag} & 41.36 {\footnotesize $\pm$ 0.74} & 61.26 {\footnotesize $\pm$ 0.73} & 60.59 {\footnotesize $\pm$ 0.69} & 73.67 {\footnotesize $\pm$ 0.65} & 60.18 {\footnotesize $\pm$ 0.73}\\ \cline{2-8}
& No Diversity Batch Sampler              & \textbf{32.70 {\footnotesize $\pm$ 0.75}} \textcolor{green}{\ddag} & 72.13 {\footnotesize $\pm$ 0.66} & 70.86 {\footnotesize $\pm$ 0.67} & 56.61 {\footnotesize $\pm$ 0.71} & 73.27 {\footnotesize $\pm$ 0.64} & 73.69 {\footnotesize $\pm$ 0.69}\\ \cline{2-8}
& No Diversity Tasks per Batch Sampler    & 30.36 {\footnotesize $\pm$ 0.71} \textcolor{green}{\ddag} & \textbf{86.29 {\footnotesize $\pm$ 0.49}} \textcolor{red}{\dag} & 72.33 {\footnotesize $\pm$ 0.69} & 72.96 {\footnotesize $\pm$ 0.68} & 63.17 {\footnotesize $\pm$ 0.70} & 56.14 {\footnotesize $\pm$ 0.72}\\ \cline{2-8}
& Single Batch Uniform Sampler            & 30.21 {\footnotesize $\pm$ 0.74} \textcolor{red}{\dag} & 75.39 {\footnotesize $\pm$ 0.66} & 63.52 {\footnotesize $\pm$ 0.69} & 72.95 {\footnotesize $\pm$ 0.71} & 73.44 {\footnotesize $\pm$ 0.66} & 42.72 {\footnotesize $\pm$ 0.74}\\ \cline{2-8}
& OHTM Sampler                            & 31.91 {\footnotesize $\pm$ 0.70} \textcolor{green}{\ddag} & 83.42 {\footnotesize $\pm$ 0.53} & 76.55 {\footnotesize $\pm$ 0.65} & 74.93 {\footnotesize $\pm$ 0.66} & 74.56 {\footnotesize $\pm$ 0.66} & 78.00 {\footnotesize $\pm$ 0.69} \\ \hline \hline

\multirow{6}{*}{\textbf{Meta-Dataset (Aircraft)}} & Uniform Sampler & 22.52 {\footnotesize $\pm$ 0.34} & 40.62 {\footnotesize $\pm$ 0.54} & 28.34 {\footnotesize $\pm$ 0.41} & 27.93 {\footnotesize $\pm$ 0.41} & \textbf{32.12 {\footnotesize $\pm$ 0.47}} & \textbf{38.64 {\footnotesize $\pm$ 0.54}}\\ \cline{2-8}
& No Diversity Task Sampler               & 22.76 {\footnotesize $\pm$ 0.32} \textcolor{red}{\dag} & 24.55 {\footnotesize $\pm$ 0.35} & 26.77 {\footnotesize $\pm$ 0.36} & 22.58 {\footnotesize $\pm$ 0.30} & 24.73 {\footnotesize $\pm$ 0.36} & 30.37 {\footnotesize $\pm$ 0.40}\\ \cline{2-8}
& No Diversity Batch Sampler              & 22.32 {\footnotesize $\pm$ 0.31} \textcolor{red}{\dag} & 33.11 {\footnotesize $\pm$ 0.48} & \textbf{28.57 {\footnotesize $\pm$ 0.40}} \textcolor{red}{\dag} & 23.63 {\footnotesize $\pm$ 0.33} & 25.76 {\footnotesize $\pm$ 0.36} & 28.84 {\footnotesize $\pm$ 0.39}\\ \cline{2-8}
& No Diversity Tasks per Batch Sampler    & 23.00 {\footnotesize $\pm$ 0.33} \textcolor{green}{\ddag} & 40.59 {\footnotesize $\pm$ 0.52} \textcolor{red}{\dag} & 28.32 {\footnotesize $\pm$ 0.41} \textcolor{red}{\dag} & 27.98 {\footnotesize $\pm$ 0.42} \textcolor{red}{\dag} & 27.13 {\footnotesize $\pm$ 0.40} & 26.09 {\footnotesize $\pm$ 0.36}\\ \cline{2-8}
& Single Batch Uniform Sampler            & 22.40 {\footnotesize $\pm$ 0.31} \textcolor{red}{\dag} & 29.19 {\footnotesize $\pm$ 0.40} & 28.51 {\footnotesize $\pm$ 0.40} \textcolor{red}{\dag} & 26.47 {\footnotesize $\pm$ 0.39} & 27.53 {\footnotesize $\pm$ 0.42} & 21.86 {\footnotesize $\pm$ 0.26}\\ \cline{2-8}
& OHTM Sampler                            & \textbf{23.34 {\footnotesize $\pm$ 0.32}} \textcolor{green}{\ddag} & 38.73 {\footnotesize $\pm$ 0.54} & 27.99 {\footnotesize $\pm$ 0.38} \textcolor{red}{\dag} & \textbf{29.53 {\footnotesize $\pm$ 0.45}} \textcolor{green}{\ddag}& 31.08 {\footnotesize $\pm$ 0.47} & 35.96 {\footnotesize $\pm$ 0.50} \\ \hline \hline

\multirow{6}{*}{\textbf{Meta-Dataset (Birds)}} & Uniform Sampler & 22.30 {\footnotesize $\pm$ 0.32} & \textbf{47.79 {\footnotesize $\pm$ 0.61}} & 29.84 {\footnotesize $\pm$ 0.42} & 34.36 {\footnotesize $\pm$ 0.53} & 33.70 {\footnotesize $\pm$ 0.48} & \textbf{47.43 {\footnotesize $\pm$ 0.62}}\\ \cline{2-8}
& No Diversity Task Sampler               & 23.13 {\footnotesize $\pm$ 0.33} \textcolor{green}{\ddag} & 27.25 {\footnotesize $\pm$ 0.38} & 27.08 {\footnotesize $\pm$ 0.38} & 25.95 {\footnotesize $\pm$ 0.39} & 26.78 {\footnotesize $\pm$ 0.41} & 40.42 {\footnotesize $\pm$ 0.61}\\ \cline{2-8}
& No Diversity Batch Sampler              & \textbf{25.31 {\footnotesize $\pm$ 0.41}} \textcolor{green}{\ddag} & 36.66 {\footnotesize $\pm$ 0.51} & 28.47 {\footnotesize $\pm$ 0.38} & 26.91 {\footnotesize $\pm$ 0.41} & 28.55 {\footnotesize $\pm$ 0.40} & 31.95 {\footnotesize $\pm$ 0.48}\\ \cline{2-8}
& No Diversity Tasks per Batch Sampler    & 23.93 {\footnotesize $\pm$ 0.37} \textcolor{green}{\ddag} & 45.36 {\footnotesize $\pm$ 0.56} & 29.48 {\footnotesize $\pm$ 0.43} \textcolor{red}{\dag} & 32.11 {\footnotesize $\pm$ 0.48} & 27.86 {\footnotesize $\pm$ 0.41} & 31.28 {\footnotesize $\pm$ 0.47}\\ \cline{2-8}
& Single Batch Uniform Sampler            & 23.69 {\footnotesize $\pm$ 0.36} \textcolor{green}{\ddag} & 33.70 {\footnotesize $\pm$ 0.47} & 30.40 {\footnotesize $\pm$ 0.46} \textcolor{red}{\dag} & 32.39 {\footnotesize $\pm$ 0.50} & 30.80 {\footnotesize $\pm$ 0.44} & 25.13 {\footnotesize $\pm$ 0.38}\\ \cline{2-8}
& OHTM Sampler                            & 23.22 {\footnotesize $\pm$ 0.35} \textcolor{green}{\ddag} & 46.38 {\footnotesize $\pm$ 0.61} & \textbf{30.83 {\footnotesize $\pm$ 0.43}} \textcolor{green}{\ddag} & \textbf{35.17 {\footnotesize $\pm$ 0.54}} \textcolor{green}{\ddag} & \textbf{37.18 {\footnotesize $\pm$ 0.50}} \textcolor{green}{\ddag} & 43.77 {\footnotesize $\pm$ 0.59} \\ \hline \hline

\multirow{6}{*}{\textbf{Meta-Dataset (Textures)}} & Uniform Sampler & 22.51 {\footnotesize $\pm$ 0.34} & \textbf{33.33 {\footnotesize $\pm$ 0.48}} & 26.63 {\footnotesize $\pm$ 0.37} & 28.74 {\footnotesize $\pm$ 0.42} & 27.44 {\footnotesize $\pm$ 0.40} & \textbf{38.10 {\footnotesize $\pm$ 0.50}}\\ \cline{2-8}
& No Diversity Task Sampler               & 22.46 {\footnotesize $\pm$ 0.35} \textcolor{red}{\dag} & 24.93 {\footnotesize $\pm$ 0.37} & 24.95 {\footnotesize $\pm$ 0.37} & 23.58 {\footnotesize $\pm$ 0.35} & 23.80 {\footnotesize $\pm$ 0.34} & 33.11 {\footnotesize $\pm$ 0.51}\\ \cline{2-8}
& No Diversity Batch Sampler              & 22.32 {\footnotesize $\pm$ 0.32} \textcolor{red}{\dag} & 29.40 {\footnotesize $\pm$ 0.44} & 26.80 {\footnotesize $\pm$ 0.37} \textcolor{red}{\dag} & 23.84 {\footnotesize $\pm$ 0.34} & 24.95 {\footnotesize $\pm$ 0.35} & 32.46 {\footnotesize $\pm$ 0.49}\\ \cline{2-8}
& No Diversity Tasks per Batch Sampler    & 22.69 {\footnotesize $\pm$ 0.35} \textcolor{red}{\dag} & 31.94 {\footnotesize $\pm$ 0.45} & 26.53 {\footnotesize $\pm$ 0.37} \textcolor{red}{\dag} & 25.65 {\footnotesize $\pm$ 0.37} & 23.99 {\footnotesize $\pm$ 0.34} & 31.77 {\footnotesize $\pm$ 0.44}\\ \cline{2-8}
& Single Batch Uniform Sampler            & 21.51 {\footnotesize $\pm$ 0.29} & 26.97 {\footnotesize $\pm$ 0.39} & 27.08 {\footnotesize $\pm$ 0.37} \textcolor{green}{\ddag}& 25.52 {\footnotesize $\pm$ 0.38} & 26.29 {\footnotesize $\pm$ 0.38} & 22.93 {\footnotesize $\pm$ 0.35}\\ \cline{2-8}
& OHTM Sampler                            & \textbf{22.94 {\footnotesize $\pm$ 0.36}} \textcolor{red}{\dag} & 31.67 {\footnotesize $\pm$ 0.48} & \textbf{28.33 {\footnotesize $\pm$ 0.40}} \textcolor{green}{\ddag} & \textbf{29.36 {\footnotesize $\pm$ 0.42}} \textcolor{green}{\ddag} & \textbf{28.34 {\footnotesize $\pm$ 0.41}} \textcolor{green}{\ddag} & 34.90 {\footnotesize $\pm$ 0.46} \\ \hline \hline

\multirow{6}{*}{\textbf{Meta-Dataset (Quick Draw)}} & Uniform Sampler & 34.84 {\footnotesize $\pm$ 0.63} & 55.31 {\footnotesize $\pm$ 0.69} & \textbf{51.80 {\footnotesize $\pm$ 0.67}} & \textbf{52.61 {\footnotesize $\pm$ 0.68}} & \textbf{56.17 {\footnotesize $\pm$ 0.67}} & \textbf{58.54 {\footnotesize $\pm$ 0.70}}\\ \cline{2-8}
& No Diversity Task Sampler               & 34.95 {\footnotesize $\pm$ 0.62} \textcolor{red}{\dag} & 37.59 {\footnotesize $\pm$ 0.58} & 48.94 {\footnotesize $\pm$ 0.64} & 42.82 {\footnotesize $\pm$ 0.62} & 51.42 {\footnotesize $\pm$ 0.68} & 44.04 {\footnotesize $\pm$ 0.64}\\ \cline{2-8}
& No Diversity Batch Sampler              & 35.45 {\footnotesize $\pm$ 0.63} \textcolor{red}{\dag} & 46.98 {\footnotesize $\pm$ 0.64} & 44.28 {\footnotesize $\pm$ 0.63} & 36.60 {\footnotesize $\pm$ 0.55} & 48.00 {\footnotesize $\pm$ 0.63} & 47.53 {\footnotesize $\pm$ 0.68}\\ \cline{2-8}
& No Diversity Tasks per Batch Sampler    & 35.31 {\footnotesize $\pm$ 0.61} \textcolor{red}{\dag} & \textbf{55.49 {\footnotesize $\pm$ 0.67}} & 50.77 {\footnotesize $\pm$ 0.65} & 47.41 {\footnotesize $\pm$ 0.67} & 48.96 {\footnotesize $\pm$ 0.63} & 41.24 {\footnotesize $\pm$ 0.63}\\ \cline{2-8}
& Single Batch Uniform Sampler            & \textbf{36.58 {\footnotesize $\pm$ 0.62}} \textcolor{green}{\ddag} & 50.10 {\footnotesize $\pm$ 0.64} & 48.52 {\footnotesize $\pm$ 0.66} & 48.02 {\footnotesize $\pm$ 0.68} & 51.96 {\footnotesize $\pm$ 0.67} & 29.29 {\footnotesize $\pm$ 0.48}\\ \cline{2-8}
& OHTM Sampler                            & 35.13 {\footnotesize $\pm$ 0.61} \textcolor{red}{\dag} & 53.42 {\footnotesize $\pm$ 0.68} & 49.90 {\footnotesize $\pm$ 0.69} & 50.59 {\footnotesize $\pm$ 0.68} & 54.11 {\footnotesize $\pm$ 0.65} & 51.99 {\footnotesize $\pm$ 0.68} \\ \hline \hline

\multirow{6}{*}{\textbf{Meta-Dataset (Fungi)}} & Uniform Sampler & 23.04 {\footnotesize $\pm$ 0.36} & \textbf{39.78 {\footnotesize $\pm$ 0.57}} & 29.43 {\footnotesize $\pm$ 0.42} & 34.45 {\footnotesize $\pm$ 0.54} & 34.65 {\footnotesize $\pm$ 0.53} & 40.37 {\footnotesize $\pm$ 0.58}\\ \cline{2-8}
& No Diversity Task Sampler               & 23.53 {\footnotesize $\pm$ 0.36} \textcolor{red}{\dag} & 29.22 {\footnotesize $\pm$ 0.45} & 27.69 {\footnotesize $\pm$ 0.40} & 24.57 {\footnotesize $\pm$ 0.35} & 27.25 {\footnotesize $\pm$ 0.43} & 35.26 {\footnotesize $\pm$ 0.50}\\ \cline{2-8}
& No Diversity Batch Sampler              & \textbf{25.05 {\footnotesize $\pm$ 0.41}} \textcolor{green}{\ddag} & 34.84 {\footnotesize $\pm$ 0.54} & 28.43 {\footnotesize $\pm$ 0.42} & 25.97 {\footnotesize $\pm$ 0.41} & 30.46 {\footnotesize $\pm$ 0.50} & 30.75 {\footnotesize $\pm$ 0.46}\\ \cline{2-8}
& No Diversity Tasks per Batch Sampler    & 24.47 {\footnotesize $\pm$ 0.41} \textcolor{green}{\ddag} & 38.80 {\footnotesize $\pm$ 0.59} & 30.38 {\footnotesize $\pm$ 0.46} \textcolor{green}{\ddag} & 32.03 {\footnotesize $\pm$ 0.54} & 28.82 {\footnotesize $\pm$ 0.44} & 31.26 {\footnotesize $\pm$ 0.47}\\ \cline{2-8}
& Single Batch Uniform Sampler            & 23.28 {\footnotesize $\pm$ 0.36} \textcolor{red}{\dag} & 31.28 {\footnotesize $\pm$ 0.49} & 30.12 {\footnotesize $\pm$ 0.46} \textcolor{green}{\ddag} & 31.94 {\footnotesize $\pm$ 0.51} & 30.49 {\footnotesize $\pm$ 0.48} & 25.90 {\footnotesize $\pm$ 0.43}\\ \cline{2-8}
& OHTM Sampler                            & 23.88 {\footnotesize $\pm$ 0.37} \textcolor{green}{\ddag} & 34.84 {\footnotesize $\pm$ 0.54} & \textbf{31.13 {\footnotesize $\pm$ 0.47}} \textcolor{green}{\ddag} & \textbf{35.63 {\footnotesize $\pm$ 0.56}} \textcolor{green}{\ddag} & \textbf{36.40 {\footnotesize $\pm$ 0.56}} \textcolor{green}{\ddag} & \textbf{41.85 {\footnotesize $\pm$ 0.57}} \textcolor{green}{\ddag}\\ \hline \hline

\multirow{6}{*}{\textbf{Meta-Dataset (VGG Flower)}} & Uniform Sampler & 30.28 {\footnotesize $\pm$ 0.51} & 65.76 {\footnotesize $\pm$ 0.64} & 50.98 {\footnotesize $\pm$ 0.55} & \textbf{58.64 {\footnotesize $\pm$ 0.71}} & 58.28 {\footnotesize $\pm$ 0.67} & \textbf{65.12 {\footnotesize $\pm$ 0.68}}\\ \cline{2-8}
& No Diversity Task Sampler               & 31.19 {\footnotesize $\pm$ 0.49} \textcolor{green}{\ddag} & 49.91 {\footnotesize $\pm$ 0.59} & 45.36 {\footnotesize $\pm$ 0.57} & 45.50 {\footnotesize $\pm$ 0.64} & 37.46 {\footnotesize $\pm$ 0.56} & 52.21 {\footnotesize $\pm$ 0.65}\\ \cline{2-8}
& No Diversity Batch Sampler              & \textbf{35.93 {\footnotesize $\pm$ 0.52}} \textcolor{green}{\ddag} & 60.13 {\footnotesize $\pm$ 0.66} & 49.97 {\footnotesize $\pm$ 0.59} & 44.95 {\footnotesize $\pm$ 0.56} & 53.61 {\footnotesize $\pm$ 0.64} & 41.95 {\footnotesize $\pm$ 0.64}\\ \cline{2-8}
& No Diversity Tasks per Batch Sampler    & 34.11 {\footnotesize $\pm$ 0.55} \textcolor{green}{\ddag} & \textbf{66.35 {\footnotesize $\pm$ 0.66}} \textcolor{red}{\dag} & 53.77 {\footnotesize $\pm$ 0.61} \textcolor{green}{\ddag} & 57.27 {\footnotesize $\pm$ 0.67} & 50.29 {\footnotesize $\pm$ 0.59} & 43.91 {\footnotesize $\pm$ 0.60}\\ \cline{2-8}
& Single Batch Uniform Sampler            & 26.84 {\footnotesize $\pm$ 0.46} & 53.39 {\footnotesize $\pm$ 0.63} & \textbf{51.14 {\footnotesize $\pm$ 0.59}} \textcolor{red}{\dag} & 54.32 {\footnotesize $\pm$ 0.67} & 52.84 {\footnotesize $\pm$ 0.61} & 27.16 {\footnotesize $\pm$ 0.44}\\ \cline{2-8}
& OHTM Sampler                            & 32.69 {\footnotesize $\pm$ 0.52} \textcolor{green}{\ddag} & 64.39 {\footnotesize $\pm$ 0.67} & 50.55 {\footnotesize $\pm$ 0.56} \textcolor{red}{\dag} & 56.07 {\footnotesize $\pm$ 0.67} & \textbf{58.33 {\footnotesize $\pm$ 0.68}} \textcolor{red}{\dag} & 62.51 {\footnotesize $\pm$ 0.68} \\ \hline \hline

\multirow{6}{*}{\textbf{Meta-Dataset (Traffic Signs)}} & Uniform Sampler & 24.53 {\footnotesize $\pm$ 0.39} & \textbf{48.85 {\footnotesize $\pm$ 0.67}} & \textbf{38.97 {\footnotesize $\pm$ 0.57}} & \textbf{39.59 {\footnotesize $\pm$ 0.63}} & 39.23 {\footnotesize $\pm$ 0.58} & \textbf{52.75 {\footnotesize $\pm$ 0.67}}\\ \cline{2-8}
& No Diversity Task Sampler               & 25.81 {\footnotesize $\pm$ 0.43} \textcolor{green}{\ddag} & 32.48 {\footnotesize $\pm$ 0.52} & 32.84 {\footnotesize $\pm$ 0.50} & 33.81 {\footnotesize $\pm$ 0.51} & 35.57 {\footnotesize $\pm$ 0.57} & 40.24 {\footnotesize $\pm$ 0.54}\\ \cline{2-8}
& No Diversity Batch Sampler              & \textbf{26.16 {\footnotesize $\pm$ 0.44}} \textcolor{green}{\ddag} & 41.59 {\footnotesize $\pm$ 0.59} & 38.12 {\footnotesize $\pm$ 0.53} & 34.10 {\footnotesize $\pm$ 0.49} & 37.67 {\footnotesize $\pm$ 0.58} & 39.69 {\footnotesize $\pm$ 0.55}\\ \cline{2-8}
& No Diversity Tasks per Batch Sampler    & 25.92 {\footnotesize $\pm$ 0.43} \textcolor{green}{\ddag} & 48.73 {\footnotesize $\pm$ 0.69} \textcolor{red}{\dag} & 38.69 {\footnotesize $\pm$ 0.54} \textcolor{red}{\dag} & 38.86 {\footnotesize $\pm$ 0.58} & 32.38 {\footnotesize $\pm$ 0.47} & 35.42 {\footnotesize $\pm$ 0.51}\\ \cline{2-8}
& Single Batch Uniform Sampler            & 25.62 {\footnotesize $\pm$ 0.44} \textcolor{green}{\ddag} & 38.42 {\footnotesize $\pm$ 0.54} & 39.18 {\footnotesize $\pm$ 0.58} & 35.77 {\footnotesize $\pm$ 0.57} & 40.87 {\footnotesize $\pm$ 0.62} \textcolor{green}{\ddag} & 26.39 {\footnotesize $\pm$ 0.45}\\ \cline{2-8}
& OHTM Sampler                            & 24.97 {\footnotesize $\pm$ 0.42} \textcolor{red}{\dag} & 45.73 {\footnotesize $\pm$ 0.65} & 38.00 {\footnotesize $\pm$ 0.51} \textcolor{red}{\dag} & 36.61 {\footnotesize $\pm$ 0.55} & \textbf{41.56 {\footnotesize $\pm$ 0.62}} \textcolor{green}{\ddag} & 47.09 {\footnotesize $\pm$ 0.63} \\ \hline \hline

\multirow{6}{*}{\textbf{Meta-Dataset (MSCOCO)}} & Uniform Sampler & 23.00 {\footnotesize $\pm$ 0.36} & 36.97 {\footnotesize $\pm$ 0.56} & 30.82 {\footnotesize $\pm$ 0.49} & 31.98 {\footnotesize $\pm$ 0.55} & 33.68 {\footnotesize $\pm$ 0.54} & 42.46 {\footnotesize $\pm$ 0.60}\\ \cline{2-8}
& No Diversity Task Sampler               & 23.98 {\footnotesize $\pm$ 0.39} \textcolor{green}{\ddag} & 30.53 {\footnotesize $\pm$ 0.51} & 30.47 {\footnotesize $\pm$ 0.49} \textcolor{red}{\dag} & 26.37 {\footnotesize $\pm$ 0.44} & 26.72 {\footnotesize $\pm$ 0.46} & \textbf{42.71 {\footnotesize $\pm$ 0.64}} \textcolor{red}{\dag} \\ \cline{2-8}
& No Diversity Batch Sampler              & \textbf{25.95 {\footnotesize $\pm$ 0.47}} \textcolor{green}{\ddag} & 33.83 {\footnotesize $\pm$ 0.53} & 29.32 {\footnotesize $\pm$ 0.45} & 26.42 {\footnotesize $\pm$ 0.42} & 29.25 {\footnotesize $\pm$ 0.49} & 34.89 {\footnotesize $\pm$ 0.54}\\ \cline{2-8}
& No Diversity Tasks per Batch Sampler    & 24.56 {\footnotesize $\pm$ 0.40} \textcolor{green}{\ddag} & 36.84 {\footnotesize $\pm$ 0.55} \textcolor{red}{\dag} & 32.71 {\footnotesize $\pm$ 0.52} \textcolor{green}{\ddag} & 32.80 {\footnotesize $\pm$ 0.57} \textcolor{green}{\ddag} & 31.59 {\footnotesize $\pm$ 0.51} & 32.88 {\footnotesize $\pm$ 0.51}\\ \cline{2-8}
& Single Batch Uniform Sampler            & 23.66 {\footnotesize $\pm$ 0.37} \textcolor{green}{\ddag} & 33.07 {\footnotesize $\pm$ 0.53} & \textbf{32.47 {\footnotesize $\pm$ 0.53}} \textcolor{green}{\ddag} & 33.14 {\footnotesize $\pm$ 0.56} \textcolor{green}{\ddag} & \textbf{34.53 {\footnotesize $\pm$ 0.57}} \textcolor{green}{\ddag} & 27.72 {\footnotesize $\pm$ 0.44}\\ \cline{2-8}
& OHTM Sampler                            & 23.83 {\footnotesize $\pm$ 0.39} \textcolor{green}{\ddag} & \textbf{37.22 {\footnotesize $\pm$ 0.58}} \textcolor{red}{\dag} & 31.02 {\footnotesize $\pm$ 0.49} \textcolor{red}{\dag} & \textbf{33.54 {\footnotesize $\pm$ 0.57}} \textcolor{green}{\ddag} & 33.87 {\footnotesize $\pm$ 0.56} \textcolor{red}{\dag} & 40.19 {\footnotesize $\pm$ 0.60} \\ \hline \hline

\end{tabular}%
}
\caption{Additional Performance metrics of our models on different task samplers in the 5-way 1-shot setting.}
\label{Results_meta}
\end{table*}

\begin{table*}[!htbp]
\centering
\resizebox{0.95\textwidth}{!}{%
\begin{tabular}{c|c|c|c|c|c|c|c}
\hline
\textbf{Dataset} &\textbf{Sampler} & {\textit{\textbf{MAML}}} & {\textit{\textbf{Reptile}}}  & {\textit{\textbf{Protonet}}}  & {\textit{\textbf{Matching Networks}}}  & {\textit{\textbf{MetaOptNet}}}  & {\textit{\textbf{CNAPs}}}\\ \hline
\multirow{4}{*}{\textbf{Omniglot}}
& No Diversity Tasks per Batch Sampler              & 97.76 {\footnotesize $\pm$ 0.20} & \textbf{94.55 {\footnotesize $\pm$ 0.31}} \textcolor{red}{\dag} & 97.18 {\footnotesize $\pm$ 0.25} & 93.97 {\footnotesize $\pm$ 0.40} & 96.80 {\footnotesize $\pm$ 0.27} & 90.84 {\footnotesize $\pm$ 0.47}\\ \cline{2-8}
& Single Batch Uniform Sampler        & 93.84 {\footnotesize $\pm$ 0.37} & 92.60 {\footnotesize $\pm$ 0.38} & 95.95 {\footnotesize $\pm$ 0.31} & 92.98 {\footnotesize $\pm$ 0.44} & 95.76 {\footnotesize $\pm$ 0.31} & 75.86 {\footnotesize $\pm$ 0.73}\\ \cline{2-8}
& SBU unbounded              & \textbf{98.06 {\footnotesize $\pm$ 0.19}} & 88.66 {\footnotesize $\pm$ 0.55} & 97.80 {\footnotesize $\pm$ 0.23} \textcolor{red}{\dag} & 93.82 {\footnotesize $\pm$ 0.40} & \textbf{97.18 {\footnotesize $\pm$ 0.26}} & 89.01 {\footnotesize $\pm$ 0.52} \\ \cline{2-8}
& SBU bounded            & 97.46 {\footnotesize $\pm$ 0.21} & 87.62 {\footnotesize $\pm$ 0.55} & \textbf{97.85 {\footnotesize $\pm$ 0.22}} \textcolor{red}{\dag} & \textbf{94.96 {\footnotesize $\pm$ 0.37}} \textcolor{red}{\dag} & 97.18 {\footnotesize $\pm$ 0.27} & \textbf{91.24 {\footnotesize $\pm$ 0.46}} \\ \hline \hline

\multirow{4}{*}{\textbf{MiniImagenet}}
& No Diversity Tasks per Batch Sampler              & \textbf{48.17 {\footnotesize $\pm$ 0.62}} & \textbf{41.49 {\footnotesize $\pm$ 0.56}} \textcolor{red}{\dag} & 47.73 {\footnotesize $\pm$ 0.60} & 42.54 {\footnotesize $\pm$ 0.53} & 50.60 {\footnotesize $\pm$ 0.62} & \textbf{64.11 {\footnotesize $\pm$ 0.68}} \textcolor{red}{\dag} \\ \cline{2-8}
& Single Batch Uniform Sampler        & 41.76 {\footnotesize $\pm$ 0.56} & 22.96 {\footnotesize $\pm$ 0.33} & 41.35 {\footnotesize $\pm$ 0.56} & 40.00 {\footnotesize $\pm$ 0.54} & 39.10 {\footnotesize $\pm$ 0.54} & 45.47 {\footnotesize $\pm$ 0.67}\\ \cline{2-8}
& SBU unbounded              & 45.81{\footnotesize $\pm$ 0.59} & 29.82 {\footnotesize $\pm$ 0.48} & 48.60 {\footnotesize $\pm$ 0.60} \textcolor{red}{\dag} & 42.92 {\footnotesize $\pm$ 0.56} & 52.24 {\footnotesize $\pm$ 0.64} & 53.99 {\footnotesize $\pm$ 0.67}\\ \cline{2-8}
& SBU bounded            & 45.65 {\footnotesize $\pm$ 0.59} & 30.03 {\footnotesize $\pm$ 0.49} & \textbf{48.64 {\footnotesize $\pm$ 0.61}} \textcolor{red}{\dag} & \textbf{44.21 {\footnotesize $\pm$ 0.56}} \textcolor{red}{\dag} & \textbf{52.63 {\footnotesize $\pm$ 0.64}} & 53.19 {\footnotesize $\pm$ 0.67}\\  \hline \hline

\multirow{4}{*}{\textbf{Tiered-Imagenet}} 
& No Diversity Tasks per Batch Sampler    & \textbf{49.62 {\footnotesize $\pm$ 0.69}} & \textbf{50.77 {\footnotesize $\pm$ 0.67}}\textcolor{red}{\dag} & 46.96 {\footnotesize $\pm$ 0.67} & 43.15 {\footnotesize $\pm$ 0.66} \textcolor{green}{\ddag} & 43.05 {\footnotesize $\pm$ 0.63} & \textbf{62.81 {\footnotesize $\pm$ 0.73}}\\ \cline{2-8}
& Single Batch Uniform Sampler            & 41.74 {\footnotesize $\pm$ 0.66} & 36.82 {\footnotesize $\pm$ 0.56} & 38.22 {\footnotesize $\pm$ 0.62} & 38.08 {\footnotesize $\pm$ 0.61} & 36.21 {\footnotesize $\pm$ 0.60} & 45.56 {\footnotesize $\pm$ 0.72}\\ \cline{2-8}
& SBU unbounded                           & \textbf{49.40 {\footnotesize $\pm$ 0.68}} & 43.68 {\footnotesize $\pm$ 0.61} & 50.46 {\footnotesize $\pm$ 0.68} \textcolor{green}{\ddag} & 43.08 {\footnotesize $\pm$ 0.68}  \textcolor{green}{\ddag} & \textbf{51.95 {\footnotesize $\pm$ 0.69}} \textcolor{green}{\ddag} & 56.23 {\footnotesize $\pm$ 0.74}\\ \cline{2-8}
& SBU bounded                          & 47.33 {\footnotesize $\pm$ 0.68} & 44.51 {\footnotesize $\pm$ 0.62} & \textbf{51.11 {\footnotesize $\pm$ 0.69}} \textcolor{green}{\ddag} & \textbf{43.23 {\footnotesize $\pm$ 0.69}} \textcolor{green}{\ddag} & 51.93 {\footnotesize $\pm$ 0.69} \textcolor{green}{\ddag} & 52.77 {\footnotesize $\pm$ 0.74}\\ \hline \hline

\end{tabular}%
}
\caption{Performance metric of our models on SBU oriented task samplers in the 5-way 1-shot setting.}
\label{Results_Single}
\end{table*}

\begin{figure}[!htbp]
    \centering
    \begin{adjustbox}{center}
    \includegraphics[height=0.4\textheight, keepaspectratio]{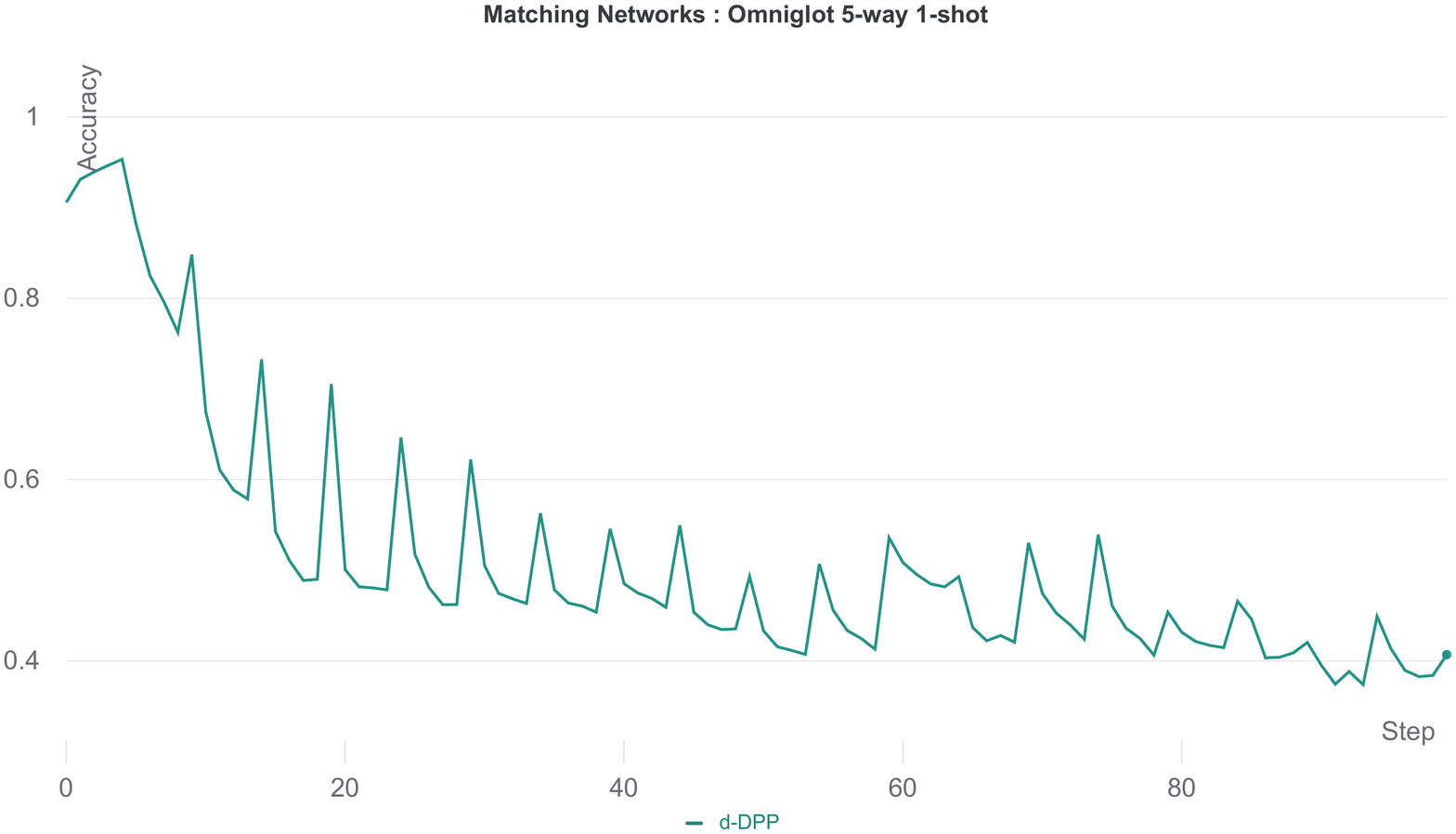}
    \end{adjustbox}
    \caption{Convergence curve of Matching Networks model on Omniglot 5-way 1-shot.}
    \label{fig:matching_networks}
\end{figure}

%%%%%%%%%%%%%%%%%%%%%%%%%%%%%%%%%%%%%%%%%%%%%%%%%%%%%%%%%%%%%%%%%%%%%%%%%%%%%%%
%%%%%%%%%%%%%%%%%%%%%%%%%%%%%%%%%%%%%%%%%%%%%%%%%%%%%%%%%%%%%%%%%%%%%%%%%%%%%%%

\newpage
\end{document}